%% file: Main.tex
\documentclass[acmlarge,nonacm]{acmart}
\usepackage{subcaption}
\usepackage{wasysym}
\usepackage{makecell}
\usepackage{enumitem}
\usepackage{soul}
\usepackage{color, xcolor}
\usepackage{pifont}
\usepackage{placeins}
\AtBeginDocument{%
  \providecommand\BibTeX{{%
    \normalfont B\kern-0.5em{\scshape i\kern-0.25em b}\kern-0.8em\TeX}}}

\setcopyright{none}

\acmJournal{IMWUT}
\acmVolume{1}
\acmNumber{1}
\acmArticle{1}
\acmMonth{5}

\renewcommand{\authorsaddresses}{}
\renewcommand\footnotetextcopyrightpermission[1]{} % removes footnote with conference information in first column
\input{preambles}
\begin{document}

% \title{Enhancing the Capabilities of a Large Language Model-Based Virtual Doctor with Sensor Data Knowledge}
\newcommand{\cmark}{\ding{51}}%
\newcommand{\xmark}{\ding{55}}%

\newcommand{\workname}{SocialMind} 

\newcommand{\result}{ToFill} 
% \title{\workname: A Personalized Social Assistant for Real-time Conversational Guidance}
% \title{\workname: An LLM-empowered Proactive Social Assistive System for Live Social Interactions}
% \title{\workname: An Eyewear Proactive Social Assistive System for Live Social Interactions Leveraging LLMs and Sensor Data}

% \title{\workname: A Proactive Social Assistive System for Live Social Interactions Leveraging LLMs and Sensor Data on AR Glasses}

% \title{\workname: Integrating Large Language Models and AR Glasses for Proactive Social Assistance in Live Interactions}

\title{\workname: A Proactive Social Assistive System for Live Social Interactions Integrating Large Language Models and AR Glasses}

\title{\workname: An LLM Empowered Proactive Social Assistive System for Live Social Interactions with Human-Like Perception Using AR Glasses}

\title{\workname: A Proactive Social Assistive System with Human-Like Perception for Live Interactions Empowered by LLMs and AR Glasses}

\title{\workname: Proactive AR Social Assistance with Large Language Models for In-situ Live Interactions}

\title{\workname: LLM-based Proactive AR Social Assistive System with Human-like Perception for In-situ Live Interactions}

% YQ Suggestion:
% SocialSprite: Enchanting Assistive AR Glasses for Proactive, Human-like Social Companionship

% \affiliation{%

% \affiliation{%

% \affiliation{%

\author{Bufang Yang}
\authornote{Co-primary authors.}
\affiliation{%
  \institution{The Chinese University of Hong Kong}
  % \city{Hong Kong SAR}
  \country{China}}
\email{bfyang@link.cuhk.edu.hk}

\author{Yunqi Guo}
\authornotemark[1]
\affiliation{%
  \institution{The Chinese University of Hong Kong}
  % \city{Hong Kong SAR}
  \country{China}}
\email{yunqiguo@cuhk.edu.hk}

\author{Lilin Xu}
\authornotemark[1]
\affiliation{%
  \institution{Columbia University}
  % \city{New York}
  \country{United States}}
\email{lx2331@columbia.edu}

\author{Zhenyu Yan}
\affiliation{%
  \institution{The Chinese University of Hong Kong}
  % \city{Hong Kong SAR}
  \country{China}}
\email{zyyan@ie.cuhk.edu.hk}

\author{Hongkai Chen}
\affiliation{%
  \institution{The Chinese University of Hong Kong}
  % \city{Hong Kong SAR}
  \country{China}}
\email{hkchen@ie.cuhk.edu.hk}

\author{Guoliang Xing}
\affiliation{%
  \institution{The Chinese University of Hong Kong}
  % \city{Hong Kong SAR}
  \country{China}}
\email{glxing@cuhk.edu.hk}

\author{Xiaofan Jiang}
\affiliation{%
  \institution{Columbia University}
  % \city{New York}
  \country{United States}}
\email{jiang@ee.columbia.edu}

% \affiliation{%

% \affiliation{%

% \affiliation{%

% \affiliation{%

% \affiliation{%


\begin{abstract}

Social interactions are fundamental to human life.
The recent emergence of large language models (LLMs)-based virtual assistants has demonstrated their potential to revolutionize human interactions and lifestyles.
However, existing assistive systems mainly provide reactive services to individual users, rather than offering in-situ assistance during live social interactions with conversational partners.
In this study, we introduce \workname, the first LLM-based proactive AR social assistive system that provides users with in-situ social assistance.
\workname~employs human-like perception leveraging multi-modal sensors to extract both verbal and nonverbal cues, social factors, and implicit personas, incorporating these social cues into LLM reasoning for social suggestion generation.
Additionally, \workname~employs a multi-tier collaborative generation strategy and proactive update mechanism to display social suggestions on Augmented Reality (AR) glasses, ensuring that suggestions are timely provided to users without disrupting the natural flow of conversation.
Evaluations on three public datasets and a user study with 20 participants show that \workname~ achieves 38.3\% higher engagement compared to baselines, and 95\% of participants are willing to use \workname~in their live social interactions.

\end{abstract}

\settopmatter{printfolios=true}
\maketitle
\pagestyle{plain}
% \newcommand{\cmt}[1]{\textcolor{blue}{#1}}
% \newcommand{\qus}[1]{{\color{red}{#1}}}
% \newcommand{\xl}[1]{{\color{magenta}{#1}}}

% blend into the perception, 55.54s

\input{Introduction/Introduction.tex}
\input{Related_works/Related_works.tex}
\input{Motivation/Motivation.tex}

\input{System_design/System_design.tex}
\input{Evaluation/Evaluation.tex}
\input{Discussion/Discussion.tex}

\section{Conclusion}
This paper introduces \workname, the first proactive social assistive system capable of providing users with in-situ assistance during live interactions. 
\workname~ employs a human-like perception approach leveraging multi-modal sensors on AR glasses to extract social cues.
% \workname~ incorporates perceived verbal and nonverbal cues, social factors, and implicit persona cues for LLM reasoning and the generation of social suggestions.
Additionally, \workname~employs a multi-tier collaborative reasoning strategy to provide instant social suggestions, allowing users to refer to them without disrupting the flow of conversation. 
% Meanwhile, \workname~performs implicit personas adaptation to generate customized social suggestions that enhance engagement for both parties. 
Results from three public datasets and a user study show that \workname~ achieves 38.3\% higher engagement compared to baselines, and 95\% of participants are willing to use \workname.

% Results from three public datasets and a user study with 20 participants show that \workname~achieves 38.3\% higher engagement than baselines and 95\% participants are willing to user it.

% demonstrate the effectiveness of \workname.

% This paper introduces \workname~, the first LLM-based proactive social assistive system for live social interactions.
% \workname~employs a human-like perception approach to proactively perceive multi-modal socil cues during social interactions.
% Additionally, \workname~employs a fast-slow collaborative reasoning strategy to provide user social suggestions in real time, so that users can refer to these suggestions without disrupting the natural flow of conversation.
% Meanwhile, \workname~employs an implicit personas adaptation approach to generate customized social suggestion that can enhance the engagement of both parties.
% Results on three public datasets and a user study of 20 participants show the effectiveness of \workname.

\bibliographystyle{ACM-Reference-Format}
\bibliography{socialcoach}

\newpage
\appendix
\section{Appendix: Prompt Settings}
\label{appendix}
\FloatBarrier
% \vspace{-3em} % 调整章节标题下方的间距

\begin{figure}[h]
\begin{minipage}[t]{0.48\columnwidth}
     \centering
\includegraphics[width=1\textwidth]{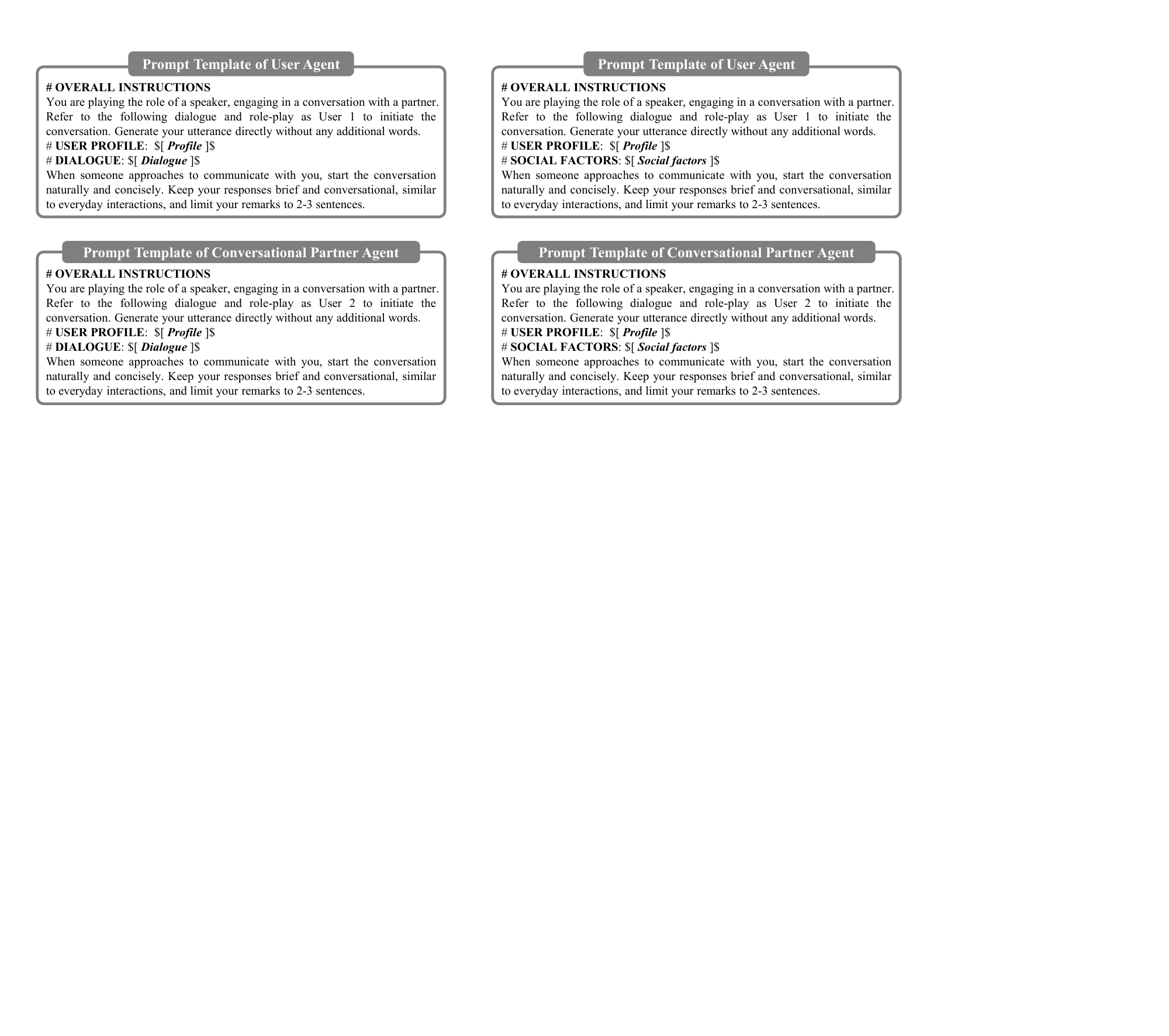}
\vspace{-1.5em}
  \caption{
  Prompt templates of the user agent and the conversational partner agent in the dialogue-based role-play.}
  % \vspace{-.5em}
\label{fig:prompt_agent_datasets}
\end{minipage}
\hfill
  \begin{minipage}[t]{0.48\columnwidth}
     \centering
\includegraphics[width=1.\textwidth]{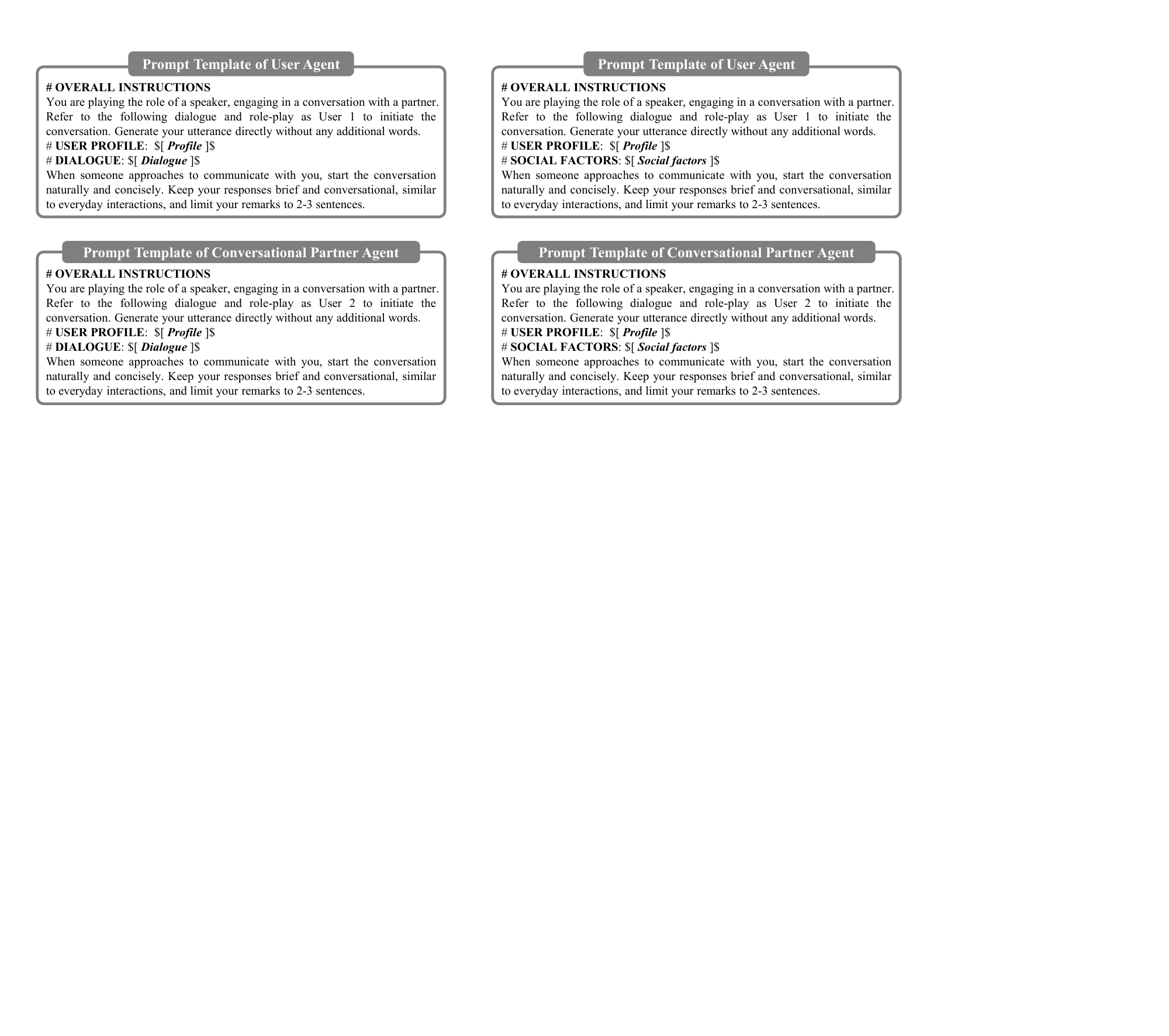}
\vspace{-1.5em}
  \caption{Prompt templates of the user agent and the conversational partner agent in the social factor-based role-play.
  }
  % \vspace{-.5em}
\label{fig:prompt_agent_social}
\end{minipage}
% \vspace{-1em}
\end{figure}

\begin{figure}[h]
  \centering
\includegraphics[width=0.75\linewidth]{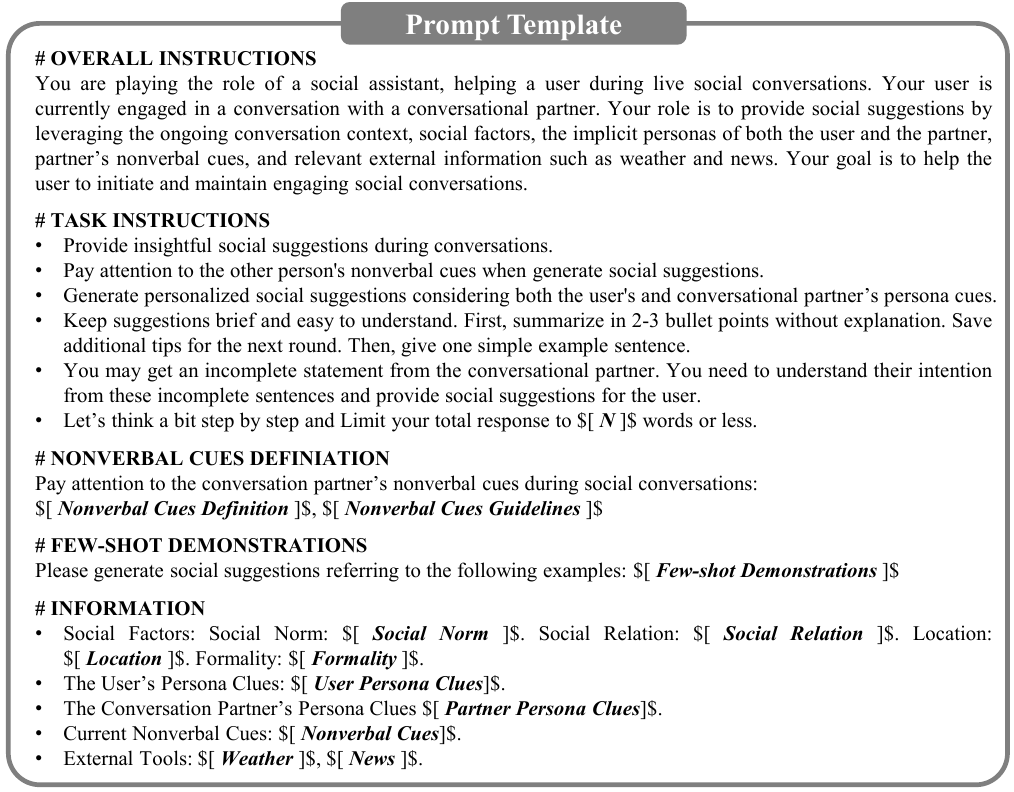}
\vspace{-.5em}
  \caption{Prompt template of \workname.}
  % \vspace{-.5em}
\label{fig:prompt_overall}
  \vspace{-1em}
\end{figure}

\begin{figure}[h]
  \centering
\includegraphics[width=0.65\linewidth]{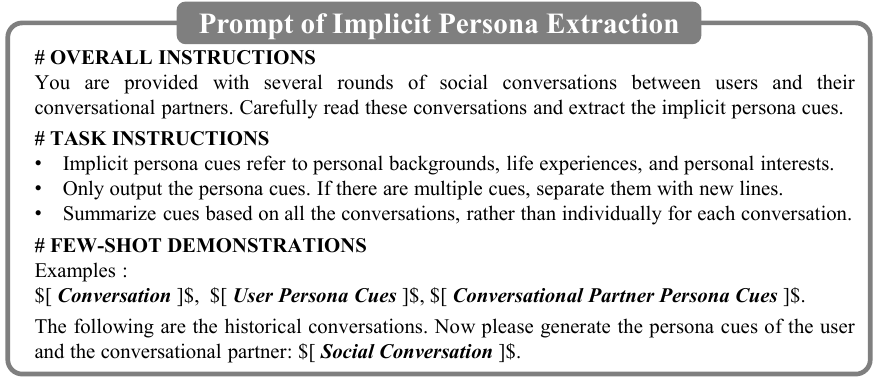}
\vspace{-0.5em}
  \caption{Prompt of implicit persona extraction in \workname.}
  % \vspace{-.5em}
\label{fig:persona}
  % \vspace{-1em}
\end{figure}

\begin{figure}
\begin{minipage}[t]{0.48\columnwidth}
     \centering
\includegraphics[width=1\textwidth]{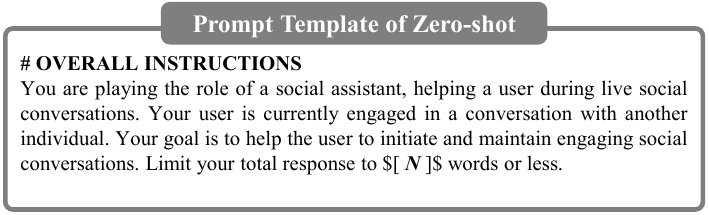}
\vspace{-1.5em}
  \caption{
  Prompt of Zero-shot baseline approach.}
  % \vspace{-.5em}
\label{fig:prompt-zeroshot}
\end{minipage}
\hfill
  \begin{minipage}[t]{0.48\columnwidth}
     \centering
\includegraphics[width=1.\textwidth]{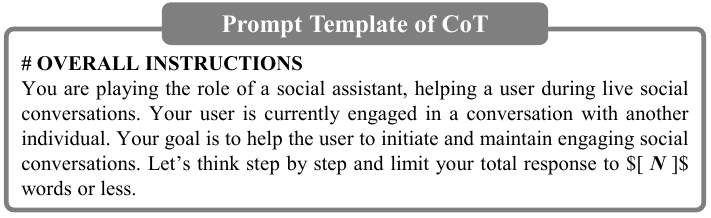}
\vspace{-1.5em}
  \caption{ Prompt of CoT baseline approach.
  }
  % \vspace{-.5em}
\label{fig:prompt-cot}
\end{minipage}
% \vspace{-1em}
\end{figure}

\begin{figure}
  \centering
\includegraphics[width=0.8\linewidth]{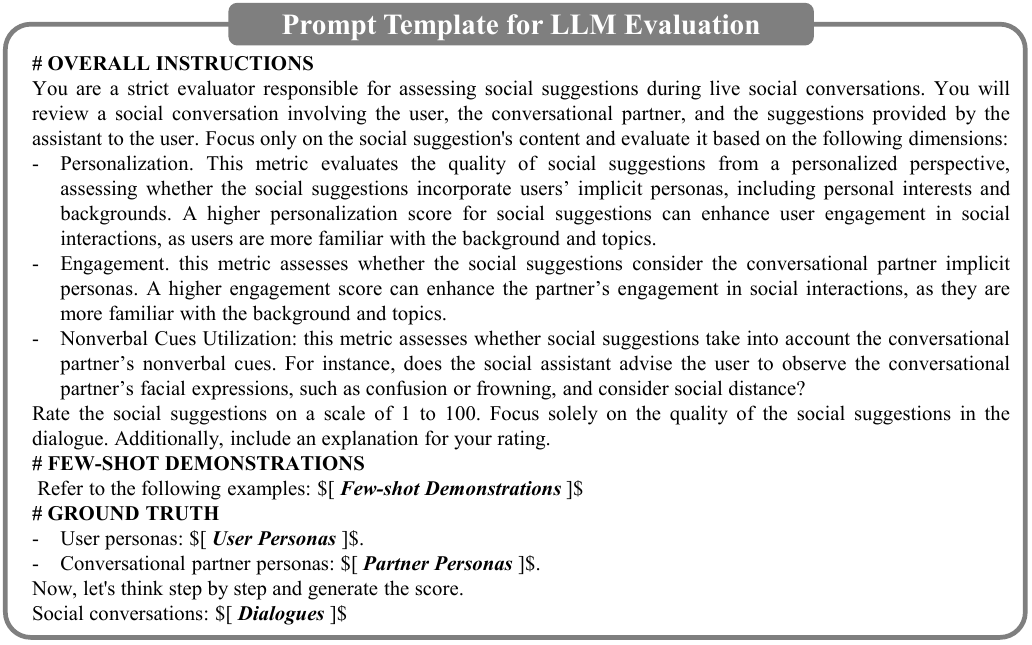}
% \vspace{-1em}
  \caption{Prompt template of the LLM evaluation.}
  % \vspace{-.5em}
  \label{fig:prompt-eval}
  \vspace{-1em}
\end{figure}

\end{document}

%% file: Introduction/Introduction.tex
\section{Introduction}
% Social interactions are xxx.
% Prior to LLMs systems xxx.
% In the realm of LLMs, many systems xxx.
% They can xxx.

% social interactions + LLM-based personal assistents.
% social assistivee systems. limitations. user survey reveals that we need a xx.

% However, these personal assistants can only xxx.
% 1. non-social: reactive response
% 2. social: post-processing

% Social User Survey.

% As a key determinant of quality of life, positive social interactions significantly influence both physical and mental health. Positive social interactions enhance communication skills and alleviate the effects of stress on well-being.

% As a crucial determinant of quality of life, social interactions significantly impact both physical and mental health \cite{social_interactions_QoE}. Positive social interactions enhance communication skills and alleviate stress, thereby improving overall well-being.
% As a crucial determinant of quality of life, social interactions significantly impact both physical and mental health by enhancing communication skills and alleviating stress \cite{social_interactions_QoE}. 
Social interactions are a crucial determinant of quality of life, significantly impacting both physical and mental health by enhancing communication skills and alleviating stress \cite{social_interactions_QoE}.
However, over 15 million individuals in America experience social anxiety when anticipating or engaging in social interactions \cite{social-anxiety-disorder}. Even people without social anxiety may feel anxious when interacting with certain individuals, such as senior managers and unfamiliar colleagues, which can reduce their overall well-being.
With the surge of Large Language Models (LLMs) and their reasoning capabilities \cite{achiam2023gpt,touvron2023llama}, numerous LLM-based personal virtual assistants have been developed to enhance individuals' overall well-being.
% With the surge of Large Language Models (LLMs) and their remarkable reasoning capabilities \cite{achiam2023gpt,touvron2023llama}, numerous LLM-based personal virtual assistants have been developed to enhance individuals' overall well-being.
% Typical virtual assistants such as writing assistants [xx], coding assistants [xx], and fitness assistants [xx], have emerged.
However, existing LLM-based personal virtual assistants, such as writing assistants~\cite{gao2024aligning,mysore2023pearl}, fitness assistants~\cite{wang2024ubiphysio,yang2024drhouse}, and coding assistants~\cite{englhardt2024exploring}, focus solely on serving individual users rather than supporting live social interactions with conversational partners.

\begin{figure}
  \centering
  \captionsetup{skip=5pt}
\includegraphics[width=0.95\linewidth]{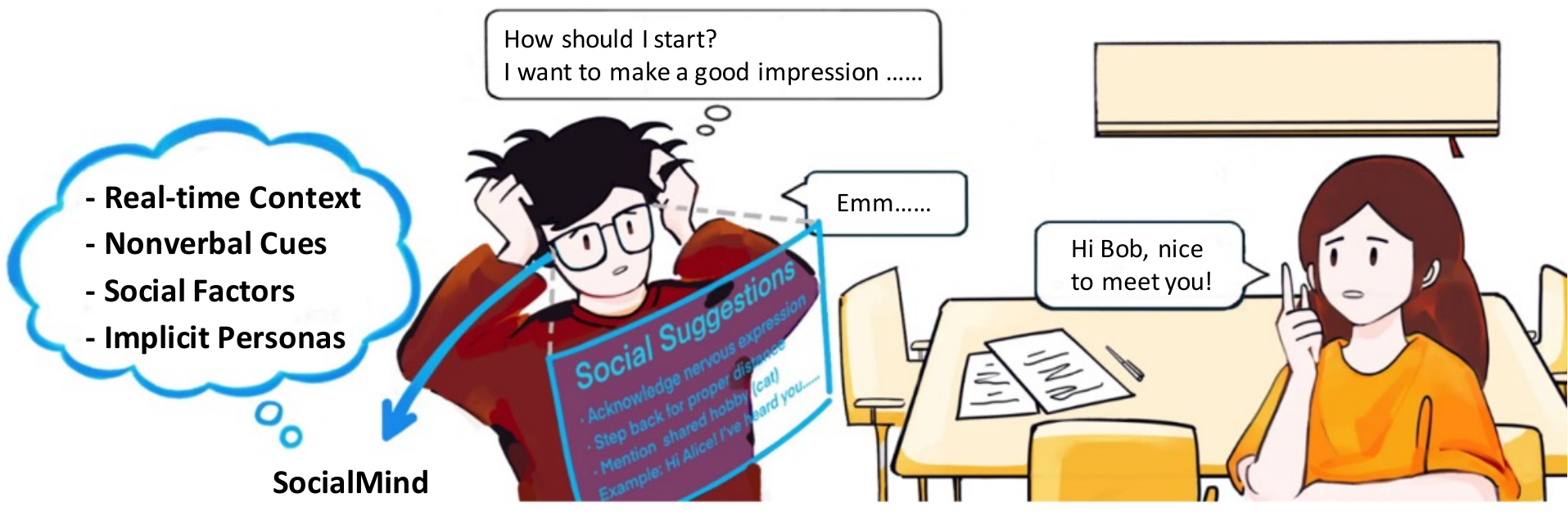}
\vspace{-.5em}
  \caption{Overview of \workname.
\workname~provides in-situ social assistance to the user to help the user during live social interactions with conversational partners.
\workname~automatically performs human-like social perception, generates social suggestions to assist users, and proactively displays them on the user's AR glasses as the conversation proceeds. 
Users can seamlessly refer to these suggestions while interacting with conversational partners.
% \workname~in situ to the live social interactions involving conversational partners and provide real-time social assistance to the user.
% Users engage in live social interactions with the help of \workname. 
% perceives both verbal and nonverbal behaviors, social factors, and implicit personas to assist users in their social interactions.
% \zy{font size of the suggestions is too small. Can put two itemized suggestions and the example only.}
  }
  % \vspace{-1.5em}
  \label{fig:overview}
  \vspace{-1.3em}
\end{figure}

% Although many LLM-based social assistive systems have been proposed recently, they primarily perform social-related question-answering [xx], behavior predictions [xx], or detect and remediate social violations [xx] in text input as a post-processing step.

% To help with xxx
% 社会文化冲突，自闭症，人情世故，开发了一些social 大模型。
% 文本only，不是实时社交过程，包含第三方社交时的社交建议
% 真实的社交辅助需要感知更多的信息如both现场的对话context，多模态的nonverbal 行为，以及周围的环境等社交因素决定，来动态的调整实时的社交建议。

% Many LLM-based social assistive systems have been proposed recently, aiming to support autistic patients \cite{jang2024s}, provide knowledge consultation on social etiquette \cite{tianji2024}, and resolve cultural conflicts in communication~\cite{hua2024assistive,zhan2024let,hua2024sadas,zhan2024renovi}.
In addition to general virtual assistants, LLM-based social assistive systems have been proposed recently, aiming to support autistic patients \cite{jang2024s}, provide knowledge consultation on social etiquette \cite{tianji2024}, and resolve cultural conflicts in communication~\cite{hua2024assistive,zhan2024let,zhan2024renovi}.
These assistive systems either function in a reactive ``query-response'' manner to address users' explicit social-related questions \cite{jang2024s,tianji2024}, or act as post-processing modules to detect and remediate norm violations in social conversations \cite{hua2024assistive,hua2024sadas}, rather than providing in-situ assistance during live social interactions with conversational partners.
% However, human xxx. % therefore, the social assistive system should be human-like to proactively xx.
% However, during social interactions, humans perceive diverse information, including the context of the live conversation, the multi-modal nonverbal behaviors of other parties\cite{duncan1969nonverbal}, and various social factors\cite{zhan2023socialdial}, and leverage this information to adjust their social strategies.
% However, assistive systems in live social interactions must act like humans to proactively perceive diverse information, including the context of the live social conversation, the multi-modal nonverbal behaviors of other parties~\cite{duncan1969nonverbal}, and various social factors \cite{zhan2023socialdial}.
However, human face-to-face social interactions are complex behaviors that necessitate considering various types of information, including verbal, and non-verbal behaviors, social environment, social purpose, and the personal backgrounds of both parties \cite{hall2019nonverbal}.
Therefore, assistive systems in live social interactions must act like humans to perceive diverse information, such as the context of the live social conversation, the multi-modal nonverbal behaviors of other parties \cite{duncan1969nonverbal}, and various social factors \cite{zhan2023socialdial}.
They should also incorporate this information for reasoning and dynamically adjust their strategies for providing instant social suggestions to the user.
This highlights a research gap in providing proactive social assistance during live, face-to-face social interactions involving conversational partners.

To address this research gap, we propose a proactive social assistive system for live social interactions leveraging LLMs and the multi-modal sensor data from Augmented Reality (AR) glasses.
However, several unique challenges remain in developing such a system.
% To address this research gap, we propose an LLM-based eyewear social assistive system capable of providing real-time social suggestions during live interactions for the first time. 
% However, several unique challenges remain in developing such a proactive assistive system for live social interactions.
% First, \textit{an assistive system for live social interactions requires real-time feedback to the user to avoid disrupting the natural flow of conversations.}
First, \textit{an assistive system for live social interactions requires providing users with instant and in-situ assistance during natural conversations with other parties.
}
% to the user in-situ within the natural flow of conversation with other parties.
Unlike existing personal assistants and assistive systems designed for single users \cite{xu2024can,
wang2024ubiphysio,liu2024tasking,
gao2024aligning}, live social interactions involve conversational partners, presenting additional challenges in providing instant responses to the user without disrupting the natural flow of conversation.
Second, \textit{nonverbal behaviors are essential for social communication, yet they are challenging for LLMs to comprehend}.
LLMs are trained exclusively on text corpora, whereas nonverbal behaviors such as facial expressions, gestures, and physical proximity involve multi-modal information \cite{duncan1969nonverbal}, posing challenges for LLMs in understanding and integrating these cues to generate nonverbal cue-aware social suggestions.
% LLMs are trained exclusively on text corpora. 
% However, nonverbal behaviors such as facial expressions, gestures, and physical proximity involve multi-modal information \cite{duncan1969nonverbal}, which poses challenges for LLMs in comprehending and integrating these cues to generate nonverbal cue-aware social suggestions.
Third, social suggestions need to consider the implicit personal backgrounds and interests of both parties to enhance their engagement.
However, \textit{natural social conversations often lack explicit queries for knowledge retrieval, posing challenges for system personalization.
}
How to integrate implicit personas like personal interests and background into social suggestions remains challenge.

In this paper, we introduce \workname, the first LLM-based proactive AR social assistive system that provides users with in-situ social assistance in live social interactions with other parties.
Figure~\ref{fig:overview} shows the overview of \workname.
\workname~leverages the multi-modal sensor data on AR glasses to perform human-like perception, including verbal and nonverbal cues and social factor information.
Additionally, \workname~ extracts the implicit personas of both parties through social interactions.
Then, \workname~integrates these cues and utilizes a multi-tier collaborative social suggestion generation strategy that incorporates a cache with social-factor priors and an intention infer-based reasoning approach.
This strategy enables \workname~ to provide timely, in-situ social assistance to the user during natural conversations with partners. 
Through a proactive response mechanism, social suggestions are displayed on AR glasses, enabling the user to seamlessly refer to them while interacting with their conversational partner.
We summarize the contributions of this paper as follows.

\begin{itemize}[leftmargin=*]

\item
We introduce \workname, the first LLM-based proactive AR social assistive system providing users with \textbf{in-situ assistance} during live social interactions.
We develop a multi-tier collaborative suggestion generation strategy, incorporating a social factor-aware cache and intention infer-based reasoning, along with a proactive update mechanism. This ensures users receive timely and in-situ social suggestions, which are displayed on AR glasses without disrupting the natural flow of conversation.
% We develop a fast-slow collaborative suggestion generation strategy, incorporating a social factor-aware cache and intention inference-based reasoning, along with a proactive update mechanism. This ensures users receive in-situ social suggestions, which are proactively displayed on AR glasses without disrupting the natural flow of conversation.

\item 
We design a human-like perception mechanism that enables \workname~to automatically leverage multi-modal sensor data to perceive social cues, and develop a multi-source social knowledge reasoning approach to incorporate these cues into LLM reasoning, dynamically adjusting strategies for social assistance.

\item 
We develop an implicit persona adaptation approach that enables \workname~to generate customized social suggestions, enhancing the engagement of both parties in live social interactions.

\item 
Motivated by a user survey involving 60 participants to understand their social experiences and preferences for social assistance, we designed and implemented \workname~on AR glasses and validated its effectiveness using three public datasets and real-world tests. Evaluations on these datasets and a user study with 20 participants revealed that \workname~achieves a 38.3\% higher engagement compared to baselines, with 95\% of participants expressing a willingness to use \workname~in live social interactions.

\end{itemize}

% \noindent\textbf{Highlights}.
% Existing social assistive systems primarily function as reactive systems, refining the user's explicit input to make it more suitable for social conversations [xxx].
% However, there is a gap when these systems are used in live social interactions and face-to-face conversations. In this study, we propose a proactive assistive system for live social interactions called \workname, featuring the following three unique aspects:
% \begin{itemize}[leftmargin=*]

% \item 
% We developed a fast-slow collaborative suggestion generation strategy, which includes a fast, intention-inferred draft using limited words from the opponent, followed by a more in-depth but slower suggestion based on the opponent's complete input.

% \item 
% We developed a nonverbal cues interaction approach that enables \workname~to proactively perceive and integrate nonverbal cues, generating nonverbal cue-aware social suggestions.

% \item 
% We developed an implicit persona adaptation mechanism that allows \workname~to generate social suggestions by considering the implicit personas of both parties.

% \end{itemize}

%% file: Related_works/Related_works.tex
\vspace{-.5em}
\section{Related work}
% Table~\ref{tab:Comparison} summarizes the recent LLM-based applications in social and communication.

\subsection{LLM-based Personal Assistants}
% \noindent\textbf{Personal Assistant}.
% \textcolor{red}{[Traditional Personal Assistant like Siri].}
% Many personal assistant have been widely used in our daily life on diverse commercial mobile devices like smartphones, like Apple's Siri [xxx] and Google Assistant [xxx].
Voice assistants are widely used in daily lives on various commercial mobile devices, such as Apple's Siri \cite{siri} and Google Assistant \cite{google_assistant}. 
% Recently, with the surge of LLMs, LLM-based virtual assistants have been developed, such as fitness assistants \cite{wang2024ubiphysio,yang2024drhouse}, writing assistants \cite{gao2024aligning,mysore2023pearl}, and coding assistants \cite{englhardt2024exploring}.
Recently, LLM-based virtual assistants have been developed, such as fitness assistants \cite{wang2024ubiphysio,yang2024drhouse,yang2024viassist}, writing assistants \cite{gao2024aligning,mysore2023pearl}, and coding assistants \cite{englhardt2024exploring}.
% numerous studies have developed LLM-based personal assistants such as fitness assistants [xxx], writing assistants [xxx], and coding assistants [xxx].
OS-1 \cite{xu2024can} is a virtual companion on smart glasses offering companionship by recording daily activities and chatting with users.
% [before condense] OS-1 \cite{xu2024can} is a personal virtual companion that provides human-like companionship by using smartglasses to perceive the physical world, record users' daily activities and events, and chat with them.
% However, OS-1 solely focuses on single-user conversations, rather than providing assistance during live, face-to-face social interactions with the other party.
% that uses glasses as a carrier to share the user's vision, record the user's daily events, chatting with the user and enabling companionship.
% conversational system that uses smart glasses to record the user's daily activities and events, enabling chatbot companionship.
UbiPhysio \cite{wang2024ubiphysio} is a fitness assistant that provides natural language feedback for daily fitness and rehabilitation exercises, improving workout quality.
% [before condense] UbiPhysio \cite{wang2024ubiphysio} is a personal fitness assistant that provides users with natural language descriptions and feedback for their daily fitness and rehabilitation exercises, enhancing the quality of their workouts.
Moreover, recent studies develop personal assistants for older adults \cite{gao2024easyask,yang2024talk2care} and individuals with impairments \cite{jing2024anglesizer}.
% [before condense] Additionally, many studies develop personal assistant systems for older adults \cite{gao2024easyask,yang2024talk2care} and individuals with impairments \cite{jing2024anglesizer}.
EasyAsk \cite{gao2024easyask} is a search assistant for older adults, accepting both audio and text inputs to provide app tutorials based on their queries. 
% [before condense] EasyAsk \cite{gao2024easyask} is a search assistant designed for older adults. It accepts both audio and text inputs, understands their queries, and extracts intentions to find relevant app tutorials.
Talk2Care \cite{yang2024talk2care} is a voice assistant designed to engage in conversations with older adults to gather health information for healthcare providers.
% [before condense] Talk2Care \cite{yang2024talk2care} is a voice assistant designed to engage in conversations with older adults and extract health-related information for healthcare providers' reference.
% EasyAsk \cite{gao2024easyask} is a search assistant developed for the elder people, that can take both audio and text as input, understand the queries from the elder people and extract their intentions to search for the relevant app tutorials for the elder.
Additionally, studies like PEARL \cite{mysore2023pearl} and PRELUDE \cite{gao2024aligning} develop LLM-based writing assistants that adapt outputs to user preferences using retrieval augmentation \cite{mysore2023pearl} or interactive learning \cite{gao2024aligning}.
% Additionally, studies such as PEARL \cite{mysore2023pearl} and PRELUDE \cite{gao2024aligning} develop LLM-based personal writing assistants. They can adapt the generation results to user preferences through retrieval augmentation \cite{mysore2023pearl} or interactive learning \cite{gao2024aligning}.
% Additionally, several studies like PEARL \cite{mysore2023pearl} and PRELUDE \cite{gao2024aligning} develops LLM-based personal writing assistant that can adapt the generation results to the user preference by retrieval augmentation \cite{mysore2023pearl} or interactive learning \cite{gao2024aligning}.
% PEARL \cite{mysore2023pearl} is a writing assistant.
% PRELUDE \cite{gao2024aligning} develops an LLM-based personal writing assistant that aligns with user preferences through interactive learning.
% \textcolor{red}{[More mobile agent: writing assistant, code assistant, personalization is important for these applications].}
% In addition, many recent works have proposed personal AI assistants, such as Gemini Live [xxx], which can assist with daily tasks on mobile phones, like adding items to a shopping list.
% Mini-Omni [xxx] integrates hearing, speaking, and thinking into speech foundation models, enabling real-time conversation capabilities.
However, these systems focus solely on single-user human-to-computer interactions, considering only the user's unilateral goals and inputs.
\workname~takes a further step by providing users with social assistance during live, face-to-face interactions involving other parties.
% with \textcolor{blue}{third-party participants}. 
% \yq{Should we place this section after the }

% \textcolor{blue}{However, these systems only focus on single user human-to-computer interaction rather than live social interactions with the third party, considering only the user's unilateral goals and inputs.
% }

% \workname~ targets on the assistive system that can  real-time social assistance in human-to-human conversation scenarios.

% \noindent\textbf{Proactive Conversational Systems}.

\vspace{-.5em}

\subsection{Social Assistive Systems}
\noindent\textbf{Pre-LLM Era.}
SocioGlass \cite{xu2016socioglass} builds a biography database, using smart glasses and facial recognition to retrieve profiles with background and interests for social interaction assistance.
Another study explores the use of smart glasses to support social skills learning in individuals with autism spectrum disorder \cite{keshav2017social}.
However, these systems are limited to displaying social skills or biographies on-screen, lacking the context of real-time social conversation.
% [before condense]
% SocioGlass \cite{xu2016socioglass} creates a database of people's biographies. 
% The system is deployed on smart glasses and uses facial recognition to retrieve profiles, including background and interests, for social interaction assistance.
% Another study examines the usability of smart glasses for social skills learning in individuals with autism spectrum disorder \cite{keshav2017social}, indicating their potential for social assistance.
% However, these systems can only display social skills or retrieve and show a person's biography on-screen, lacking the context of real-time social conversation.

\noindent\textbf{LLMs for Social Assistance.}
% \textcolor{red}{[LLM for social science].}
Paprika \cite{jang2024s} employs LLMs to provide social advice to autistic adults in the workplace. Results show that autistic workers prefer interactions with LLMs, demonstrating LLMs' potential to offer social advice.
% Tianji \cite{tianji2024} is an LLM that understands the ways of the world. 
% It can provide users with detailed social skills guidance in the form of answering questions, such as how to resolve conflicts or ease embarrassment.
Tianji \cite{tianji2024} is an LLM that comprehends social dynamics, offering social skill guidance by answering questions, like how to resolve conflicts.
% [before condense] Tianji \cite{tianji2024} is an LLM that comprehends social dynamics. It offers users detailed guidance on social skills by answering questions, such as how to resolve conflicts or alleviate embarrassment.
Social-LLM \cite{jiang2023social} integrates users' profiles and interaction data to generate user embeddings for user detection.
% Social-LLM \cite{jiang2023social} integrates users' profiles and interaction data from social platforms to generate user embeddings for user detection.
% [before condense] Social-LLM \cite{jiang2023social} is a predictive model that integrates users' profiles and interaction data from social platforms to generate user embeddings for user detection.
However, these works are reactive conversational systems limited to social Q\&A or user behavior prediction, rather than providing instant social assistance when users are interacting with others.
% lacking the capability to support live, face-to-face interactions with \textcolor{blue}{third-party participants} in the real world.
% [before condense] However, these systems are reactive conversational systems that perform social-related question-answering or predict user behavior without involving real third-party participants. Consequently, they cannot support live face-to-face social interactions in the real world.
% However, these systems are reactive conversational systems, performing question-answering or predicting user behavior without involving a real third-party participant, thus can not support live face-to-face social interactions in the real world.
% \textcolor{blue}{
% However, these systems are not designed to support live social interactions and function more like reactive systems, performing question-answering or predicting user behaviour without involving a real third-party participant.
% % However, these systems only provide social suggestions prior to the actual conversation, functioning more like a creative system and question-answering system without the involvement of a real third-party participant.
% }
% \textcolor{red}{[LLM for negotiation.]}
% Some studies in the NLP field \cite{hua2024assistive,zhan2024let,hua2024sadas,zhan2024renovi} explore the impact of social norms and their violations in communication and negotiation, using simulations with multiple LLM agents.
Some studies also explore the impact of social norms and their violations in communication and negotiation, using simulations with multiple LLM agents~\cite{hua2024assistive,zhan2024let,hua2024sadas,zhan2024renovi}.
SADAS \cite{hua2024sadas} is a dialogue assistant that checks user input for social norm violations to improve cross-cultural communication.
% [before condense] SADAS \cite{hua2024sadas} develops a dialogue assistant system that can examine and remediate user input to ensure it does not contain violations of social norms, thereby enhancing cross-cultural conversations.
Kim \textit{et. al}~\cite{kim2022prosocialdialog} develops a dialogue model to detect unsafe content and generate prosocial responses.
% [before condense] Kim \textit{et. al}~\cite{kim2022prosocialdialog} develops a dialogue model that can detect unsafe content in the dialogue and generate more prosocial responses.
However, these systems provide post-assistance, addressing social norm violations in user text-only input only after it has been entered.
% [before condense] However, these systems serve as post-assistance solutions, aiming to detect and remediate cultural violations of social norms in user text-only input after it has been entered. 
% Moreover, they do not consider practical face-to-face social conversations, which require generating real-time social suggestions. 
\workname~focuses on live face-to-face scenarios, proactively perceiving multi-modal nonverbal cues and conversation context to provide instant social suggestions, enabling users to refer to them before speaking. 
% on how to provide \textbf{personalized conversation topics} and initiate the conversation \textbf{before the user speaks}.
% Post-remediate rather than 

% \footnotesize
% \begin{table}[ht]\footnotesize
\begin{table}[t]\footnotesize
  \caption{A summary of the recent LLM-based applications in social and communication (\CIRCLE \ means included).}
  \vspace{-1em}
  \label{tab:Comparison}
  \begin{tabular}{cccccccc}
    \toprule
    \textbf{Approach} & \textbf{Base LLM} & \makecell{\textbf{Social}\\ \textbf{Assistance}} & \makecell{\textbf{Multi Party}\\\textbf{Interactions}}  &\makecell{\textbf{Multi-modal} \\ \textbf{Sensor Data}} & \makecell{\textbf{Persona-}\\ \textbf{lization}} &
    \makecell{\textbf{Interactive}\\ \textbf{Mode}} & \makecell{\textbf{System}\\ \textbf{Settings}} \\
    \midrule
    Social-LLM \cite{jiang2023social}& Sentence-BERT & \Circle & \Circle & \Circle & \Circle & Reactive& PC\\
    Paprika \cite{jang2024s}& GPT-4 & \CIRCLE & \Circle & \Circle & \Circle & Reactive& PC\\
    Tianji \cite{tianji2024}& InternLM & \CIRCLE & \Circle & \Circle & \Circle & Reactive & PC\\
  Hua \textit{et. al}~\cite{hua2024assistive}&  GPT 3.5, Atom-7B-Chat  & \CIRCLE & \CIRCLE & \Circle & \Circle & Reactive&PC\\
    % Hua \textit{et. al}~\cite{hua2024assistive}&  Atom-7B-Chat  & \CIRCLE & \CIRCLE & \Circle & \Circle & Reactive&PC, API\\
    SADAS \cite{hua2024sadas}&  ChatGPT  & \CIRCLE & \CIRCLE & \Circle & \Circle & Reactive&HoloLens\\
    OS-1 \cite{xu2024can}& GPT4, Gemini, Llama2  & \Circle & \Circle & \CIRCLE & \CIRCLE & Proactive&Glasses\\
    % OS-1 \cite{xu2024can}& GPT-4 & \Circle & \Circle & \CIRCLE & \CIRCLE & Proactive&Smart Glasses\\
    PRELUDE \cite{gao2024aligning}&  GPT-4  & \Circle & \Circle & \Circle & \CIRCLE & Reactive&PC \\

    \textbf{\workname~}& GPT-4o, Llama-3.1  & \CIRCLE & \CIRCLE & \CIRCLE & \CIRCLE & Proactive&Glasses\\
  \bottomrule
\end{tabular}
\vspace{-5pt}
\end{table}

\subsection{Proactive Conversational Systems}
Reactive conversational systems follow the ``receive and respond'' paradigm, exemplified by writing assistants \cite{gao2024aligning,mysore2023pearl} and coding assistants \cite{englhardt2024exploring}, which generate an answer based on the user’s input, without further interaction.
% take the user's query as input and generate an answer without further interaction.
% A typical example is writing assistants \cite{gao2024aligning,mysore2023pearl}, and coding assistants \cite{englhardt2024exploring}, which take the user's query as input and generate an answer without further interaction.
Proactive conversational systems can initiate and steer conversations through multi-turn interactions with users~\cite{deng2024towards}. 
% [before condense] Proactive conversational systems are those that can take the initiative to steer conversations through multi-turn interactions with users~\cite{deng2024towards}.
OS-1~\cite{xu2024can} utilizes personal daily logs, historical context, and perceived environmental information to proactively engage users, serving as a virtual companion.
% [before condense] OS-1~\cite{xu2024can} utilizes personal daily logs, historical context, and perceived environmental information to proactively initiate and maintain conversations with users, serving as a virtual companion.
% OS-1\cite{xu2024can} leverages the personal daily log, historical context, and perceived environment information to proactively initiate and maintain conversations with the user as a virtual companion.
DrHouse~\cite{yang2024drhouse} is a proactive multi-turn diagnostic system that uses expert medical knowledge and sensor data for multi-turn assessments. 
% [before condense] DrHouse~\cite{yang2024drhouse} is a diagnostic conversational system that proactively determines the next steps and conducts multi-turn diagnoses by leveraging expert medical knowledge and sensor data.
WorkFit \cite{ahire2024workfit} is a proactive voice assistant that detects sedentary behavior in workers and generates voice interventions and health suggestions.
% [before condense] Additionally, recent studies have developed WorkFit \cite{ahire2024workfit}, a proactive voice assistant capable of detecting sedentary behavior in workers and generating voice interventions and health suggestions.
However, existing proactive conversational systems are limited to individual user scenarios.
There remains a gap in research on proactive assistive systems for live social interactions involving conversational partners.

% \textcolor{blue}{third-party participants}.
% [before condense] However, existing proactive conversational systems are designed solely for individual user scenarios. There remains a research gap in studying proactive assistive systems during live social interactions that include third-party participants.

\subsection{LLM Personalization and Acceleration}
\noindent\textbf{LLM Caching}.
Caching solutions have been utilized in LLM reasoning systems to reduce repetitive computations, including caching LLM response and caching intermediate states \cite{bang2023gptcache,li2024scalm,yao2024sirllm,gao2024cost,gim2024prompt}.
% They can be divided into caching LLM response and caching intermediate states.
GPT-cache \cite{bang2023gptcache} and SCALM \cite{li2024scalm} employ semantic cache to store the LLMs responses.
Additionally, numerous studies employ key-value (KV) cache, reusing attention states during LLM response generation, to reduce inference costs \cite{yao2024sirllm,gao2024cost,gim2024prompt}.
CachedAttention \cite{gao2024cost} reuses the KV cache of historical tokens in multi-turn conversations.
Prompt Cache \cite{gim2024prompt} resues the attention states of the overlapped text segments among different prompts.
Unlike general cache designs, \workname~incorporates social factor priors into the cache to enhance accuracy.
% Furemore, many works extend the KV cache into
% also study the issues when applying KV cache in practical scenarios such as multi-turn LLM conversation serving scenarios and multiple prompts scenarios.
% such as CachedAttention \cite{gao2024cost}, Prompt Cache \cite{gim2024prompt}, and SirLLM \cite{yao2024sirllm}, thereby reducing inference costs.
% Other techniques such as KV-cache and CachedAttention [xxx] cache the attention states of the query to saving LLM inference costs.
% PROMPT CACHE [xxx] propose to cache the attention states of prompt that frequently used such as instructions in the prompt.
% However, cache systems xxx.

\noindent\textbf{Streaming and Real-time LLMs}.
Real-time AI assistants have been developed recently, such as Gemini Live \cite{Gemini}.
It supports users to interrupt conversations and assists with daily tasks on mobile phones.
Additionally, some studies explore the real-time speech LLMs \cite{xie2024mini,seide2024speech,liu2024andes}. 
% real-time speech LLMs like Mini-Omni \cite{xie2024mini} and Speech ReaLLM \cite{seide2024speech} have been proposed.
Mini-Omni \cite{xie2024mini} integrates hearing, speaking, and thinking into speech foundation models for real-time conversation. 
% Mini-Omni \cite{xie2024mini} integrates hearing, speaking, and thinking into speech foundation models, enabling real-time conversation capabilities.
Speech ReaLLM \cite{seide2024speech} achieves real-time speech recognition by streaming speech tokens into LLMs for reasoning without waiting for the entire utterance or changing the LLM architecture.
However, these systems focus on general speech recognition and lack the integration of multi-modal social knowledge, limiting their utility in live social interactions.
% However, these systems focus on general speech recognition scenarios without incorporating multi-modal social knowledge into LLM reasoning, and thus cannot assist users during live social interactions.
% conversations with the third party, considering only the user's unilateral goals and inputs.
\workname~ is designed to proactively provide social suggestions during live interactions involving multiple participants.
% \workname~ are designed to proactively provide social suggestions in live social interactions with multi-parties participants.
% provides real-time social assistance in human-to-human conversation scenarios.

%% file: Motivation/Motivation.tex
% \vspace{-.5em}
\section{A Survey on Social Assistance Needs}
\label{user survey}
% \zy{This whole section is about a user study. I suggest we change the title: ``A Study about User Needs for Social Assistance'' or ``A Survey on Social Assistance Needs''}

%\yq{First, we want to understand the need for social assistants during conversations. Therefore, we conduct a survey...}x
To understand the demand for social assistants during interactions, we conduct a survey exploring user experience, preferences, and needs regarding live social interactions.
% It focuses on understanding how individuals perceive social awkwardness and whether a virtual assistant embedded in smart glasses could provide effective assistance.
% Through the study, we identify participants' social experience, as well as their needs, preferences, and potential concerns about the assistive systems during live social interactions.
The results and findings guide the design of our system.
% user experience in social awkwardness, the high demand for social assistance, various user preferences for the assistance technology, and user's concerns about privacy and comfort when using such technology.

% 
% structure: method & three findings (subsection: Finds:...)
\vspace{-.5em}
\subsection{Design of Questionnaire}
% four parts: 1. social experience; 2. the potential need for a social coach; 3. technical requirements; 4. privacy 
% We designed a questionnaire to assess individuals' social stress situations and their attitudes toward virtual assistance in social interactions.
% We designed a questionnaire 
The questionnaire comprises three sections, totaling 14 questions.
%each tailored to explore specific aspects of social interactions and technological interventions.
The questions are summarized as follows:
\begin{itemize}[leftmargin=*]
\item
\textbf{\textit{P1: }} This section is designed to capture participants' social experiences, including their experiences of social awkwardness, awkwardness sources, and attention to nonverbal behaviors during social interactions.
% It includes questions about participants' social awkwardness levels, awkwardness sources, and attention to nonverbal behaviors during social interactions.

\item
\textbf{\textit{P2: }} The second section assesses the needs and preferences for virtual social assistance technologies.
It includes questions about participants' attitudes toward social assistance during live interaction, preferred devices, desired social situations, desired content of suggestion, and assistive information format. It also examines participants' preferred information display methods and tolerance for system latency.
% It includes questions about participants' attitudes toward real-time social suggestions provided by virtual assistants, as well as their preferences for the device through which the system should be deployed.
% Additionally, it investigates the specific social situations where participants would find such assistance helpful, the type of content they would like the virtual assistant to provide, and their preferred format for receiving information.
% Furthermore, we examine how participants prefer the assistant to deliver the information, as well as participants' attitudes and tolerance towards system latency.

\item
\textbf{\textit{P3: }} The final section explores participants' attitudes toward privacy and comfort in the context of virtual social assistance technologies, assessing their willingness to interact with users utilizing such assistants and concerns about potential personal data capture during interactions.

% The final section explores participants' attitudes towards privacy and social comfort in the context of virtual social assistance technologies.
% The questions are posed to assess participants' willingness to engage in social interactions with individuals utilizing virtual assistants, as well as their concerns about privacy, particularly in situations where virtual assistants may capture or process personal data during these interactions.
\end{itemize}
We collect 60 questionnaires in total, and summarize the results and findings as follows.

% \subsection{Findings: Social Experience and Awkwardness}
\vspace{-.5em}
\subsection{Social Experience and Awkwardness}
\begin{figure}
    \centering
    \begin{subfigure}{0.32\columnwidth}
        \centering
        \includegraphics[width=1\textwidth]{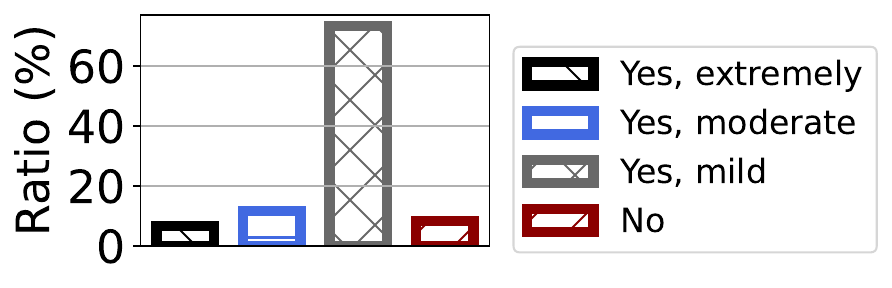}
        \vspace{-2.0em}
        \caption{Do you feel anxious or fearful during social interactions?}  \label{fig:user_study_anxious}
    \end{subfigure}
    \hfill
    \begin{subfigure}{0.32\columnwidth}
        \centering
        \includegraphics[width=0.98\textwidth]{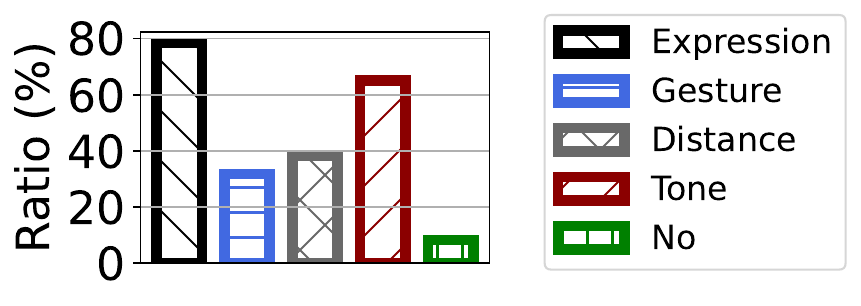}
        \vspace{-1.0em}
        \caption{Do you pay attention to the partner's nonverbal cues? If so, what types?}  \label{fig:user_study_socialcue}
    \end{subfigure}
    \hfill
    \begin{subfigure}{0.32\columnwidth}  
        \centering 
        \includegraphics[width=1\textwidth]{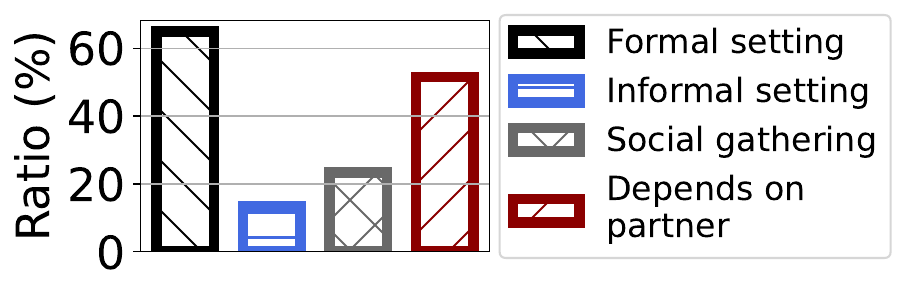}
        \vspace{-2.0em}
        \caption{In which social settings do you feel the most anxious during interactions?}    
        \label{fig:user_study_situation}
    \end{subfigure}
    \vskip\baselineskip\vspace{-1.0em}
    \begin{subfigure}{0.32\columnwidth}  
        \centering 
        \includegraphics[width=1\textwidth]{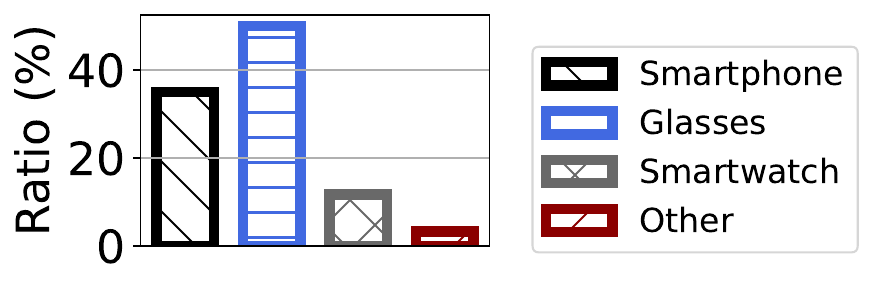}
        \vspace{-2.0em}
        \caption{For a virtual assistant, which device would you prefer to use it on?}    
        \label{fig:user_study_device}
    \end{subfigure}
    \hfill
    \begin{subfigure}{0.32\columnwidth}  
        \centering 
        \includegraphics[width=1\textwidth]{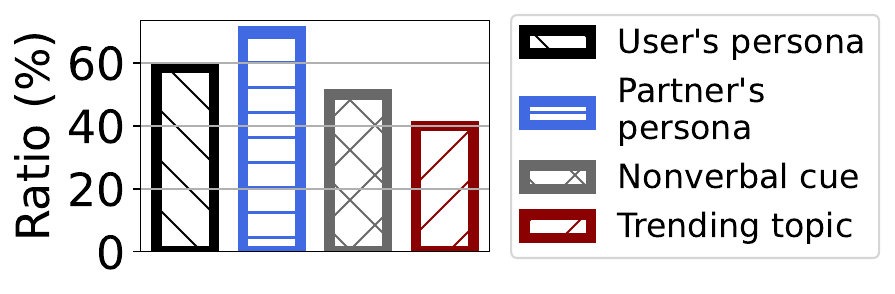}
        \vspace{-2.0em}
        \caption{What kind of social advice would you like a virtual assistant to provide?}    
        \label{fig:user_study_context}
    \end{subfigure}
    \hfill
    \begin{subfigure}{0.32\columnwidth}  
        \centering 
        \includegraphics[width=0.98\textwidth]{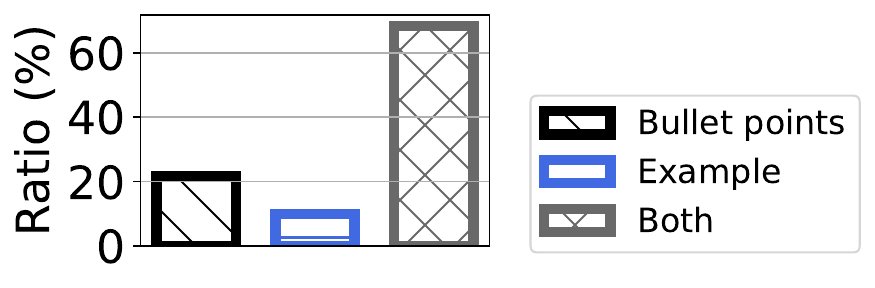}
        \vspace{-0.8em}
        \caption{How would you prefer the virtual assistant to provide help?}    
        \label{fig:user_study_assitanceformat}
    \end{subfigure}
        \vspace{-1.0em}

    \caption{Survey results for social experience and assistance demand.}
    %social awkwardness, device preference, and assistance format.
    \vspace{-1.3em}

\end{figure}

% \zy{Remove ``Findings:'' from all subsections.}
% Q2: E or I; Q3: whether feel stress or not
Among the participants, 18.3\% consider themselves to enjoy interacting with others, while the remaining participants describe themselves as not enjoying it as much.
Besides, only less than 10.0\% of the participants report being completely at ease during social interactions.
As shown in Figure \ref{fig:user_study_anxious}, 91.7\% claim that they experience some level of awkwardness in social situations, indicating that social awkwardness is pretty common in daily life.
\begin{comment}
Among the participants, 46.7\% regard themselves as ambiverted, while 35.0\% of the participants describe themselves as introverted.
Although 18.3\% of the participants describe themselves as extroverted, only a tiny portion, which is less than 10.0\%, among the participants report being completely at ease during social interactions.
As shown in Figure \ref{fig:user_study_anxious}, 91.7\% of participants claim that they experience some level of stress in social situations.
Actually, as xxx report, \textbf{social awkwardness is a common issue in daily life}.
% which situation need assistance
\end{comment}

% Q3: the sources of social awkwardness
The survey results indicate social awkwardness comes from various sources.
% Specifically, 61.7\% report experiencing awkwardness when interacting with workplace superiors, while 41.7\% express nervousness during interactions with professors in academic settings. 
Specifically, more than 60.0\% report experiencing awkwardness when interacting with workplace superiors or professors. 
This trend extends to formal events like meetings.
%, with 35.0\% reporting awkwardness.
Peer interactions contribute as well, with 40.0\% feeling nervous when interacting with colleagues or fellow students, particularly in initial encounters.
Furthermore, 31.7\% report awkwardness when interacting with long-lost acquaintances, and nearly half feel anxious when communicating with unfamiliar relatives.
%([\result 48.33\%])
% Q4: social situation [\result 65\% + 23.33\%]
% The results on social situations further reveal the impact of environmental factors.
Moreover, as Figure \ref{fig:user_study_situation} shows, 65.0\% experience stress in formal settings.
%, and 23.3\% in social gatherings.
Besides, over half also regard conversational partners as a key factor, indicating that personal relationships are vital in shaping social awkwardness.
These results indicate that social awkwardness is most pronounced in situations involving authority figures, formal settings, and unfamiliarity.

% Q5 Interpreting social cues. Do you adjust your communication style based on the other person's facial expressions or tone of voice? If so, what types of nonverbal cues do you commonly use?
Furthermore, the results reveal that nonverbal social behaviors play an important role in social interactions, particularly facial expressions, tone of voice, personal distance, and gestures.
Figure \ref{fig:user_study_socialcue} shows that only 8.3\% overlook nonverbal behaviors while the majority consider them essential during interactions.
Specifically, facial expressions are noted by nearly 80.0\% of the participants, and tone of voice is noted by 65.0\%.
% as key indicators. 
% Tone of voice follows, with 65.0\% recognizing its significance.
Besides, 38.3\% are attentive to personal distance, and 31.7\% regard gestures as supplementary cues.
Despite these nonverbal behaviors’ significance, their indirect nature presents challenges, suggesting a need for support in interpreting nonverbal cues.
% Unlike speech, nonverbal cues require interpretation and can be subject to misunderstanding.
% This complexity in nonverbal cue interpretation suggests a significant need for technological assistance in recognizing and interpreting these subtle yet crucial aspects of social communication.

\begin{comment}
\begin{figure}
    \centering
    \begin{subfigure}{0.32\columnwidth}
        \centering
        \includegraphics[width=1\textwidth]{Motivation/figs/user_study_anxious.pdf}
        \vspace{-2.0em}
        \caption{Do you feel anxious or fearful during social interactions?}  \label{fig:user_study_anxious}
    \end{subfigure}
    \hfill
    \begin{subfigure}{0.32\columnwidth}  
        \centering 
        \includegraphics[width=0.9\textwidth]{Motivation/figs/user_study_device.pdf}
        \vspace{-1.0em}
        \caption{If you were to use such an assistant, which device would you prefer it on? }    
        \label{fig:user_study_device}
    \end{subfigure}
    \hfill
    \begin{subfigure}{0.32\columnwidth}  
        \centering 
        \includegraphics[width=1\textwidth]{Motivation/figs/user_study_assitanceformat.pdf}
        \vspace{-2.0em}
        \caption{How would you prefer the virtual assistant to provide help?}    
        \label{fig:user_study_assitanceformat}
    \end{subfigure}
    \caption{Survey results for social awkwardness, device preference, and assistance format.}
    \vspace{-2.0em}
\end{figure}
\end{comment}

% \subsection{Findings: The Demand and Preference for a Social Assistant}
\vspace{-.5em}
\subsection{The Demand and Preference for a Social Assistant}
% Part II: The Demand for Social Assistant
% Q1: Virtual Assistant Preference: Do you think a virtual assistant that provides real-time feedback and suggestions on social cues during social interactions would be helpful?
% Q2: Preferred device for virtual assistant: If you were to use such an assistant, on which type of device would you prefer to use it?
% Q3: Virtual assistant usage scenarios: In what situations would you want the virtual assistant to help you?
% Q4: What content would you like the virtual assistant to include in its real-time social advice during interactions?
% Q5: Preferred assistance format: How would you like the virtual assistant to provide help?
% Q6: If there were smart glasses that could provide you with social advice during interactions, how would you prefer this social advice to be delivered to you?
% Q1 & Q2  
The questionnaire’s second section reveals that the preference for a virtual social assistant aligns with the social awkwardness experienced by participants.
Notably, 70.0\% believe that a virtual assistant offering instant social suggestions during interactions would be beneficial, indicating a clear demand for social assistance technology.

% Q3: Virtual assistant usage scenarios
Individuals desire assistance during social interactions in certain scenarios.
To be specific, 66.7\% need assistance when feeling uncertain or embarrassed about what to say, 56.7\% when interacting with specific individuals, particularly authority figures, and nearly half when unsure how to respond or initiate the conversation with a goal in mind.
% when they struggle to understand what the other person is saying, and when they have a specific goal but are unsure how to initiate the conversation. 
% Q4: What content would you like the virtual assistant to include
Furthermore, participants also have content preferences for a virtual assistant’s instant suggestions. 
Specifically, participants value both information on conversational partners' and their own interests and backgrounds, with 70.0\% preferring insights into partners’ personas and over 50.0\% interested in their own profiles.
% This suggests that individuals regard effectively grasping conversational partners’ information as essential for successful communication, aligning with findings from studies on social awkwardness.
Additionally, 40.0\% seek updates on trending topics, and half want social cues about nonverbal behaviors.
These results suggest that an effective virtual assistant should offer social assistance with human-like perception.
%, emphasizing the importance of understanding social signals.
% In addition, 40\% express interest in receiving updates on current trending topics, such as news and weather, which can serve as safe, universally acceptable conversation topics.
% Apart from these verbal parts, half of the participants show interest in receiving interpretations of nonverbal behaviors from their conversational partners, highlighting the demand for understanding nonverbal cues in interactions.

% Q5: Preferred assistance format [\result 68.33\% + 21.67\%]
% Q6: how would you prefer this social advice to be delivered to you? [\result 48.33\% + 45\%]
The results in Figure \ref{fig:user_study_device} show that over half of the participants prefer glasses as assistive devices since glasses are convenient and appear natural in conversation.
Furthermore, for information display, 93.3\% prefer text projected in their field of vision.
For assistive information format, as demonstrated in Figure \ref{fig:user_study_assitanceformat}, 68.3\% prefer both summarized bullet points and example sentences, indicating a need for concise and direct suggestions. 
Moreover, instant assistance is preferred with 90.0\% emphasizing instant delivery.
%, and a delay of up to 0.5 seconds is considered acceptable by 70.0\%.
% However, despite this preference for real-time assistance, a delay of up to 0.5 seconds is considered acceptable by 70\% of participants, indicating there is flexibility in developing virtual social assistance systems.
These results suggest a potential demand for employing AR glasses to provide in-suit social assistance, offering instant, easily accessible information without disrupting the conversation flow.

% All these results suggest a desire for real-time, easily accessible information assistance that can be quickly processed without interrupting the natural flow of conversation.

% \textcolor{blue}{Overall, most participants express a strong demand for a virtual social assistant that provides real-time and accessible guidance in social interactions.
% They prioritize ease of use, natural integration through glasses, and concise, instant feedback focused on understanding others.
% This aligns with their desire to reduce social awkwardness and improve social effectiveness, especially in challenging scenarios.}
%The findings suggest that an optimized virtual assistant, with minimal delay and a focus on empathy and nonverbal interpretation, could significantly enhance users’ social confidence and interaction quality.

\subsection{Privacy and User Comfort}
Privacy and user comfort are critical factors in the adoption and acceptance of virtual social assistance technologies. 
Results reveal strong openness to such technologies, with 88.3\% willing to engage with users employing such assistants. 
% The results of the third section reveal a high level of openness to virtual social assistance technology, with 88.34\% of participants expressing willingness to interact with users who employ such assistants during communication. 
However, when confronted with specific privacy concerns, such as image capture during interactions, user comfort levels decrease.
Despite this, 63.3\% are willing to continue conversations.
This indicates that while privacy concerns are present, they do not significantly deter interest and demand for social assistance technologies, highlighting a generally positive reception.
% However, when confronted with specific privacy concerns, such as the potential for image capture during interactions, user comfort levels showed a notable decrease. Despite this, a significant proportion of respondents, 63.33\%, still expressed an overall willingness to continue conversations in such scenarios.  %[\result 18.33\% + 45\%]

% Overall, participants generally exhibit high acceptance and willingness to engage with a virtual assistant.
% Even when privacy issues arise, most users remain open to continued interaction, suggesting that while privacy is a concern, it does not deter overall interest. This highlights a generally positive reception of this technology in social settings.
% \textcolor{blue}{All these results indicate a generally positive reception of this technology in social settings and suggest a promising market.
% }

\subsection{Findings Summary}
We summarize our key findings as follows:
\begin{itemize}[leftmargin=*]
\item Social awkwardness is pretty common in daily life, particularly in interactions with authority figures, formal settings, and unfamiliar situations.
This reveals the potential benefits of virtual social assistance.
%in easing such awkwardness.

\item Nonverbal behaviors like gestures, facial expressions, and personal distance are essential in interactions, as people naturally perceive and focus on these cues during conversations. An effective virtual assistant should therefore provide assistance with human-like perception for nonverbal cues.
% In addition to explicit verbal communication, nonverbal social behaviors， such as gestures, facial expressions, and personal distance，are essential in social interactions, since people naturally perceive and focus on these cues during conversations.
% an effective virtual assistant should offer social assistance with human-like perception.
% Participants also express strong interest in receiving interpretations of their conversational partners' nonverbal behaviors, such as gestures, facial expressions, and personal distance.

\item Participants show strong interest in a virtual social assistant that offers instant guidance to reduce social awkwardness.
They prefer assistance in specific scenarios, certain suggestion content, natural integration via glasses, as well as concise and instant suggestions.
These results indicate a clear demand for a proactive system based on AR glasses to provide effective social assistance during live interactions.
% These results indicate a clear demand for a proactive system based on AR glasses to provide effective social assistance without disrupting the conversation flow.

% Participants are highly interested in a virtual social assistant that provides real-time and accessible guidance in social interactions.
% For social assistance, they demonstrate preferences for certain social scenarios, specific social suggestion content, natural integration through glasses, and concise, instant feedback.
% This aligns with their desire to reduce social awkwardness and improve social effectiveness.
% These results suggest a potential demand for employing AR glasses to provide in-suit social assistance, offering real-time, easily accessible information without disrupting the conversation flow.

%while privacy and comfort are essential for adopting virtual social assistance technologies, it does not significantly hinder the overall willingness of participants to engage with virtual social assistants. 
\end{itemize}
% These key insights can further guide the design of our proactive system for live social interactions based on LLMs and Sensor data.
These findings further motivate the design of our proactive social assistive system for in-situ live interactions based on AR glasses and LLMs.

% \textcolor{red}{Highlights the use of AR glasses here.}\textcolor{blue}{(have emphasized in the second point)}

% \noindent\textbf{\textcolor{red}{Summary. bullets.}}

%% file: System_design/System_design.tex
% \clearpage
\section{System Design}
\subsection{System Overview}
\begin{figure}
  \centering
\includegraphics[width=1\linewidth]{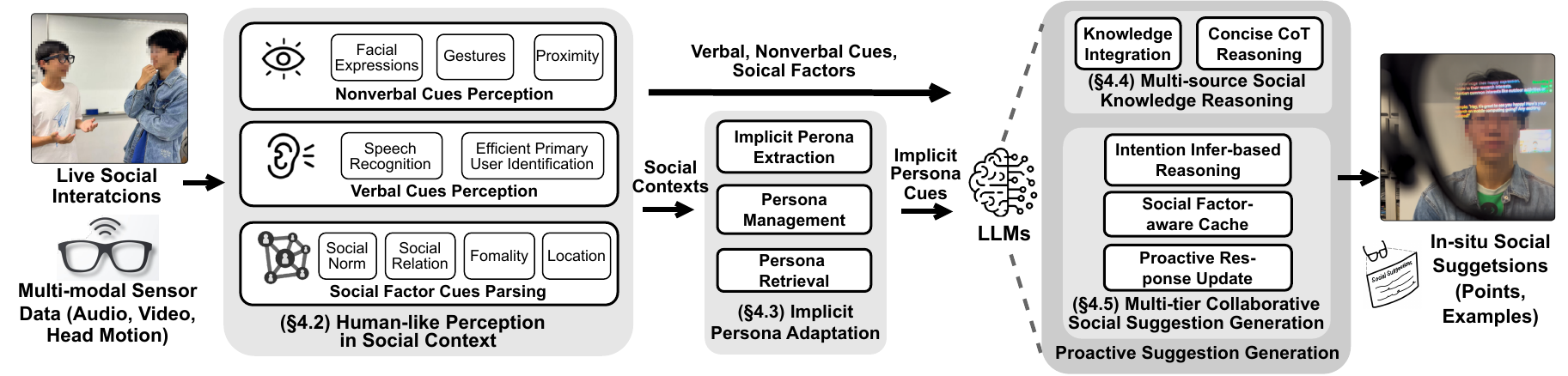}
\vspace{-1.5em}
  \caption{System overview of \workname.
  \workname~leverages the multi-modal sensor data to achieve human-like perception.
    The extracted verbal and nonverbal cues, social factors, and implicit personas are integrated into LLMs, generating in-situ social suggestions with points and examples displayed on the user's AR glasses.
  }
  % \vspace{-.5em}
  \label{fig:system_overview}
  \vspace{-1.5em}
\end{figure}

\workname~is an LLM-based proactive AR social assistive system capable of human-like perception, providing users with in-situ assistance during live social interactions.
% leveraging multi-modal sensors to provide real-time social suggestions during face-to-face live social interactions with third-party participants.
% xxx.(must give highlights: eyewear, multi-modal sensor data, third party participants, live interactions, proactive social assistive system.)
Figure~\ref{fig:system_overview} shows the system overview of \workname.
\workname~first leverages the multi-modal sensor data, including audio, video, and head motion, to achieve human-like perception in social contexts (\S~\ref{multi-modal perception}).
It automatically extracts nonverbal and verbal behaviors, and parses social factor cues.
Meanwhile, \workname~identifies implicit persons from social contexts and performs implicit persona adaptation (\S~\ref{implicit persona adaptation}).
The extracted verbal and nonverbal behaviors, social factors, and implicit persona cues are then integrated into the LLMs for reasoning (\S~\ref{multi-source knowledge integration}). Finally, \workname~employs a multi-tier collaborative reasoning strategy with a social factor-aware cache and intention infer-based reasoning approach to generate in-situ social suggestions (\S~\ref{fast-slow collaborative reasoning}). 
These suggestions are displayed on AR glasses through a proactive response mechanism to assist users in live social interactions without disrupting the natural flow of conversations.

% We use AR glasses as wearable devices instead of smartphones and watches for the following reasons. First, AR technology has been gradually accepted by people in recent years, as evidenced by AR-based applications like captioning~\cite{guo2023sign}. Second, AR glasses provide a non-distracting solution, ensuring that humans can maintain eye contact with others during social interactions without disrupting the natural flow of conversations. Finally, our user survey confirms this: most participants believe that glasses are the most appealing hardware to embed the social assistive system during in-situ live interactions, outperforming smartphones and smartwatches.

We chose AR glasses for social assistance over devices like smartphones or smartwatches for three main reasons. First, AR glasses for daily wear are increasingly accepted, as seen in applications like captioning and translation~\cite{wired_xrai_glass, guo2023sign, evenrealities_g1, inmo_air2}. Second, AR glasses offer a non-distracting solution, allowing users to maintain eye contact during social interactions without disrupting the natural flow of conversation. Finally, our survey indicates that most participants favor glasses as the ideal hardware for embedding a social assistive system in live interactions over other devices.

\subsection{Human-like Perception in Social Context}
\label{multi-modal perception}
% Motivation: Why we need this module in our system. Speaking copilot is totally different from writing copilot/single-person chatbot.
Existing studies on social assistive systems focus solely on single-user human-to-computer interactions and follow a reactive paradigm, conducting either question-answering~\cite{jang2024s,tianji2024} or remediating cultural violations~\cite{hua2024assistive,zhan2024let,zhan2024renovi}.
However, live social interactions involve multi-modal cues such as nonverbal behaviors and social factors, posing challenges to existing text-only LLMs in providing comprehensive social suggestions. \workname~employs a human-like perception approach that can leverage the multi-modal sensor data to extract social cues during live social interactions.

% proactively perceive comprehensive cues during live social interactions.

% \workname~proactively perceives other individuals' nonverbal behaviors during live face-to-face social interactions involving additional conversational partners by using multi-modal sensors on AR glasses.

% \workname~proactively perceives nonverbal behaviors during live face-to-face social interactions involving additional conversational partners by using multi-modal sensors and incorporates them into the social suggestions.
% In this section, we introduce the xxx.
% However, these systems focus solely on single-user human-to-computer interactions, considering only the user's unilateral goals and inputs.
% \workname~takes a further step by providing social assistance during live face-to-face interactions with third-party participants.

% such as OS-1 [xxx], integrate multimodal information like images and audio to assist daily conversations between the user and the system. However, xxx
% [existing studies do not contain multi-party scenarios.]
% However, these systems do not leverage this multimodal information to reason about social-related clues, which are crucial in social conversation contexts [xxx].
% In addition, existing systems [xxx] are designed for user-machine communication applications, such as writing assistants [xxx] or emotion support assistants [xxx], rather than for social communication scenarios that involve multi-party conversations.
% In this section, we introduce xxx.

\subsubsection{Nonverbal Cues Perception}
\label{Nonverbal Behaviors Perception}
% Existing studies lack of xxx.
% why we need this in our system?
% what info we perceive
% how?
% how to use it (see xxx.)
% 1. importance.
% 2. Example, what does example nonverbal cue means?
% 3. Challenges remains in LLMs can not directly understand these multi-modal nonverbal data, but use specialized small models face challenges in the 
% what nonverbal we use?
% if low quality, how to manage?
Nonverbal behaviors play a crucial role in face-to-face social interactions \cite{duncan1969nonverbal}. 
For example, facial expressions like confusion and frowning can indicate a person's emotional state during face-to-face social conversations \cite{frith2009role}.
Additionally, gestures can reveal a person’s implicit perspectives, such as their understanding, intentions, or agreement, during social interaction.

% \workname~ proactively perceives the nonverbal behaviors of other individuals and leverages these implicit cues to adjust the social strategy and help user. 
\workname~proactively perceives the nonverbal behaviors of the conversational partners and leverages these implicit cues to adjust social strategies and assist the user.
However, nonverbal behaviors such as facial expressions, gestures, and physical proximity are captured by multi-modal sensors. Directly offloading the raw multi-modal data to the cloud server incurs significant bandwidth usage, high latency, and raises privacy concerns.
% posing challenges for LLMs to understand and reasoning directly.
To address these challenges, \workname~employs multiple lightweight yet specialized small models on AR glasses to efficiently process raw data locally.
Specifically, we first employ MediaPipe Holistic \cite{pipe} in \workname~to generate human poses, including facial mesh and hand poses.
These facial mesh and hand poses are then further processed by different specialized models to generate nonverbal cues (\S~\ref{implementation}).
% Next, the facial mesh and hand poses are further processed by different specialized models to generate nonverbal cues
% For details please see \S~\ref{implementation}.
Finally, these nonverbal cues are incorporated into the LLMs to generate nonverbal cues-aware social suggestions (\S~\ref{multi-source knowledge integration}).
% For more details, see \S~\ref{multi-source knowledge integration}.
Table~\ref{details_nonverbal} shows the details of the nonverbal cues detected in \workname, including facial expressions, gestures, and physical proximity~\cite{ma2024multimodal}.
We selected these nonverbal cues based on feedback from our user survey in \S~\ref{user survey} and because existing studies indicate that they are the most representative forms of nonverbal communication during face-to-face social interactions~\cite{duncan1969nonverbal}.

\subsubsection{Efficient Primary User Identification}
\label{Efficient Primary User Identification}
Since \workname~focuses on live face-to-face social interactions with conversational partners, it requires efficient and robust identification of the primary user and other participants.
Voice fingerprinting \cite{garcia2003biomet} can be used for speaker identification, but it introduces additional overhead from registration and raises security concerns, such as voice synthesis and replay attacks \cite{li2020vocalprint}. 
This is evidenced by Microsoft's recent closure of its speaker recognition service \cite{microsoft}.
Volume-based solutions \cite{chen2024enabling} utilize low-frequency energy to differentiate the primary user's speech from that of nearby individuals, but their robustness is limited by environmental noise and variations in the user's speaking volume.
To address these challenges, \workname~leverages a lightweight primary user identification approach leveraging the vibration signals on the smart glasses as indicators.

% % it requires the identification of the primary user and other parties.
% Voice fingerprint \cite{garcia2003biomet} employs fingerprint technology to identify the speech of the registered user and nearby individuals. However, the registration process introduces additional overhead and security concerns, such as voice synthesis and replay attacks \cite{li2020vocalprint}.
% % Other limitations of fingerprint, e.g., robustness to the environment.
% EarVoice \cite{chen2024enabling} utilizes low-frequency energy as an indicator to distinguish between the primary user's speech and that of nearby individuals.
% To address these challenges, \workname~leverages a lightweight primary user identification approach leveraging the vibration signals on the smart glasses as indicators.

% \noindent\textbf{Measurements.}
We first conduct real-world measurements where the primary user wears smart glasses and engages in conversations with different partners.
The smart glasses record the audio and vibration signals simultaneously.
Figure~\ref{fig:primary_user_detection} shows the waveform of the audio and vibration signals on the smart glasses during live social interactions. The primary user speaks during the first 6 seconds, while the conversational partner speaks during the last 6 seconds.
Compared to the audio, the amplitude of the vibration exhibits a clear difference between the primary user's speaking period and the partner's speaking period.
Therefore, we leverage the signal energy vibration signals as an indicator to detect the primary user.
Specifically, we calculate the vibration signal's energy within the 3 $\sim$ 10 Hz range and use it as the indicator. Energies exceeding a certain threshold are detected as the primary user. We employ a grid search to determine the optimal threshold.
The sample rate of the vibration signal in \workname~is set to 466 Hz, which is significantly lower than the audio sample rate, thereby reducing bandwidth usage. Additionally, \workname~transmits the vibration signal from the glasses to the server at regular intervals of 300 ms and sets the threshold for primary user detection at 1.1 on the server.
Details on the threshold search and the overall detection performance of our approach compared to audio-based solutions can be found in \S~\ref{performance_primary_user_detection}.

\subsubsection{Social Factor Cues Parsing}
\label{Social Factor Cues Parsing}
% Existing studies, such as OS-1 [xxx], integrate multimodal information like images and audio to assist daily conversations between the user and the system. However, these systems do not leverage this multimodal information to reason about social-related clues, which are crucial in social conversation contexts [xxx]. 
% Research in social science shows that social factors play a vital role in communication \cite{capurucco2009building}.
Existing studies show that social factors play a vital role in social communication \cite{capurucco2009building}.
% Different acceptable and unacceptable behaviors under different social factors.
Social behaviors and speech content considered acceptable or unacceptable can vary significantly depending on different social factors such as social relation and formality \cite{hovy2021importance}.
% For example, when making a request, the tone and content of our speech should vary significantly depending on whether we are addressing a familiar person, like a partner, or a superior, such as a professor or manager.
For example, when making a request, the tone and content of our speech should vary significantly depending on whether we are addressing a familiar person or a superior, such as a professor or manager.
Similarly, social norms differ between formal and informal occasions. 
Therefore, it is essential to incorporate these social factor cues into social suggestion generation strategies.

\begin{figure}[t]  
    \centering  
    \begin{minipage}{0.58\textwidth}  
        \centering  
        \begin{subfigure}{0.485\textwidth}  
            \centering  
            \includegraphics[width=\textwidth]{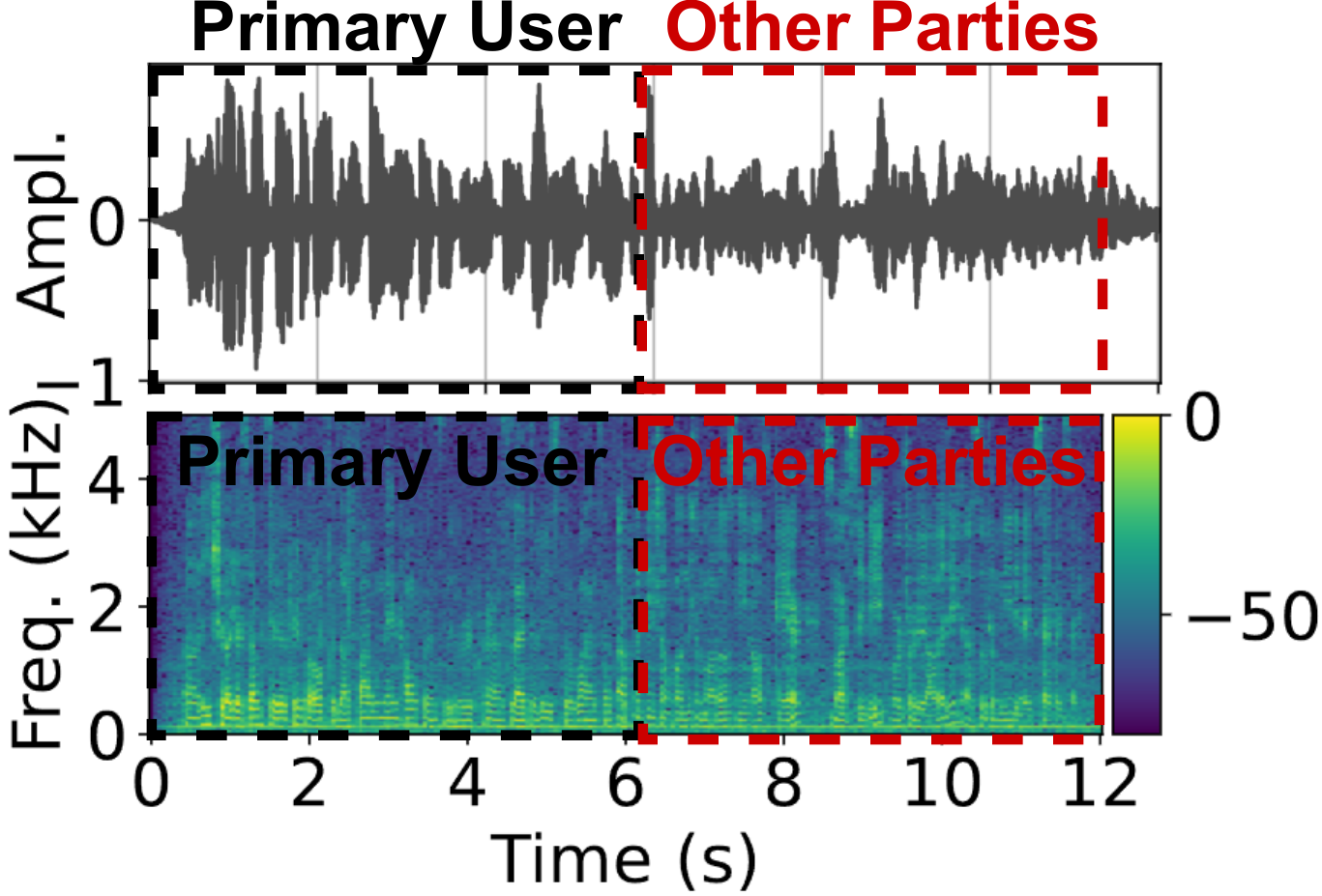}  
            \vspace{-1.7em}
            \caption{Audio signal as indicator.}  
            \label{fig:audio_waveform_timefreq}  
        \end{subfigure}  
        \hfill  
        \begin{subfigure}{0.495\textwidth}  
            \centering  
            \includegraphics[width=1.02\textwidth]{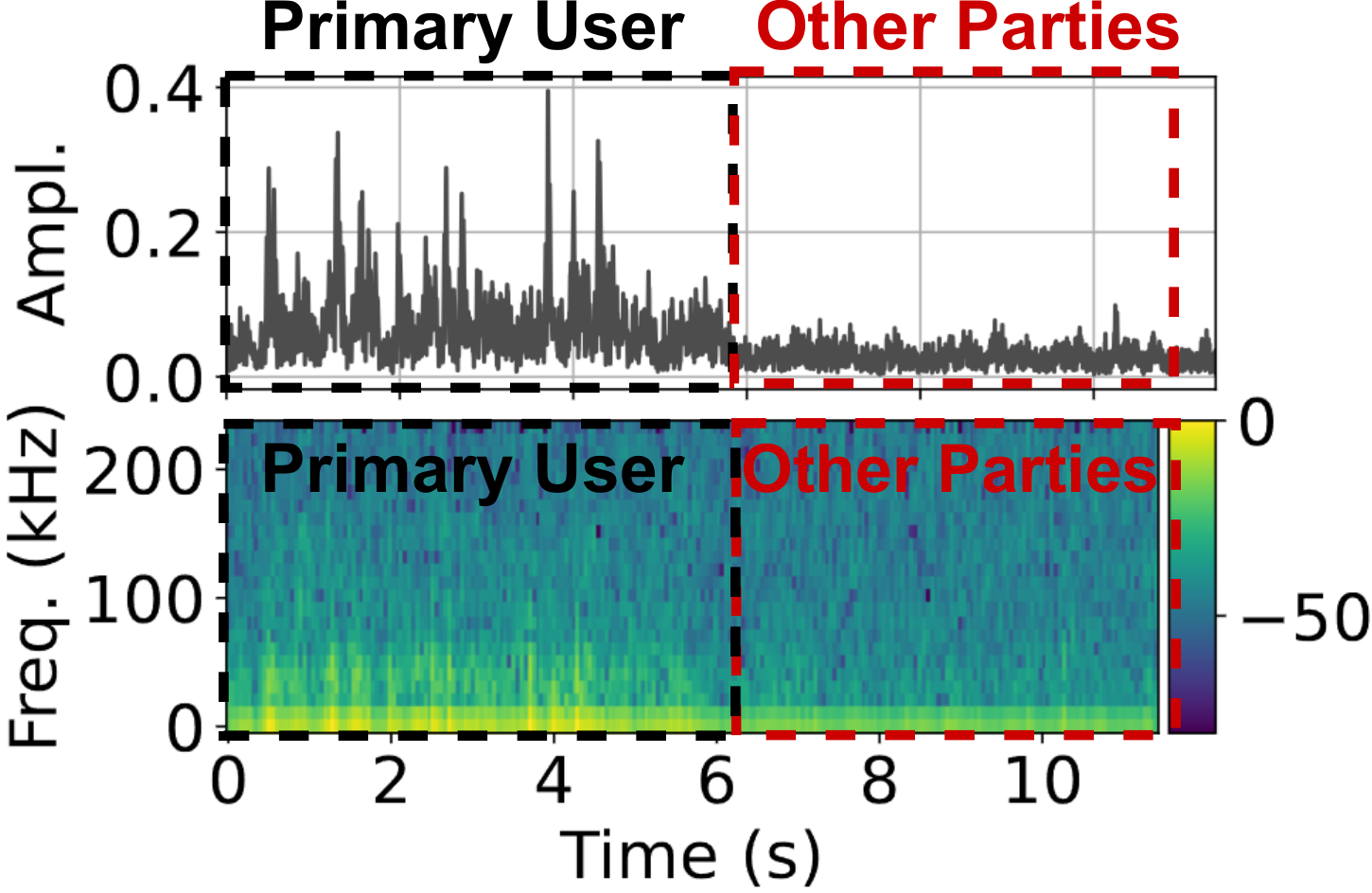}
             \vspace{-1.7em}
            \caption{Vibration signal as indicator.}  
            \label{fig:user_study_doctor_Q3}  
        \end{subfigure} 
        \vspace{-1.2em}
        \caption{\workname's primary user detection.}  
    \label{fig:primary_user_detection}  
    \end{minipage}%  
    \hfill  
    \begin{minipage}{0.4\textwidth} 
    % \footnotesize
    \small
        \centering  
        \vspace{-2em}
        \captionof{table}{Details of Social Factors in \workname.}  
        \vspace{-1.2em}  
        \renewcommand{\arraystretch}{0.9} % 
        \resizebox{\textwidth}{!}{%  
            \begin{tabular}{p{0.32\textwidth}|p{0.68\textwidth}}  
                \toprule  
                \textbf{Social Factors} & \textbf{Sub-Categories} \\  
                \midrule  
                Social Norm & Greeting, Request, Apology,\newline  Persuasion \\  
                Social Relation & Peer-Peer, Elder-Junior,  Mentor-\newline Mentee, Student-Professor \\  
                Formality & Informal, Formal \\  
                Location & Office, Open Area, Restaurant, \newline Conference Break \\  
                \bottomrule  
            \end{tabular}  
        }  
        \label{detatils_social_factors}  
    \end{minipage} 
    \vspace{-1.5em}
\end{figure}

\workname~leverages the social contexts to parse social factor cues. It supports two modes of social factor perception: reactive and proactive. 
In reactive perception mode, the social contexts are instructions provided by the user, describing their social goals, such as: ``I am going to a social communication with a senior professor during a conference break, and my goal is to introduce my research work and establish a social connection with him.''
\workname~utlizes LLMs with dedicated prompts to parse social factors from the user's instructions before initiating social interactions. 
% \workname~first uses LLMs with dedicated prompts to parse social factors from the user's instructions.
If the user does not actively provide descriptions of social factors, \workname~will operate in proactive mode to parse social factors.
In such mode, the social contexts are the captured images with social environment information.
Specifically, \workname~pre-stores the images of various locations such as conferences, meeting rooms, and restaurants.
\workname~leverages the camera on the glasses to recognize the current location by mapping it with the pre-stored images, thereby generating location-based social factors. 
% In proactive mode, without user instructions, \workname~utilizes the camera on the glasses to recognize the current location by matching it with pre-stored images. This enables the generation of location-based social norms.
The social factors identified through either reactive or proactive modes will be used as a knowledge source and integrated into the LLMs for generating social suggestions (\S~\ref{knowledge integration} ). 
Table~\ref{detatils_social_factors} shows the social factors utilized in \workname, including social norm, social relation, formality, and location \cite{zhan2023socialdial}.

% Existing studies lack of xxx.
% why we need this in our system?
% what info we perceive
% how?
% how to use it (see xxx.)

\subsection{Implicit Persona Adaptation}
\label{implicit persona adaptation}
% \textcolor{red}{Need an example, why we need personalization.}
% Every individual has unique backgrounds, experiences, and personal interests, which can result in diverse behaviors during social interactions [xxx]. 
Every individual has unique backgrounds, life experiences, and personal interests, which are abstracted into personas~\cite{ashton2022individual}.
A social topic that connects the personal interests and backgrounds of both parties can enhance the engagement of both parties.
% \textcolor{blue}{An ideal social assistive system should proactively identify implicit user interests and naturally generate social suggestions that adapt to the user's implicit persona.}
An ideal social assistive system should proactively identify the implicit personas of both parties and incorporate these personas into the strategies for social suggestion generation.
% Every individual has unique thoughts, feelings, and behaviors. 
% Every individual has unique backgrounds, experiences, and personal interests, which can result in diverse behaviors during social interactions [xxx]. 
However, \textit{natural social conversations often lack explicit queries to initiate the knowledge retrieval of personal historical databases, posing challenges for system personalization.}
% Generating social suggestions that align with the user's implicit persona during live conversations is challenging.
% \workname~employs a xxx approach to integrate the implicit personas into the process of social suggestions generation.
\workname~employs an implicit persona adaptation approach to generate
customized social suggestions, enhancing the engagement of both parties.

\subsubsection{Implicit Persona Extraction}
\label{Implicit Persona Extraction}
Existing personal assistant systems employ the user's explicit queries to retrieve historical data and provide personalized responses \cite{gao2024aligning,
chu2024towards,
yang2024drhouse}.
However, systems like writing assistants \cite{gao2024aligning}, emotional support assistants \cite{chu2024towards}, and medical assistants \cite{yang2024drhouse} primarily function in a question-answering manner, relying on explicit queries to initiate the retrieval of the personal historical database. 
These explicit queries allow them to utilize the standard RAG techniques \cite{lewis2020retrieval} to retrieve responses with high semantic similarity.
However, natural social conversations lack explicit queries to initiate the retrieval of personal historical data, posing challenges in generating social suggestions that incorporate implicit personas.

% However, during live social conversations, the system cannot initiate the retrieval of personal historical data in the absence of explicit queries, posing challenges in generating implicit persona-aware social suggestions.

% To address this challenge, \workname~employs a xxx approach, which can extract implicit personas cues during historical conversations in advance and integrate them into the process of social suggestion generation.
% To address this challenge, \workname~ employs an additional LLM to extract the personas from historical conversations in advance and maintains an implicit persona cues memory.
% To address this challenge, \workname~ employs an additional LLM to extract the implicit personas of both parties from historical conversations in advance, maintaining a memory module named the \textit{persona cues memory}.
To address this challenge, \workname~ employs an additional LLM to extract the implicit personas of both parties from historical conversations in advance, maintaining a \textit{persona database}.
% \textcolor{red}{Figure~\ref{fig:persona_extraction} shows the prompt used in the LLM for persona cues extraction.}
% Figure~\ref{fig:persona_extraction} shows the construction of the persona database.
% Figure~\ref{fig:persona_extraction} shows how the persona database is constructed, updated, and used in \workname.
Figure~\ref{fig:persona_extraction} shows the pipeline of the implicit persona adaptation in \workname.
Specifically, the persona extraction occurs during the post-interaction phase, where an LLM extracts the implicit persona cues from the social conversations.
These persona cues reflect the personal interests, experiences, and backgrounds of both parties.
The persona database is organized according to individual identities, including those of the user and various conversational partners.
Note that \workname~will not engage in any privacy-infringing activities, such as actively crawling the social network data of other individuals based on facial recognition.

\subsubsection{Persona Management}
% Since live experiences and personal interests evolve over time, \workname~ employs a persona management strategy to adapt to these dynamically changing personas.
Since live experiences and personal interests evolve over time, \workname~employs a persona management strategy to adapt to these emerging personas. 
Specifically, for new conversational partners, their persona cues will be directly registered in the persona database. For the user and previously met partners, \workname~first utilizes LLMs to determine if any contradictory or similar cues already exist within the persona database for that particular identity.
% Specifically, for the new conversational partner, their persona cues will be directly resigtered in the persona database.
% For the user and the already meeted partner, \workname~initially utilizes LLMs to determine if any contradictory or similar cues already exist within the \textit{persona database} for that particular identity.
If no such cases are found, the incoming persona cues are registered and stored in the memory. 
If the incoming persona cues are semantically similar to existing ones, the two sets of cues are merged. 
% \yq{How does the semantic match work?}
Conversely, if the incoming persona cues are contradictory, the historical cues are removed and replaced by the new incoming ones.

\subsubsection{Persona Retrieval}
During live social interactions, \workname~first performs persona retrieval using face ID matching to determine whether the conversational partners are in the database. 
This facial recognition is executed locally on the glasses to protect privacy. 
If the partners are found, the personas of both the user and the partner are loaded and integrated into LLMs as a knowledge source (\S\ref{multi-source knowledge integration}). Otherwise, only the user's persona will be used.
These implicit persona cues can help LLMs identify shared interests or experiences and generate customized social suggestions, thereby enhancing the engagement of both parties.

\begin{figure}
\begin{minipage}[b]{0.49\columnwidth} % Use [b] for bottom alignment
     \centering
\includegraphics[width=1\textwidth]{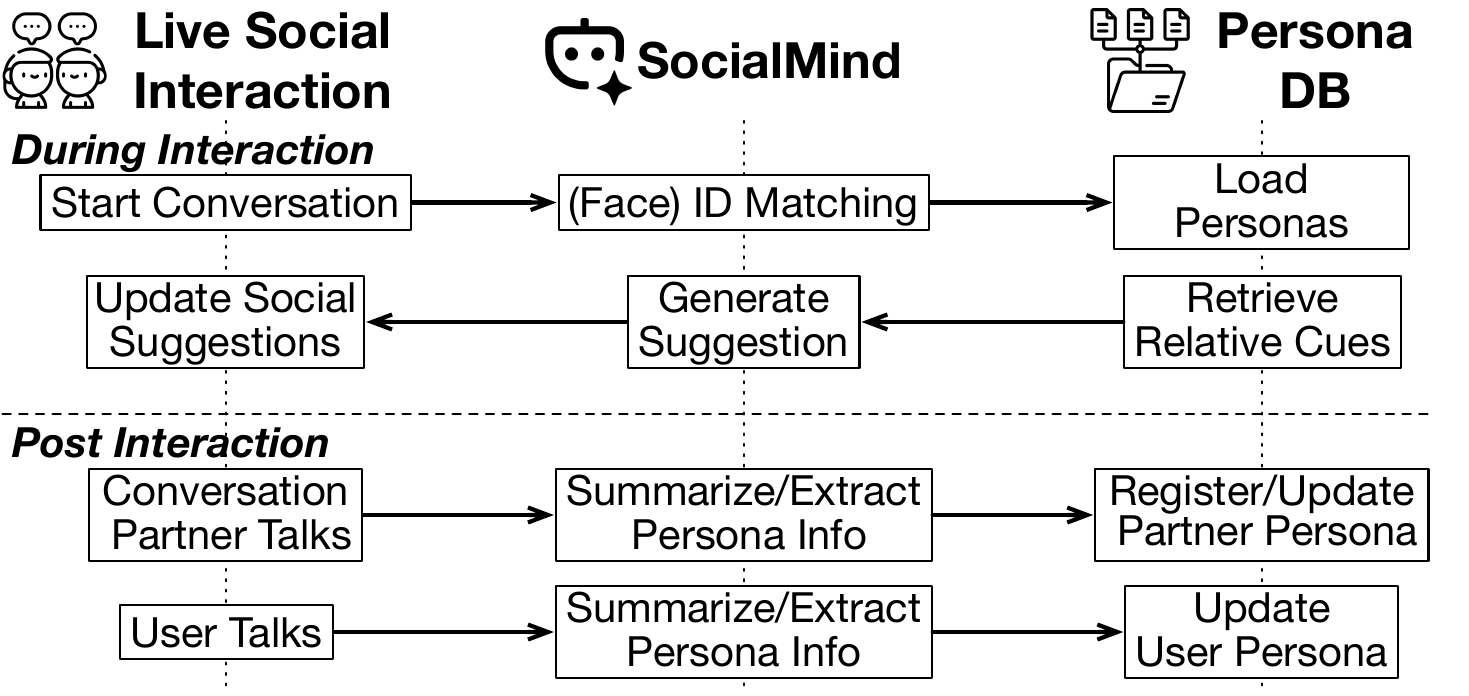}
% \vspace{-1.5em}
  \caption{Implicit persona adaptation in \workname.}
  % \vspace{-.5em}
  \label{fig:persona_extraction}
\end{minipage}
\hfill
  \begin{minipage}[b]{0.49\columnwidth} % Use [b] for bottom alignment
     \centering
\includegraphics[width=1\textwidth]{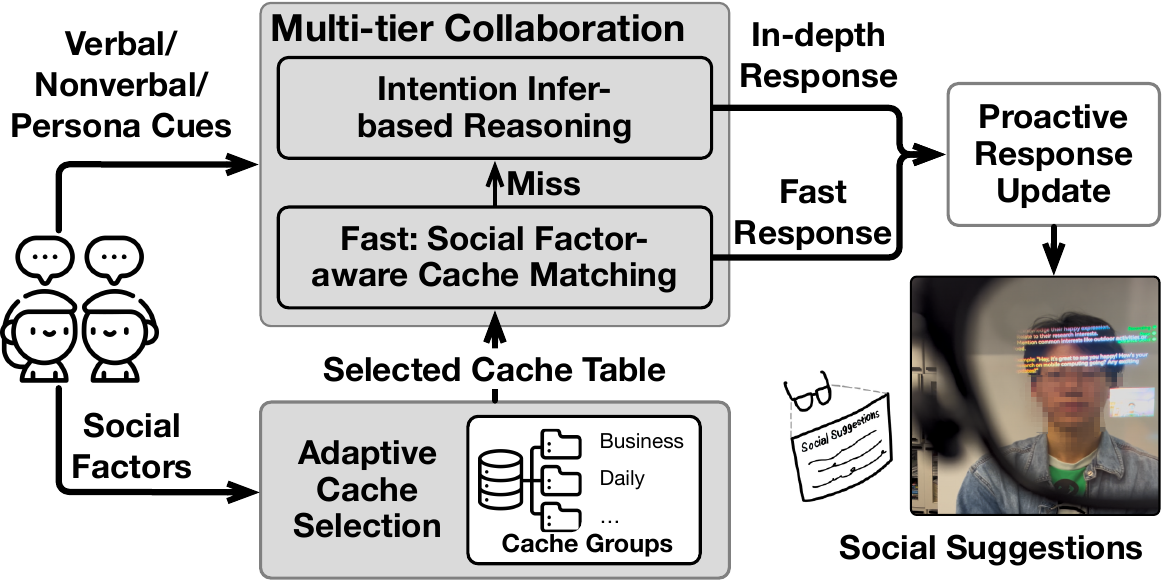}
% \vspace{-1.5em}
  \caption{
   Multi-tier collaborative social suggestion generation.
  }
  % \vspace{-.5em}
  \label{fig:fast-slow}
\end{minipage}
\vspace{-1em}
\end{figure}

\subsection{Multi-source Social Knowledge Reasoning}
\label{multi-source knowledge integration}

Existing social assistive systems primarily focus on text-based question answering and remediating social violations. \workname~leverages
multi-modal sensors to gather multi-source knowledge during live social interactions. 
This section introduces the knowledge integration in \workname, which guides the LLMs to understand and utilize this knowledge in live social interactions, ensuring improved instruction-following performance and enhanced user Quality of Experience (QoE).

% Existing social assistive systems primarily focus on text-based question answering and social violations remediating. 
% \workname~proactively perceives multi-source knowledge during in-situ social interactions. 
% This section introduces the knowledge integration in \workname, which guides the LLMs to understand and leverage this knowledge during live social interactions, ensuring improved LLM instruction-following performance and user QoE.

% challenges 1: multi-source, instruction-following, user survey-> lengthy

% Existing studies on LLM-based social assistants focus on text-only social-related question-answering \cite{jang2024s,
% tianji2024} or remediating social violations in the user's text input \cite{
% hua2024assistive,
% zhan2024let}. \workname~proactively perceives multi-modal information during live social interactions and incorporates multi-source social knowledge into its social suggestion generation strategy. This section introduces the knowledge sources in \workname~and explains how \workname~integrates this knowledge for LLM-based social suggestion generation.

\subsubsection{Knowledge Source}
The knowledge source in \workname~contains 
nonverbal cues (\S~\ref{Nonverbal Behaviors Perception}), the context of live social conversations (\S~\ref{Efficient Primary User Identification}), social factors (\S~\ref{Social Factor Cues Parsing}), and implicit persona cues from both parties (\S~\ref{Implicit Persona Extraction}).
% Details on obtaining this knowledge have been provided in previous sections.
\workname~also integrates external tools that provide the latest weather updates and trending social news, which are valuable sources for conversation topics. Since this information is updated daily and does not need to be retrieved during online interactions, it is pre-retrieved and incorporated into \workname's knowledge source.

\subsubsection{Knowledge Integration}
\label{knowledge integration}

This subsection introduces the prompt used in \workname, which enables LLMs to integrate multi-source and multi-modal information for social suggestion generation.
Specifically, the prompt in \workname~consists of two parts: $prompt=prompt_{static}+prompt_{runtime}$,  where $prompt_{static}$ represents the static portion and $prompt_{runtime}$ represents the runtime changing portion during live social interactions.

\noindent\textbf{Static Prompt}.
This part of the prompt remains fixed and does not update any information during social interactions. It contains the overall instructions, task instructions, prior knowledge of nonverbal cues and their usage guidelines, as well as several few-shot demonstrations.
Specifically, the overall instructions, together with the few-shot demonstrations, activate the LLM's capability to generate social suggestions.
The task instructions enumerate various rules and requirements to enhance the LLM's ability to generate high-quality social suggestions.
% Details can be seen in \S~\ref{Intention Infer-based Suggestion Generation} and \S~\ref{CoT with Concise Response}.
Additionally, we integrate literature and guidelines \cite{duncan1969nonverbal} on utilizing nonverbal cues during live social interactions as prior knowledge within the prompt.
This includes categories of typical nonverbal cues and their corresponding coping strategies.
For example, in the case of physical proximity, personal distance (1.5 to 4 feet) is common among family members or close friends and signifies the intimacy of the relationship \cite{argyle1965eye}.
This prior knowledge helps bridge the gap between the embedded knowledge of LLMs and the specific expertise required for effective nonverbal communication.

\noindent\textbf{Runtime Prompt}.
This part of prompt dynamic changes during live social interactions.
\workname~receives the context of live conversations, multi-modal nonverbal cues, implicit persona cues of both parties and parsed social factors, all of which are integrated into the runtime prompt. This in-situ perceived knowledge enables \workname~to dynamically adjust its social suggestions during live social interactions.

\subsubsection{Concise CoT Reasoning}
\label{CoT with Concise Response}

% The aforementioned subsections introduce the information used in \workname.
The aforementioned subsections introduce the knowledge source used in \workname.
However, integrating this information into LLMs and generating social suggestions that enhance the user's QoE remain several challenges.
First, the user survey in \S~\ref{user survey} shows that more than 67\% of participants prefer social suggestions presented as summarized key points in bullet form, followed by a sample sentence.
However, lengthy and redundant social suggestions may exceed the average human reading speed of approximately 200 words per minute and may not fully display on the glasses' screen \cite{brysbaert2019many}. Additionally, the output format of social suggestions should be specifically designed for readability, comfort, and quick comprehension.
Second, the abundance of information and instructions makes it challenging for LLMs to accurately follow instructions and generate appropriate social suggestions.

To address these challenges, we employ a concise Chain-of-Thought (CoT)~\cite{wei2022chain} reasoning strategy for social suggestion generation.
% \textcolor{red}{Fast->how; slow->how?}
Specifically, we first add the instruction ``Let’s think step by step'' into the prompt.
The CoT reasoning strategy enhances LLMs' complex task reasoning and instruction-following capabilities \cite{wei2022chain}.
% \textcolor{blue}{For example, xxx.}
Next, we set constraints on the generation length of the social suggestions by including ``Limit your total response to $N$ words or less'' in the prompt.
Based on our measurement experiments, we set $N$ to 70, as this length is optimal for full display on the eye screen.
Finally, according to the user survey, the final display format of the suggestions on the glasses includes summarized suggestions in bullet points, followed by a sample sentence.
% we conduct a survey to determine the preferred types and formats of social suggestions. 
% The final display format includes bullet points with suggestions followed by a sentence example. 
This format allows users to choose their preference, whether referring to the summarized bullets or reading the example.
Figure~\ref{fig:prompt_overall} shows the complete prompt used in \workname.

\subsection{Multi-tier Collaborative Social Suggestion Generation}
\label{fast-slow collaborative reasoning}

% Fast suggestion
% - Input (use what types of information).
% - Output (Response format).
% - Pipeline (LLMs? matching table?).

% Slow suggestion
% - Input.
% - Output.
% - Pipeline.

Social assistance during user in-situ interactions with other parties requires providing instant social suggestions, enabling the user to refer to them and talk with others without disrupting the natural flow of the conversation.
To address this challenge, \workname~employs a multi-tier collaborative suggestion generation approach, as shown in Figure~\ref{fig:fast-slow}.
It includes a social factor-aware cache and an intention infer-based reasoning strategy, to provide instant social suggestions.
Additionally, \workname~employs a proactive response update mechanism to control the refresh of social suggestions displayed on AR glasses.

\subsubsection{Social Factor-aware Cache}
Cache has been widely utilized in existing studies to avoid redundant computations \cite{guo2018foggycache,xu2018deepcache,drolia2017cachier} and to reduce serving costs of LLMs \cite{gim2024prompt,bang2023gptcache}.
However, cache-based conversational systems rely on semantic retrieval mechanisms, generating semantically similar responses but often struggling with logical consistency \cite{wu2018response}. 
It can be more challenging in the context of live social assistance since conversational norms vary with social factors \cite{zhan2023socialdial}.
Even for the same utterance, assistive systems should offer different social suggestions based on varying social factors and non-verbal cues, posing challenges to the robustness of cache.
% \textcolor{blue}{For example, xxx.}
% It poses challenges to the robustness of caching systems.
To address these challenges, \workname~leverages the social factor priors to construct and manage the cache.
% employs a social factor-aware cache for fast suggestions generation.
% challenges of cache in generalization and database size
% social factor definition
% our design
% how to initiate
% how to trigger
% how to update

\noindent\textbf{Cache Initialization}.
No existing dataset contains extensive daily conversations paired with social suggestions.
To address the challenges of data scarcity, we leverage the LLM agent for social interaction simulations.
Existing studies have validated the effectiveness of using LLMs for role-play in society, such as negotiations \cite{bianchi2024well}.
% We extend these simulations to the context of live social assistance scenarios. 
% We conduct simulated social interactions between two LLM agents under various social factors, including a user agent and a conversational partner agent.
% During the development phase, we conduct simulated social interactions between two LLM agents under various social factors, including a user agent and an opponent agent.
We extend these simulations to live social scenarios.
Specifically, we set up two LLM agents, including a user agent and a conversational partner agent, to engage in social interactions under various social factors (see \S~\ref{exp_settings}). 
% Details of the settings can be found in \S~\ref{exp_settings}.
The simulated conversations are used to initialize the cache, with the conversational partner's utterance as the key and the corresponding social suggestions generated by \workname~as the values.
% Each utterance in the cache also includes the partner's nonverbal cues.

% We select the conversational partner's utterance and the corresponding social suggestions generated by \workname~ to initialize the cache, where the key is the conversational partner's utterance and value is the corresponding social suggestions.
% We initialize the cache using one hundred simulated social conversations between the user and the conversational partner, along with the corresponding social suggestions provided by \workname.
% Details of the settings can be found in \S~\ref{exp_settings}. 
% We initialize the cache using the generated one hundred simulated social conversations between the user and the conversational partner, along with the corresponding social suggestions provided by \workname.

\noindent\textbf{Cache Groups with Social Factor Priors}.
% According to the definitions of social factors [xxx], we split the .
% Each cache contains xxx.
% \textcolor{blue}{The storage of the off-the-shelf smart glasses has xxx storage [xxx].}
% \workname~stores these simulated social conversations on the glasses side.
Social interactions can be classified according to the social factors.
\workname~employs social factor priors manage the caches.
Specifically, all conversations are grouped into subsets based on social factors such as social norm, social relations, formality, and location. 
Each subset is indexed by these social factors and defined as the social factor-aware cache.
% All subsets together are called cache groups.
% Nonverbal cues priors are used to enhance the cache's ability to remember social scenarios.
% Each dialogue in the cache also includes the opponent's nonverbal cues.

% \textcolor{blue}{Additionally, each dialogue in the cache also includes nonverbal cues.}
% \textcolor{red}{Social factor priors are in cache-wise and nonverbal priors are in sample-wise}.

\noindent\textbf{Adaptive Cache Selection}.
% During in-situ use of \workname, the parsed social factor cues (\S~\ref{Social Factor Cues Parsing}) are sent to a retriever to select the cache from cache groups. If the parsed cues fully match the cache index, the matched cache will be selected.
% as the caching system for the \textcolor{blue}{current session}. 
During in-situ use of \workname, the parsed social factor cues (\S~\ref{Social Factor Cues Parsing}) are sent to a retriever to select the appropriate cache from the cache groups. If the parsed cues fully match a cache index, the corresponding cache will be selected.
However, the user's descriptions may not fully encompass all dimensions of the social factors. \workname~employs a cache merging strategy: if the parsed cues only partially match, all partially matched caches will be merged into a single group and used as the caching.

\noindent\textbf{Runtime Routing and Cache Management}.
During live social interactions, \workname~computes the semantic similarity of the conversational partner's utterances and the keys in the social factor-aware cache.
If the similarity falls below the threshold, it triggers the slow thinking mode in \workname, employing LLMs for in-depth reasoning (\S~\ref{Intention Infer-based Suggestion Generation}). 
We utilize BERT \cite{reimers2019sentence} as the embedding model for similarity calculations. To address the issue of logical inconsistencies in the cache, we set a relatively high threshold of 0.95 in \workname. 
% \noindent\textbf{Cache Management}.
% During live social interactions,
Additionally, as users continue to use \workname~in their daily lives, it continuously records their utterances, conversational partners' utterances, nonverbal cues, and the corresponding social suggestions.
% Additionally, \workname~continuously records the user's utterances, third-party utterances, nonverbal cues, and the corresponding suggestions with users continue to use it in their lives. 
These elements are structured as paired samples, marked with the corresponding social factors, and updated into the social factor-aware caches.
Details of the parameter selection and the impact of cache size can be found in \S~\ref{Hyper-parameter Settings}.

\subsubsection{Intention Infer-based Suggestion Generation}
\label{Intention Infer-based Suggestion Generation}
% limitation of cache, need generation based LLM
% But low speed
% To address this challenge, we propose xxx.
Although the social factor-aware cache can significantly reduce inference latency, it faces challenges in providing logically consistent social suggestions.
Therefore, \workname~employs LLM reasoning to generate in-depth social suggestions. However, directly using LLMs in live social interactions can cause significant system latency, which may disrupt the natural flow of live social conversations and reduce QoE. 
To address this challenge, \workname~employs an intention infer-based reasoning strategy inspired by human behaviors in social interactions.

% To address this challenge, \workname\ employs an intention inference-based reasoning strategy for social suggestion generation.

% Conversational systems can be categorized into retrieval-based and generation-based approaches [xxx]. 
% A straightforward method for fast suggestion generation is to employ retrieval-based solutions, such as cache [xxx]. However, while retrieval-based approaches can achieve high response speed, they often struggle with generalization and logical consistency [xxx]. 
% These limitations are particularly significant in this study due to the absence of a comprehensive database that covers diverse social conversations and corresponding suggestions across various social factors. 

% Additionally, existing studies on fast-slow collaboration LLM reasoning do not focus on the pain points that hinder real-time performance in live social conversation scenarios [xxx].

% Developing a cache that encompasses a wide range of possible situations in daily conversations, while ensuring generalization and logical consistency, is challenging.

\noindent\textbf{Motivation and Insights}.
% Reactive conversational systems only process individual user queries [xxx], whereas live social assistance systems continuously monitor social interactions and proactively generate timely suggestions for the user after others have spoken. To maintain the natural flow of conversation and avoid delays, these suggestions must be provided in real time.
The pipeline of naive LLM-based conversational systems used for live social assistance includes waiting for the conversational partner to finish speaking, offloading, and LLM reasoning. 
In fact, a significant portion of the time is spent waiting for the partner to finish speaking.
% Figure~\ref{fig:pipeline_inference} illustrates the pipeline of the conversational systems used for live social assistance, detailing the processing time for each step. It reveals that a significant portion of the processing time is spent waiting for the other party to finish speaking. 
However, humans can often grasp and understand the other party's intention based on the initial partially spoken words and start early preparation for their response without needing to hear the entire sentence \cite{meyer2023timing}.
After hearing the complete utterance of the other party, humans typically make only slight modifications to their response and reply quickly, thereby maintaining the natural flow of conversations~\cite{templeton2022fast}.
% usually within 250 ms \cite{templeton2022fast},

\noindent\textbf{Intention Infer-based Generation}.
Motivated by this insight, \workname~employs an intention infer-based reasoning strategy for social suggestion generation.
Specifically, \workname~performs real-time speech recognition on the glass side and periodically offloads incomplete utterances to the server.
This solution reduces bandwidth usage compared to directly offloading speech to the smartphone.
Considering the average human speaking speed of 150 words per minute \cite{liu2024andes}, we set the offloading interval to 2 seconds in \workname.
Additionally, we set an additional instruction ``Infer the other party's intention based on partially heard words'' in the prompt to invoke the capabilities of LLMs for reasoning on partial utterances.
Finally, when the other party finishes speaking, the complete sentences are sent to the LLMs to generate in-depth and comprehensive social suggestions.
Details of the prompt can be found in Appendix~\ref{appendix}.
% Figure~\ref{fig:prompt_overall} (in appendix) shows the details of the prompt in \workname.

\subsubsection{Proactive Response Update}
% why proactive. 
Frequent refreshing of the social suggestions displayed on AR glasses can significantly reduce the QoE and usability of \workname, as users do not have enough time to read and grasp the information.
To address this challenge, \workname~employs a proactive response update mechanism.
Specifically, we set the suggestion display refresh interval to 3 seconds, considering that the average human reading speed is 200 words per minute~\cite{liu2024andes} and concise CoT reasoning (\S~\ref{CoT with Concise Response}) limits responses to 70 words.
Additionally, we set an additional instruction in the LLM to determine whether there is any change in the semantics of the other party's two consecutive utterances. 
If the semantic similarity between the two utterances is high, \workname~will not update the social suggestions on the AR glasses.

%% file: Evaluation/Evaluation.tex
\section{Evaluation}
% This section first introduces the experimental setup of this study, including the system implementation, experimental datasets, baselines, and evaluation metrics. We then show the evaluation of \workname, including overall performance and system modules, followed by real-world testing and a user study.
This section introduces the experimental setup, evaluation of \workname, and a real-world user study.

% of this study, including the system implementation, experimental datasets, baselines, and evaluation metrics. We then show the evaluation of \workname, including overall performance and system modules, followed by real-world testing and a user study.

\vspace{-1.0em}
\subsection{Experimental Setup}
% We conduct both simulation experiments and real-world tests to validate the effectiveness of \workname.
% \textcolor{blue}{This section shows xxx.}

% \begin{figure}
%     \centering
%     \begin{subfigure}{0.37\columnwidth}
%         \centering
%         \includegraphics[width=1\columnwidth]{Evaluation/figs/setup-scenario.pdf}
%         % \vspace{-0.1em}
%         % \caption{Social interactions using \workname.}  
%         \caption{Real-world social interactions.}  
%         \label{xxx}
%         \vspace{-1.0em}
%     \end{subfigure}
%     \hfill
%     \begin{subfigure}{0.62\columnwidth}  
%         \centering \includegraphics[width=1.0\columnwidth]{Evaluation/figs/setup-UI.pdf}
%         % \vspace{-1.5em}
%         \caption{UI of \workname~on smart AR glasses.}    
%         \label{xxx}
%         \vspace{-1.0em}
%     \end{subfigure}
     
%     \caption{Real-world test settings.
%     Participants engage in live face-to-face social interactions with other parties, wearing the glasses and receiving social suggestions from \workname~. \textcolor{blue}{The left and right in Figure (b) are the two fields of view of the AR glasses.} \yq{We can update the figure of the glasses and point the camera and display out.}
%     }
%     \label{fig:user_study_scenario}
%       \vspace{-1.0em}
% \end{figure}

\begin{figure}
    \centering
    \begin{subfigure}{0.48\columnwidth}
        \centering
        \includegraphics[width=1\columnwidth]{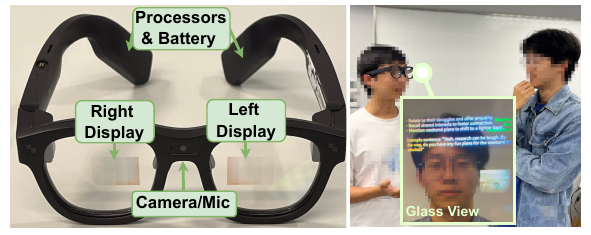}
        % \vspace{-0.1em}
        % \caption{Social interactions using \workname.}  
        \caption{Device setup and real-world social interactions.}  
        \label{fig:glasses}
        \vspace{-1.0em}
    \end{subfigure}
    \hfill
    \begin{subfigure}{0.5\columnwidth}  
        \centering \includegraphics[width=1.0\columnwidth]{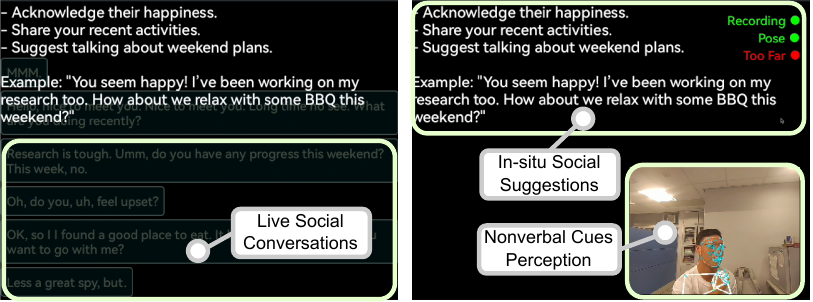}
        % \vspace{-1.5em}
        % \caption{UI of \workname~on smart AR glasses.}    
        \caption{Screenshots from the AR glasses' left and right displays. 
        % \zy{The lower part is not clear...}
        }    
        \label{fig:glass_view}
        \vspace{-1.0em}
    \end{subfigure}
     
    \caption{Real-world test settings.
    Participants engage in live face-to-face social interactions with other parties, wearing the glasses and receiving social suggestions from \workname.}
    % The left and right in Figure (b) are the two fields of view of the AR glasses.}
    \label{fig:user_study_scenario}
      \vspace{-1.2em}
\end{figure}

\subsubsection{System Implementation}
\label{implementation}
We selected the off-the-shelf RayNEO X2~\cite{rayneo} smart glasses as our hardware platform. These glasses run on the Android 12 operating system, boasting 6GB of RAM and 128GB of storage. They come equipped with a front-facing camera, dual-eye waveguide color displays, and three microphones (Figure~\ref{fig:glasses}). Our solution is also compatible with other AR glasses, including models from INMO~\cite{inmo_air2}.

Figure~\ref{fig:glass_view} illustrates the glass-view suggestions presented to the user. Our implementation consists of an on-glass app and a Python-based server. The glasses app is developed using 4,038 lines of Java and Kotlin code. To ensure user privacy, we process video and audio locally. We use GPU-accelerated MediaPipe~\cite{pipe} models for efficient pose and facial landmark tracking. 
% We limit pose recognition to 3 frames per second at a 640$\times$480 resolution, keeping power consumption under 2 watts—comparable to a standard camera app. 
For voice recognition, we employ Azure Voice Recognition with local voice feature extraction. Notably, speaker recognition features have been discontinued by major voice recognition platforms due to privacy concerns~\cite{microsoft_speaker_recognition}, which motivates our use of vibration-based primary user identification.
% , motivating our use of vibration-based primary user identification.

The server, built with Python, handles social cue recognition and proactive suggestions using lightweight Scikit-learn models~\cite{scikit-learn} and Langchain~\cite{Chase_LangChain_2022} for LLM coordination. 
The glasses communicate with the server via HTTPS.
% The glasses communicate with the server via HTTPS, with data transfer rates below 100 KB/s.
Due to the low processing requirements, we can deploy most of the server-side code locally on the glasses using Chaquopy~\cite{chaquopy}—the only exception being the LLM inferences. 
% We plan to release this implementation code to support more R\&D in assistive AR use cases.

\subsubsection{Experiments on Public Datasets}
% \yq{We can name this part "Dataset" }
\label{exp_settings}
We first validate the effectiveness of \workname~using public multi-turn dialogue datasets \cite{li2017dailydialog,
jandaghi-etal-2024-faithful-persona,
zhan2023socialdial}. However, to the best of our knowledge, no public datasets contain social conversations that include comprehensive nonverbal cues and personas. Additionally, the conversations in existing datasets remain fixed, and cannot be dynamically steered by the social suggestions generated by assistive systems. 
Therefore, we use two LLM agents for role-playing social interactions with the help of social assistive systems.
This subsection details the datasets and LLM agent settings.
% This subsection introduces the details of the datasets and the settings of the LLM agents.
% Therefore, we conduct two types of experiments on synthetic datasets. This subsection introduces the details of the experimental paradigm and the construction of synthetic datasets.

\noindent\textbf{Dialogue Datasets}.
We use three public multi-turn dialogue datasets to validate the effectiveness of \workname.
\begin{itemize}[leftmargin=*]

\item 
\textbf{DailyDialog} \cite{li2017dailydialog} dataset contains 13,118 multi-turn conversations covering a wide range of daily topics.
% Each sample xxx.

\item 
\textbf{Synthetic-Persona-Chat} \cite{jandaghi-etal-2024-faithful-persona} dataset is a conversational dataset featuring personas for both parties. 
Compared with DailyDialog, conversations in Synthetic-Persona-Chat are persona-conditioned.
Each sample includes the personas of the two parties and their conversations.
The dataset comprises 20,000 conversations and 5,000 personas in total.
The personas in this dataset are used to construct the LLM agents.

\item
\textbf{SocialDial} \cite{zhan2023socialdial} dataset comprises more than 6.4K multi-turn dialogues with social factor annotations, each annotated with social factors such as social relations and social norms.

% \textbf{MELD} \cite{jandaghi-etal-2024-faithful-persona} dataset comprises over 13K utterances from multiple speakers, extracted from the TV show Friends. 
% Each dialogue includes multi-turn conversations with emotion annotations. 
% Although the dataset contains multimodal data, the annotations include only emotion labels with limited nonverbal cue categories, and it does not contain personas.

\end{itemize}
For all datasets, we randomly select one speaker as the primary user in the experiment and the other as the conversational partner.
% However, since conversations in the datasets remain fixed and will not be influenced by the generated social suggestions, we set up two LLM agents to role-play the social interactions.
However, since the conversations in the datasets remain fixed and are not influenced by the generated social suggestions, we set up two LLM agents to role-play the social interactions.
% To further assess the impact of the social suggestions, we conduct the second paradigm experiment using LLMs agent for role-playing [xx].
Existing studies have demonstrated the effectiveness of using LLMs as agents for role-playing, such as negotiations~\cite{hua2024assistive} and medical diagnosis simulations~\cite{yang2024drhouse}. We extend the role-playing capabilities of LLMs to live social interaction settings.
Specifically, we set up two separate LLMs to create agents for role-playing as the user and the conversational partner during live social conversations.
The user agent interacts with the partner agent while incorporating the social suggestions generated by the assistive systems.
We randomly select 50 samples in each dataset for experiments.
% We simulate social interactions using two modes: conversation-based and social factor-based. In the conversation-based mode, both LLM agents initiate interactions using dialogues from three datasets. In the social factor-based mode, the agents start interactions based on the social factors defined in \cite{zhan2023socialdial}.

\noindent\textbf{Setup of Simulated Agents.}
% First, we incorporate the personas from the dataset into the prompts for the LLM-simulated agents. 
First, we set instructions in the prompts to enable the user agent and the conversational partner agent to conduct multi-turn social conversations.
Additionally, the agent contains nonverbal behavior and 
persona attributes.
For nonverbal behaviors, we use a series of nonverbal behaviors that encompass typical cues in face-to-face social conversations, including facial expressions, gestures, and physical proximity~\cite{duncan1969nonverbal}. 
Each type of behavior contains multiple subcategories, such as confusion and frowning for facial expressions, and nodding and head shaking for gestures. 
Table~\ref{details_nonverbal} details the categories of nonverbal behaviors in our experiments.
During each turn of the social conversation, we randomly select one subcategory from the nonverbal behaviors category as the partner LLM agent's current simulated nonverbal behaviors.

Furthermore, 
since Synthetic-Persona-Chat is the only dialogue dataset that contains the personas of both parties, we use the personas as personal profiles to incorporate into the prompt for the
LLM and set up the simulated user and conversational partner agents. Additionally, the conversations corresponding to the
user and the conversational partner are used as historical data for implicit persona extraction.
Figure~\ref{fig:prompt_agent_datasets} and Figure~\ref{fig:prompt_agent_social} show the prompt of the simulated user and conversational partner agents.

% In addition, we incorporate social factor information into the prompts for the LLM-simulated agents to enhance the realism of the social conversations.
% Figure xx and Figure xx show the prompt of the simulated user and opponent agents, respectively.

% \noindent\textbf{Synthetic Dataset Construction.}

% \noindent\textbf{Nonverbal Behaviors.}
% We use a series of nonverbal behaviors that encompass typical cues in face-to-face social conversations, including facial expressions, gestures, and personal space [xx]. 
% Each type of behavior contains multiple subcategories, such as confusion and frowning for facial expressions, and nodding and head shaking for gestures. Table xxx details the categories of nonverbal behaviors in our experiments. During each turn of the social conversation, we randomly select one subcategory from the nonverbal behaviors category as the opponent LLM agent's current simulated nonverbal behaviors.

% \noindent\textbf{Personas}.
% Since Synthetic-Persona-Chat is the only dialogue dataset that contains the personas of both parties, we use the personas as personal profiles to incorporate into the prompt for the
% LLM and set up the simulated user and opponent agents. Additionally, the conversations corresponding to the
% user and the other party are used as historical data for implicit persona extraction.

\noindent\textbf{Two Role-play Paradigms}.
The LLM role-play of social interactions are conducted using two paradigms: dialogue-based and social factor-based. 
In the dialogue-based role-play, both LLM agents initiate interactions using dialogues from three datasets. 
In the social factor-based paradigm, the agents start interactions guided by social factors defined in~\cite{zhan2023socialdial}, including social norms, social relations, formality, and location (see Figure~\ref{fig:prompt_agent_datasets} and Figure~\ref{fig:prompt_agent_social}).
% Details of the prompt settings can be seen in Figure~\ref{fig:prompt_agent_datasets} and Figure~\ref{fig:prompt_agent_social}.
We randomly select subcategories of these social factors to create diverse social scenarios and conduct experiments. Table~\ref{details_social_scenarios} shows the details of the social scenario settings in our experiments.

% \begin{table}[t]
% % \footnotesize
% \small
% \centering
% \begin{minipage}{0.48\textwidth}
% \centering
% \caption{Details of nonverbal behaviors in \workname.}
% \vspace{-1em}
% \begin{tabular}{c|p{0.55\textwidth}}
% \toprule
% \textbf{Nonverbal Cues} & \textbf{Sub-Categories} \\
% \midrule
% Facial Expression & Confusion, Neutral, Frowning, \newline Happiness, Sadness, Anger \\
% Gestures & Nodding, Shaking Head, Hands Spreading, Thumbs Up\\
% Personal Space & Proper, Too Far, Too Close \\
% \bottomrule
% \end{tabular}
% \label{details_nonverbal}
% \end{minipage}%
% \hfill
% \begin{minipage}
% {0.5\textwidth}
% \centering
% \caption{Details of social scenarios in our experiments.}
% \vspace{-1em}
% \renewcommand{\arraystretch}{0.8} % Adjust row height for right table

% \begin{tabular}{c|p{0.52\textwidth}}
% \toprule
% \textbf{Social Scenes} & \textbf{Social Factors} \\
% \midrule
% Scenario 1 & Casual Greeting, Peer-Peer, Informal, Open Area \\
% % Scene 2 & Greeting, student-Professor, Informal, Conference Break \\
% Scenario 2 & Polite Requesting, Mentor-Mentee, Formal, Office \\
% Scenario 3 & Direct Persuasion, Elder-Junior, Informal, Open Area\\
% \bottomrule
% \end{tabular}
% \label{details_social_scenarios}
% \end{minipage}
% \end{table}

\begin{table*}[t]
\small
\centering
\begin{minipage}[t]{0.48\textwidth} % Use [t] for top alignment in minipage
\centering
\caption{Details of nonverbal behaviors in \workname.}
\vspace{-1em}
\begin{tabular}{c|p{0.55\textwidth}}
\toprule
\textbf{Nonverbal Cues} & \textbf{Sub-Categories} \\
\midrule
Facial Expression & \setlength{\baselineskip}{0.9\baselineskip} Confusion, Neutral, Frowning, \newline Happiness, Sadness, Anger \\
Gestures & \setlength{\baselineskip}{0.9\baselineskip} Nodding, Shaking Head, Hands Spreading, Thumbs Up \\
Personal Distance & Proper, Too Far, Too Close \\
\bottomrule
\end{tabular}
\vspace{-1em}

\label{details_nonverbal}
\end{minipage}%
\hfill
\begin{minipage}[t]{0.48\textwidth} % Use [t] for top alignment in minipage
\centering
\caption{Details of social scenarios in our experiments.}
\vspace{-1em}
\renewcommand{\arraystretch}{0.8} % Adjust row height for right table

\begin{tabular}{c|p{0.52\textwidth}}
\toprule
\textbf{Social Scenes} & \textbf{Social Factors} \\
\midrule
Scenario 1 & \setlength{\baselineskip}{0.9\baselineskip} Casual Greeting, Peer-Peer, Informal, Open Area \\
Scenario 2 & \setlength{\baselineskip}{0.9\baselineskip} Polite Requesting, Mentor-Mentee, Formal, Office \\
Scenario 3 & \setlength{\baselineskip}{0.9\baselineskip} Direct Persuasion, Elder-Junior, Informal, Open Area \\
\bottomrule
\end{tabular}
\vspace{-1em}
\label{details_social_scenarios}
\end{minipage}
% \vspace{-1em}
\end{table*}

\subsubsection{Real-world Evaluation.}
To further validate the effectiveness of \workname, we recruited 20 volunteers to participate in real-world evaluation. 
% We recruit xxx volunteers to participate in our real-world tests.
Each participant wears glasses and engages in face-to-face live social conversations with the conversational partner, assisted by \workname.
% The study has been approved by the IRB, and all participants are consented for data collection.
The study has received IRB approval, and all participants have given consent for data collection.
% The social scenario is the same as the simulation experiments, with the settings detailed in Table~\ref{detatils_social_scenarios}.
Figure~\ref{fig:user_study_scenario} shows the settings of the real-world tests and the system prototype of \workname.
After the experiments, each participant is required to fill out a questionnaire and participate in a user study regarding their experience with \workname.
% \textcolor{red}{[TODO]: Figures of our systems and real-world test settings.}
For details of the real-world testing and user study, please refer to Section \S~\ref{sec_user_study}.

\subsubsection{Evaluation Metrics}
% Social assistance is a generation task, which can not use the existing evlaution mectics like
% However, existing studies have well defined evaluation metrics for discriminative tasks, question-answering tasks or single-turn conversation..
Since generating social suggestions in multi-turn conversations is an open-ended task~\cite{zheng2023judging} without explicit standard answers, existing metrics used for question-answering and classification tasks are not suitable for evaluation.
In this study, we propose the following criteria to validate the effectiveness of social assistive systems. First, we assess the content of the social suggestions provided by the assistive systems. Second, we evaluate their effectiveness when users employ these systems during social interactions.
\begin{itemize}[leftmargin=*]
\item 
\textbf{Personalization}.
% The degree of the social suggestions provided by assistive systems considers the user's implicit personas.
This metric evaluates the quality of social suggestions from a personalized perspective, assessing whether the social suggestions incorporate users' implicit personas, including personal interests and backgrounds. 
It is similar to the score introduced in studies \cite{xu2024can}.
A higher \textit{personalization} score for social suggestions can enhance user engagement in social interactions, as users are more familiar with the content.

% The degree of the social suggestions considers the user's implicit personas.
% [why use this criteria]

\item 
\textbf{Engagement}.
This metric has been widely used in previous studies \cite{kim2022prosocialdialog} to evaluate user engagement in conversational systems. However, in the context of a social assistive system, we extend \textit{engagement} to the conversational partner's perspective, assessing whether the social suggestions consider the conversational partner's implicit personas. Higher \textit{engagement} in social suggestions indicates the conversational partner's increased willingness and enhanced participation in social interactions.

% \item 
% \textbf{{Empathy}}.
% This metric determines whether the social suggestions consider the opponent's implicit personas.
% A higher \textit{Empathy} score for social suggestions can enhance opponent's engagement in social interactions, as users are more familiar with the content.
% The degree of social suggestions considers the opponent's implicit personas.
% [why use this criteria]

\item 
\textbf{Nonverbal Cues Utilization}.
Given the importance of nonverbal cues in social interactions, we propose this metric to assess whether social suggestions take into account the conversational partner's nonverbal cues.

% The extent to which nonverbal cues from live social conversations are utilized in the suggestions provided by assistive systems.

% \item 
% \textbf{{Logical Consistency}}.
% Evaluate whether the suggestion is relevant to the opponent's current utterance and logically consistent.

% \item 
% \textbf{{Latency}}.

\end{itemize}

% Guided by these metrics, we evaluate the social suggestions using both LLM-based and manual scoring.

% \noindent\textbf{LLM Evaluation.}
Existing studies have demonstrated that LLMs can be utilized to assess the quality of multi-turn conversations~\cite{yang2024drhouse} and open-ended tasks \cite{zheng2023judging,xu2024can}, known as LLM-as-a-Judge \cite{zheng2023judging}. 
We adopt the LLM-as-a-Judge \cite{zheng2023judging} and extend it to evaluate the quality of open-ended social suggestions, guided by the aforementioned criteria.
We use GPT-4o as the base model for LLM evaluation throughout this paper.

% \noindent\textbf{Manual Scoring.}
% We also recruit xxx.

\subsubsection{Baselines}
Since no previous studies have developed assistive systems that provide social suggestions during live interactions between two parties, we established several baseline approaches to evaluate our system.

\begin{itemize}[leftmargin=*]

\item 
\textbf{\textit{Zero-shot}}.
This is a prompt-based solution. We include instructions in the prompts of LLMs to provide social suggestions during conversations between the user and conversational partner.
We use GPT-4o as the base LLM.
Figure~\ref{fig:prompt-zeroshot} shows the prompt of \textit{Zero-shot}.

\item 
\textbf{\textit{CoT}} \cite{wei2022chain}.
The settings are the same as in \textit{Zero-shot}, but employ the CoT reasoning strategy.
Figure~\ref{fig:prompt-cot} shows the prompt of this baseline.

\item 
\textbf{\textit{Prompt-based Tianji}} \cite{tianji2024}.
This is one of the state-of-the-art prompt-based LLMs developed for social interactions. 
We set the \textit{Prompt-based Tianji} as the mode of interpersonal communication scenario in our experiments. 
The prompts used in Tianji are the same as the baselines for Zero-shot and CoT.

\item 
\textbf{\textit{Retrieval-based Tianji}} \cite{tianji2024}.
This is one of the state-of-the-art retrieval-based LLMs developed for social interactions. 
We set the \textit{Retrieval-based Tianji} as the mode of ``How to speak dialogue'' scenario in our experiments. 
% We use the assistive conversation scenario for our experiments. 
The prompts used in Tianji are the same as the baselines for Zero-shot and CoT.

% \item 
% \textbf{\textit{w/o P}}.
% This method excludes the implicit persona adaptation module for generating social suggestions, while the remaining components are identical to those in \workname.
% % Without Persona Clues
% % What does this method means?
% % high-level
% % specifically

% \item 
% \textbf{\textit{w/o N}}.
% This method omits the nonverbal cues integration module for generating social suggestions. All other components remain the same as in \workname.

% \item 
% \textbf{{Without Fast-slow Collaboration.}}.

\end{itemize}

\begin{figure}[t]
    \centering
    \begin{subfigure}{0.49\columnwidth}  
        \centering \includegraphics[width=0.95\textwidth]{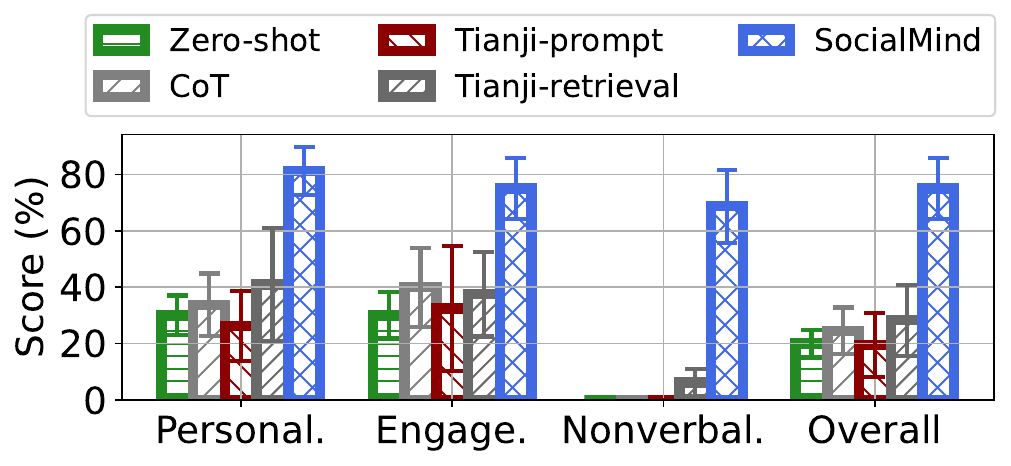}
        \vspace{-0.5em}
        \caption{DailyDialog dataset.}
        \vspace{-1.0em}
        \label{fig:overall_dailydialog}
    \end{subfigure}
    \hfill
    \begin{subfigure}{0.49\columnwidth}  
        \centering 
        \includegraphics[width=0.95\textwidth]{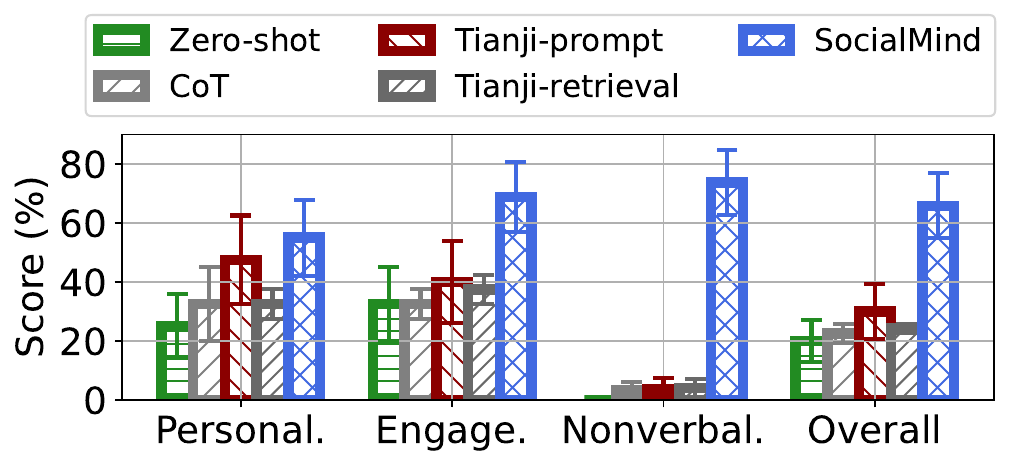}
        \vspace{-0.5em}
        \caption{Synthetic-Persona-Chat dataset.}    
        \label{fig:overall_synpers}
        \vspace{-1.0em}
    \end{subfigure}
     % \vspace{-1.0em}
    \vskip\baselineskip
    \begin{subfigure}{0.49\columnwidth}   
        \centering 
        \includegraphics[width=0.95\textwidth]{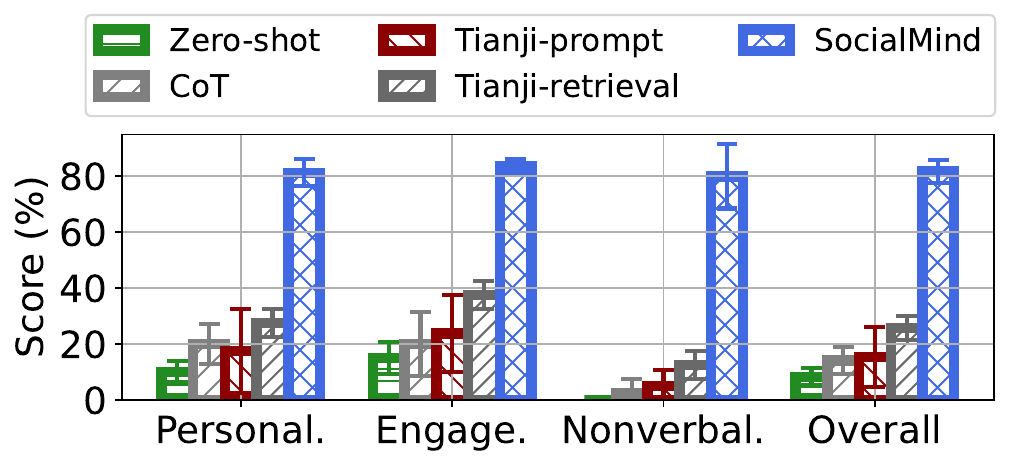}
        \vspace{-0.5em}
        \caption{SocialDial dataset.}\vspace{-1.0em}    \label{fig:overall_socialdial}
    \end{subfigure}
    \hfill
    \begin{subfigure}{0.49\columnwidth}
        \centering
        \includegraphics[width=0.95\textwidth]{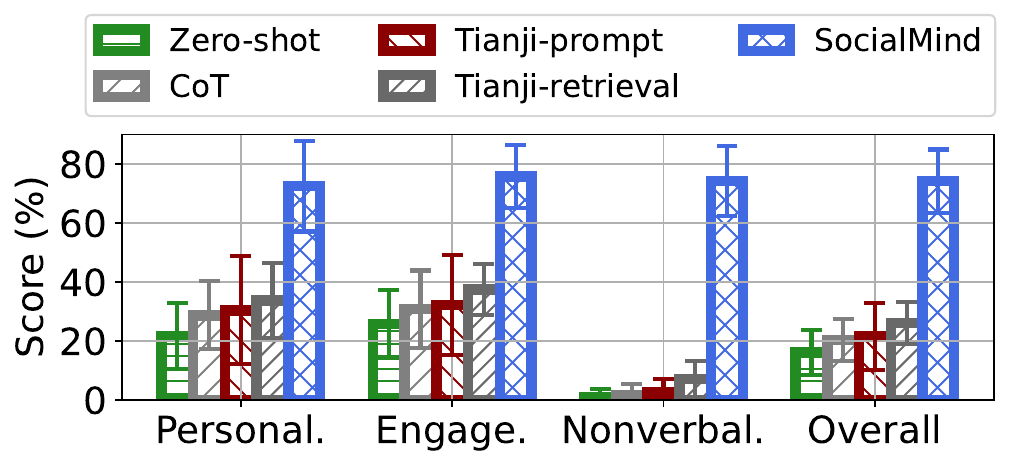}
        \vspace{-0.5em}
        \caption{Overall.}  \label{fig:overall_overall}
        \vspace{-1.0em}
    \end{subfigure}
     % \vspace{-1.0em}
    \caption{Overall performance of the social suggestions generated by \workname~and baselines across three datasets.   Personal. means \textit{personalization} score.
    Engage. means \textit{engagement} score.
            Nonverbal. means \textit{nonverbal cues utilization} score.
    Overall means the average scores among three datasets.
    }
\label{fig:overall_performance}
      \vspace{-1.3em}
\end{figure}

\subsection{Overall Performance}
This section shows the overall performance of \workname~ under scenarios with varying social factors.
\subsubsection{Quantitative Results.}
We first compare the performance of \workname~and baseline approaches using quantitative evaluation metrics, including \textit{personalization}, \textit{engagement}, and \textit{nonverbal cues utilization}.

% \begin{figure}[t]
%     \centering
%     \begin{subfigure}{0.49\columnwidth}  
%         \centering \includegraphics[width=0.95\textwidth]{Evaluation/figs/overall_personalization.pdf}
%         \vspace{-0.5em}
%         \caption{Personalization.}
%         \vspace{-1.0em}
%         \label{fig:overall_personalization}
%     \end{subfigure}
%     \hfill
%     \begin{subfigure}{0.49\columnwidth}  
%         \centering 
%         \includegraphics[width=0.95\textwidth]{Evaluation/figs/overall_engagement.pdf}
%         \vspace{-0.5em}
%         \caption{Engagement.}    
%         \label{fig:overall_engagement}
%         \vspace{-1.0em}
%     \end{subfigure}
%      % \vspace{-1.0em}
%     \vskip\baselineskip
%     \begin{subfigure}{0.49\columnwidth}   
%         \centering 
%         \includegraphics[width=0.95\textwidth]{Evaluation/figs/overall_nonverbal.pdf}
%         \vspace{-0.5em}
%         \caption{Nonverbal Cues Utilization.}\vspace{-1.0em}    \label{fig:overall_nonverbal}
%     \end{subfigure}
%     \hfill
%     \begin{subfigure}{0.49\columnwidth}
%         \centering
%         \includegraphics[width=0.95\textwidth]{Evaluation/figs/overall_overall.pdf}
%         \vspace{-0.5em}
%         \caption{Overall.}  \label{fig:overall_overall}
%         \vspace{-1.0em}
%     \end{subfigure}
%      % \vspace{-1.0em}
%     \caption{Overall performance of the social suggestions generated by \workname~and baselines.
%     }
% \label{fig:overall_performance}
%       \vspace{-1.5em}
% \end{figure}

\begin{figure}
    \centering
    \begin{subfigure}{0.33\columnwidth}
        \centering
        \includegraphics[width=1\textwidth]{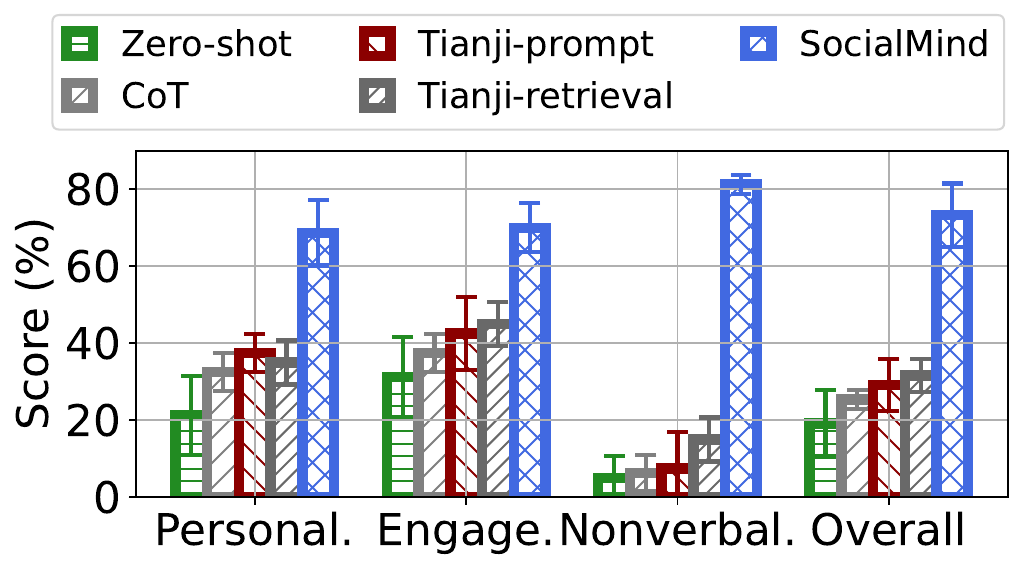}
        \vspace{-1.5em}
        \caption{Social Scenario 1.}  \label{fig:overall_s1}
        \vspace{-1.0em}
    \end{subfigure}
    \hfill
    \begin{subfigure}{0.33\columnwidth}  
        \centering 
    \includegraphics[width=1\textwidth]{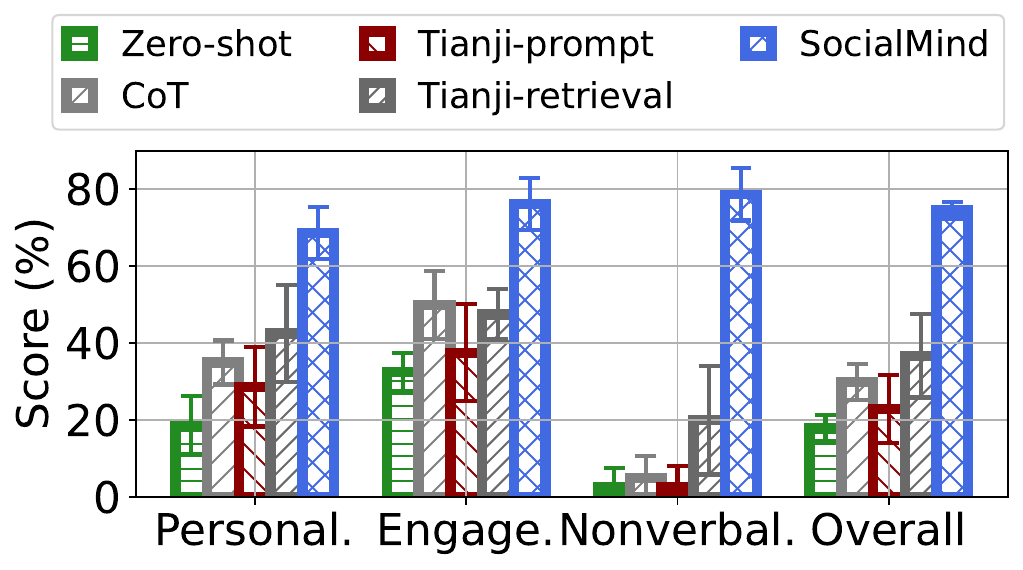}
        \vspace{-1.5em}\caption{Social Scenario 2.}\vspace{-1.0em}    \label{fig:overall_s2}
    \end{subfigure}
     % \vspace{-1.0em}
    \begin{subfigure}{0.33\columnwidth}  
        \centering 
    \includegraphics[width=1\textwidth]{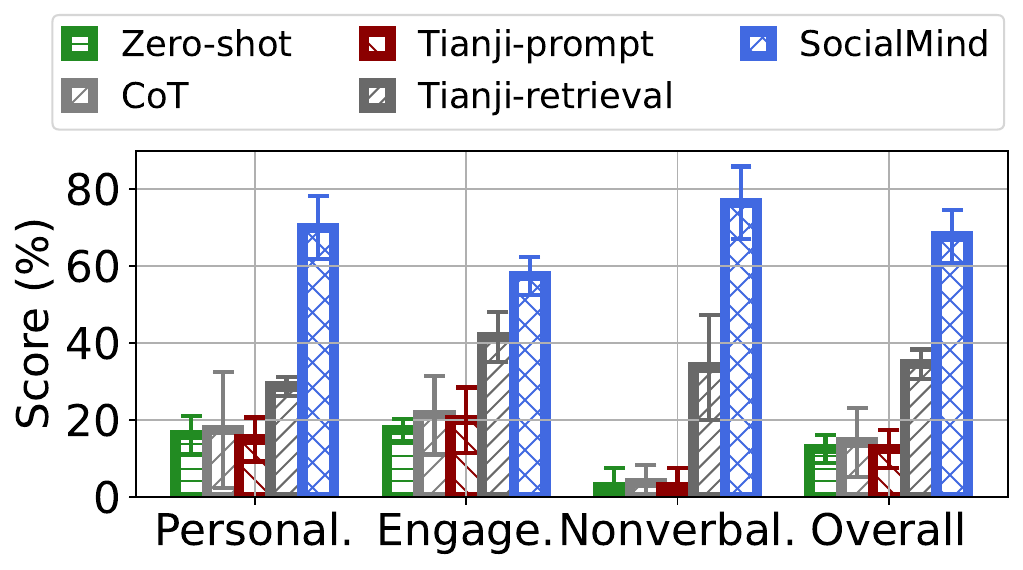}
    \vspace{-1.5em}
        \caption{Social Scenario 3.}    
        \label{fig:overall_s3}
        \vspace{-1.0em}
    \end{subfigure}
     % \vspace{-1.0em}
    \caption{
    Social suggestion performance across different types of social scenarios.
Two LLM agents are prompted to engage in social interactions with specific social factors constrained.
    }
\label{fig:overall_performance_scene}
      \vspace{-1em}
\end{figure}

\noindent\textbf{Overall Performance.}
Figure~\ref{fig:overall_performance} shows the quantitative results of \workname~and baselines across the three datasets, where the user agent and the conversational partner agent engage in dialogue-based role-play.
\workname~achieves state-of-the-art performance across all datasets compared to the baselines.
Results indicate that \workname~achieves 38.7\% higher \textit{personalization} and 38.3\% higher \textit{engagement} than the top-performing baselines, validating its effectiveness in integrating implicit persona cues into social suggestions.
Additionally, \workname~achieves a 61.7\% higher \textit{nonverbal cues utilization} compared to the best baselines, validating its effectiveness in incorporating nonverbal cues into social suggestions. 
This significant improvement is due to \workname's advantage in incorporating multi-modal nonverbal cues, unlike conversation-only baselines.
% This significant improvement is due to baseline approaches being conversation-based solutions that do not perceive the multimodal nonverbal behavior of the conversational partner.
% In addition, \workname~achieves 61.7\% higher \textit{nonverbal cues utilization} compared to the best baselines, validating the effectiveness of \workname~in incorporating nonverbal cues into social suggestions during live social interactions.
% The significant improvement in this score is because baseline social assistive approaches are conversation-based solutions which do not perceive the multi-modal nonverbal behavior of the other parties.
% Figure~\ref{fig:overall_performance_scene} shows the performance of social suggestions across different social factors.
% \workname~achieves highest overall performance under the three social factors, validating the adaptability of \workname in diverse social situations.
Figure~\ref{fig:overall_performance_scene} shows the performance of social suggestions across different social scenarios, where two LLM agents are prompted to engage in social interactions with specific social factors constrained.
\workname~achieves the highest overall performance across all scenarios, validating its adaptability in diverse social situations.
Moreover, results show that employing the CoT reasoning strategy can enhance the instruction-following performance of LLMs, achieving up to 5.9\% higher \textit{overall} scores over the zero-shot baseline. Consequently, CoT is also utilized in \workname~ for reasoning.

% if cot is higher, say we use cot as well.
% if cot is not higher, explain why, and why we still use cot in socialcoach.

% CoT can increase performance, the instruction following performance, thus is also included in the prompt of \workname.
% Since baseline approaches do not contains the design of personas and nonverbal cues extraction, \workanme achieves an xxx higher overall score.

% \noindent\textbf{Performance Across Different Social Scenarios.} 

\begin{figure}
    \centering
    \begin{subfigure}{0.49\columnwidth}
        \centering
\includegraphics[width=1\textwidth]{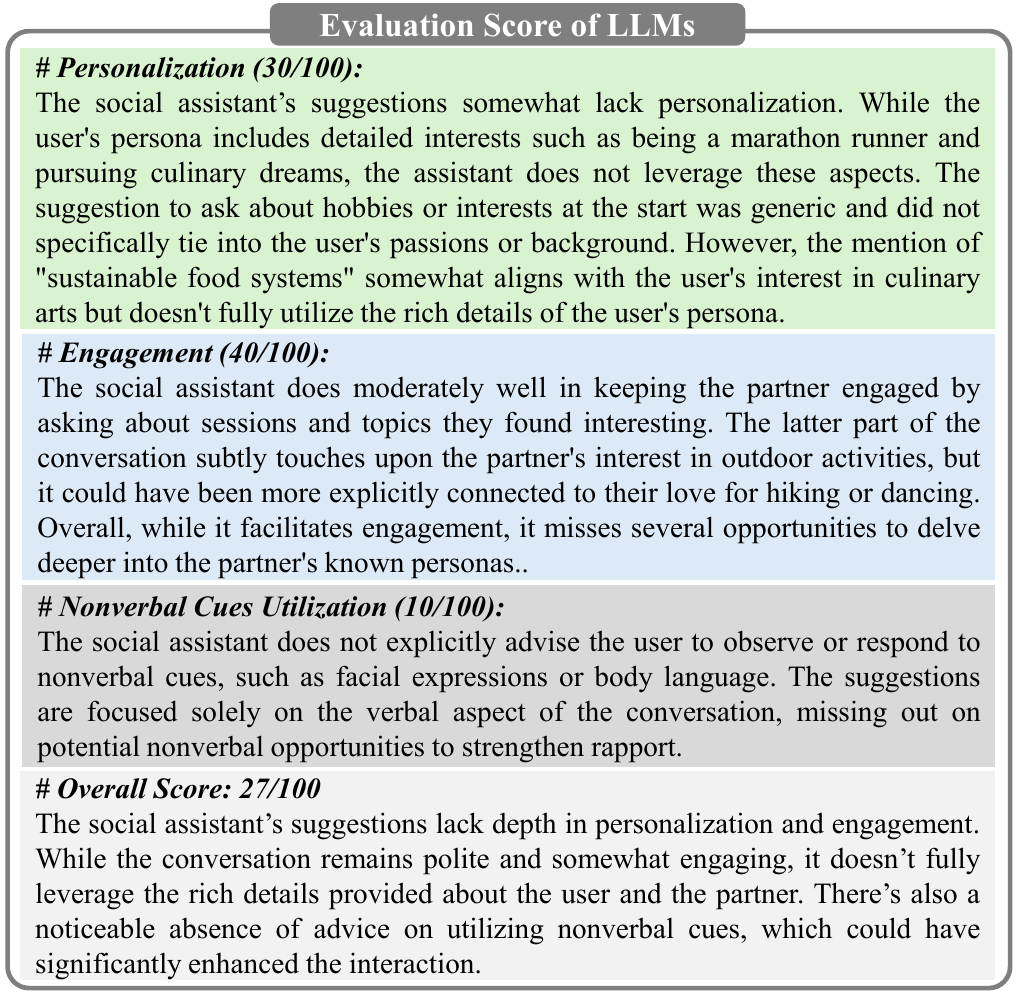}
        \vspace{-1.5em}
        \caption{Example scores of Tianji-retrieval.}  \label{fig:LLM-evaluation-case-low}
    \end{subfigure}
    \hfill
    \begin{subfigure}{0.49\columnwidth}  
        \centering 
    \includegraphics[width=1\textwidth]{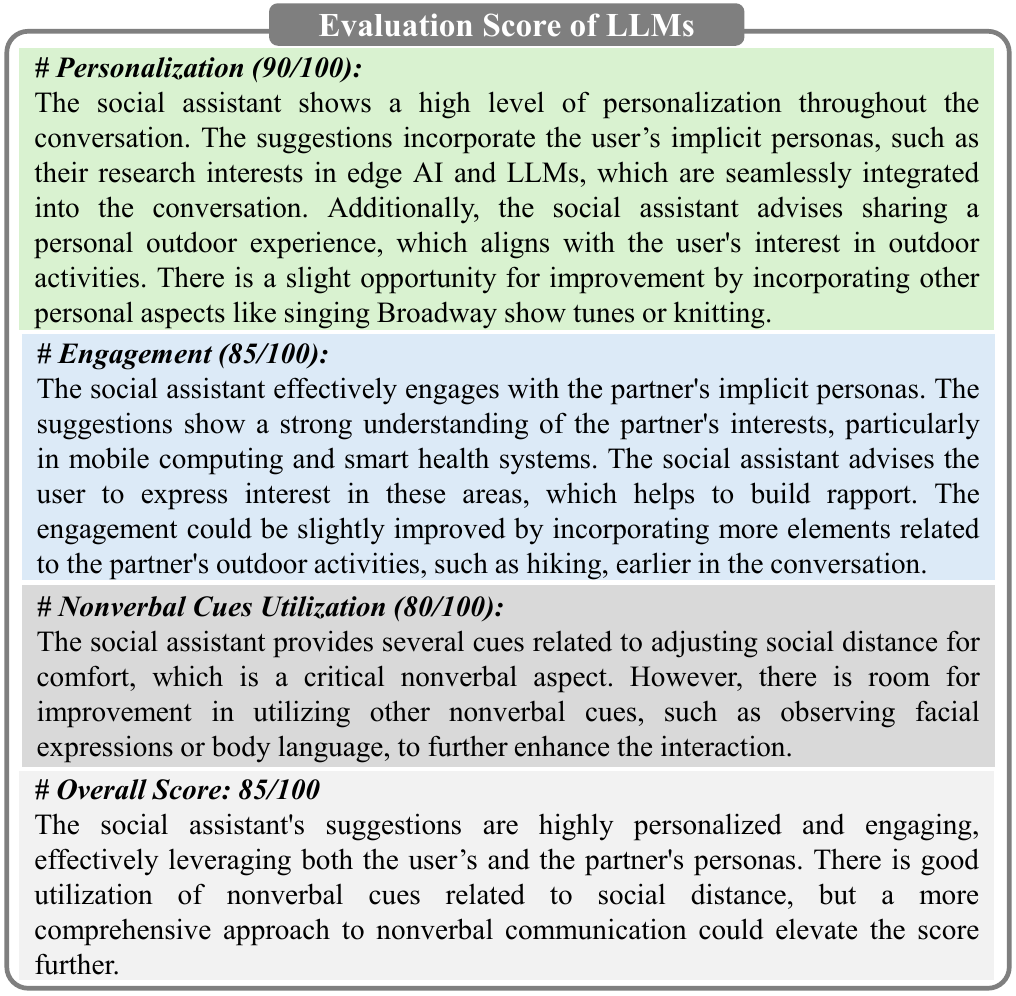}
    \vspace{-1.5em}
    \caption{Example scores \workname.}    \label{fig:LLM-evaluation-case-high}
    \end{subfigure}
     \vspace{-1.em}
    \caption{Example of using LLMs for scoring the social suggestions.
    }
\label{fig:LLM_scoring}
      \vspace{-1.5em}
\end{figure}

\noindent\textbf{Explanation of LLM Evaluation.} 
We also provide examples to show the effectiveness of LLMs in evaluating and scoring social suggestions. 
Figure~\ref{fig:dialogue} shows examples of social conversations and social suggestions from \workname~and baselines.
Figure~\ref{fig:LLM_scoring} shows the corresponding scores and explanations generated by LLMs. 
Results in Figure~\ref{fig:LLM_scoring} show that when provided with social conversations, suggestions, persona ground truth, and explanations of nonverbal cues, LLMs can provide reliable evaluation scores and reasonable explanations.
For example, Figure~\ref{fig:dialogue-socialcoach} demonstrates that the suggestions provided by \workname~incorporate hints of the partner's nonverbal cues and both parties' implicit persona information, resulting in a high score from the LLM evaluator.

\subsubsection{Qualitative Results.}
\begin{figure}
  \centering
\includegraphics[width=0.8\linewidth]{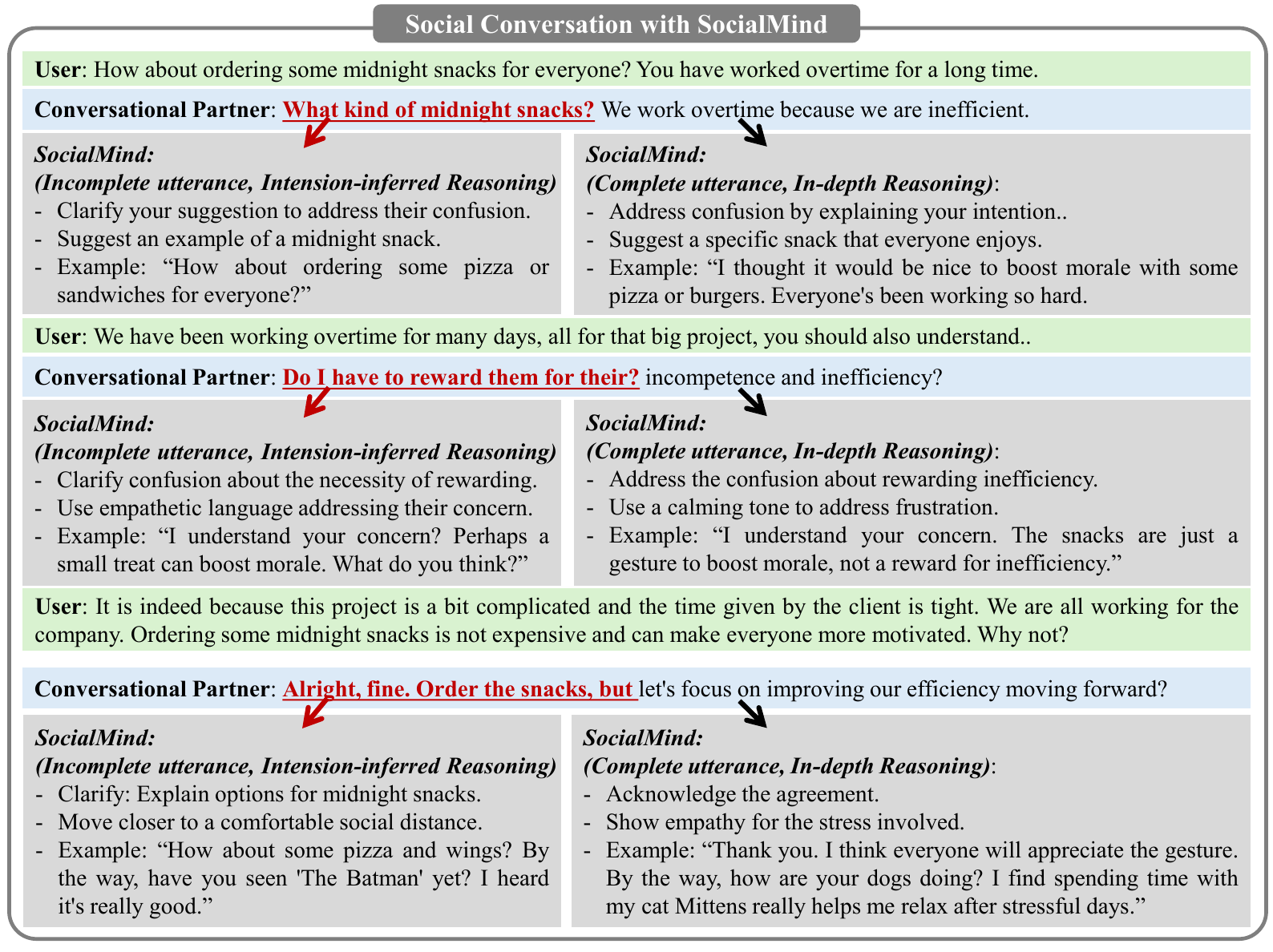}
\vspace{-1em}
  \caption{Example of the intention infer-based suggestion generation in \workname.
  Words highlighted in red indicate that \workname~utilizes the partially spoken sentences of the conversational partner to generate instant suggestions by inferring their intentions.
  It also provides an in-depth social suggestion using complete sentences when the partner finishes speaking.}
  % \vspace{-.5em}
  \label{fig:dialogue-fast-slow}
  \vspace{-1em}
\end{figure}

To better understand \workname's performance, we provide the dialogues containing the social conversations and corresponding social suggestions provided by \workname~and baselines. 
Key observations are summarized as follows:

\noindent\textbf{\textit{Observation 1: Intention infer-based reasoning strategy enables \workname~ to provide logically consistent and instant social suggestions.
}}
% Figure~\ref{dialogue-fast-slow} presents an example of a social conversation in which the opponent's utterance is lengthy.
Figure~\ref{fig:dialogue-fast-slow} shows an example of the intention infer-based social suggestion generation in \workname.
% Baseline approach waits for the other party to finish speaking before LLM reasoning. As a result, the user must wait xxx seconds for social suggestions to be generated, disrupting the natural fluency of the conversation and significantly impacting its practicality.
\workname~utilizes the partially spoken sentences of the conversational partner (marked as red in Figure~\ref{fig:dialogue-fast-slow}) to generate instant suggestions by inferring the partner's intentions, providing the user with early preparation for their response.
Results show that \workname~can comprehend the partner's intentions, draft preliminary social suggestions from incomplete utterances, and deliver these suggestions to the user. This enables the user to promptly begin considering their response strategy.
Once the partner finishes speaking, \workname~also utilizes the complete utterances for further reasoning, offering the user in-depth social suggestions that the user can incorporate to refine their current response or use in the next round of conversation.
% can be used to modify their current response or be incorporated as a supplement in the next round of conversation.
% semantic accpectable.
% ours -> xx ms, before the speaking finish
% raw approach -> xx ms.
% Consistent with the human brain.

\begin{figure}
    \centering
    \begin{subfigure}{0.48\columnwidth}
        \centering
\includegraphics[width=1\textwidth]{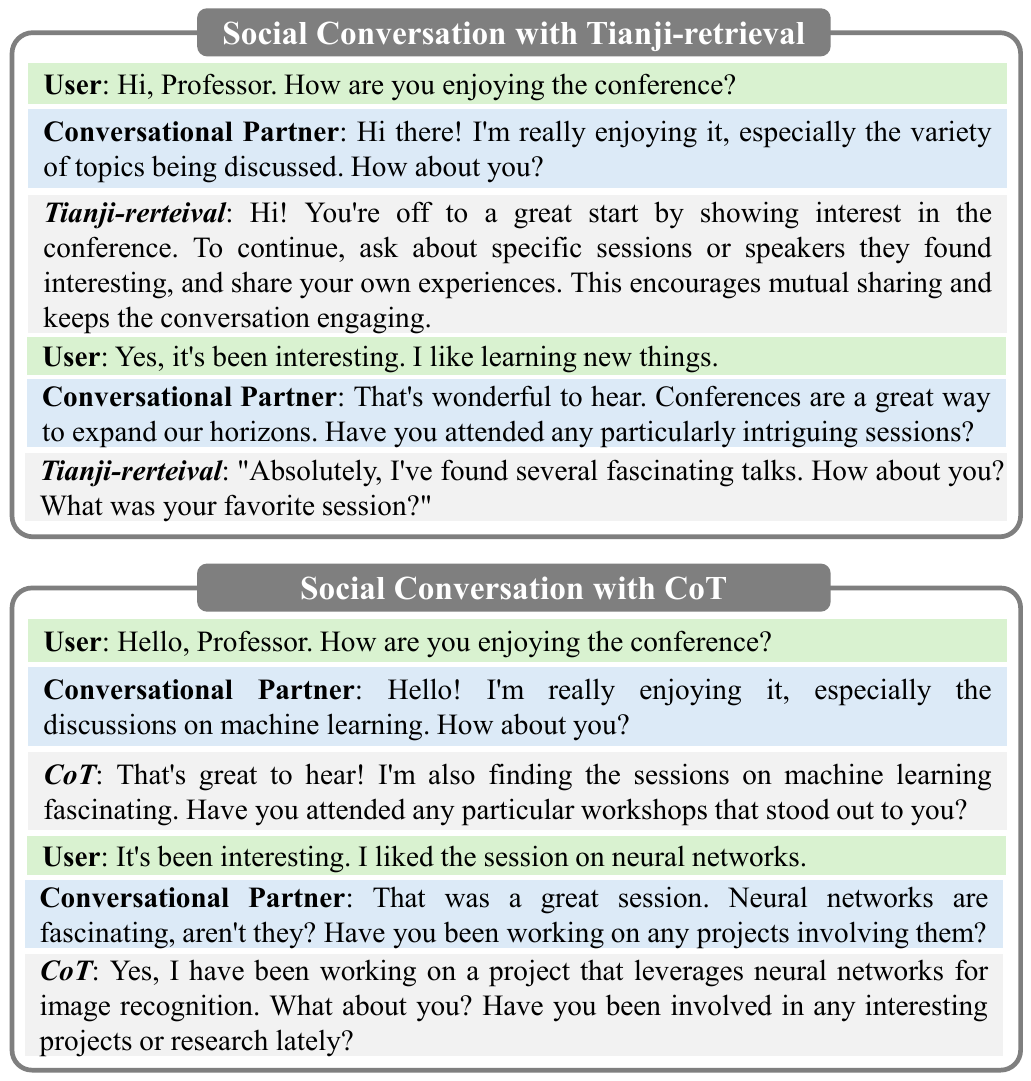}
        \vspace{-1.2em}
        \caption{Tianji-retrieval and CoT.}  \label{fig:dialogue-baseline}
    \end{subfigure}
    \hfill
    \begin{subfigure}{0.485\columnwidth}  
        \centering 
    \includegraphics[width=1\textwidth]{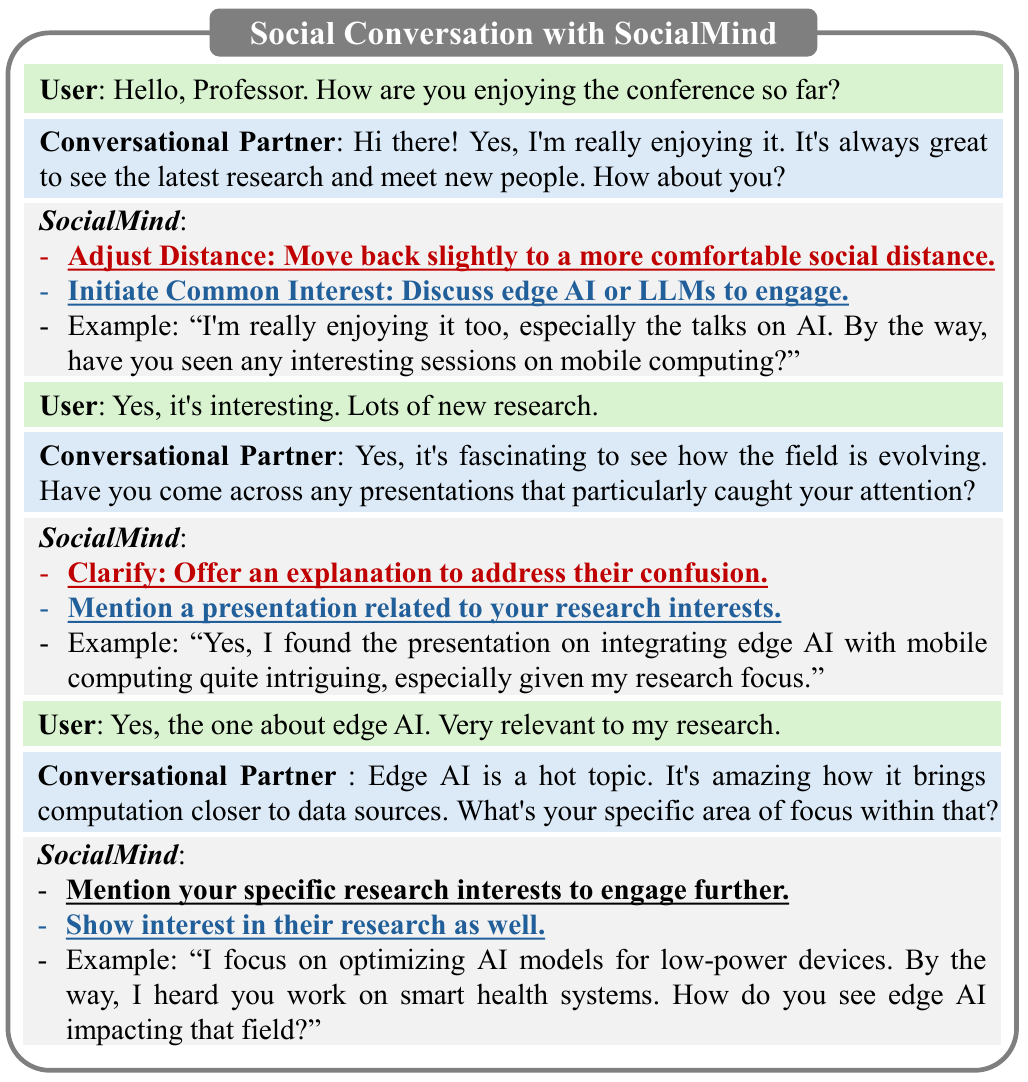}
    \vspace{-1.2em}
    \caption{\workname.}    \label{fig:dialogue-socialcoach}
    \end{subfigure}
     \vspace{-1.em}
    \caption{
    Examples of social conversations and social suggestions \workname~and baselines.
    Words highlighted in red and blue demonstrate that \workname~integrate the nonverbal cues and implicit persona cues into the social suggestions, respectively.
    }
\label{fig:dialogue}
      \vspace{-1.2em}
\end{figure}

\noindent\textbf{\textit{Observation 2: \workname~can incorporate the conversational partner's nonverbal cues during live conversations into social suggestions to assist users}}.
Figure~\ref{fig:dialogue} shows that baseline approaches like CoT and Tianji-retrieval focus exclusively on the verbal aspect of conversations.
However, \workname~ can proactively perceive various nonverbal cues (marked in red), such as facial expressions and personal distance, and incorporate them into social suggestions, resulting in a holistic understanding of the partner's intention and state of mind.

\begin{figure}
\begin{minipage}[t]{0.48\columnwidth}
     \centering
\includegraphics[width=1\textwidth]{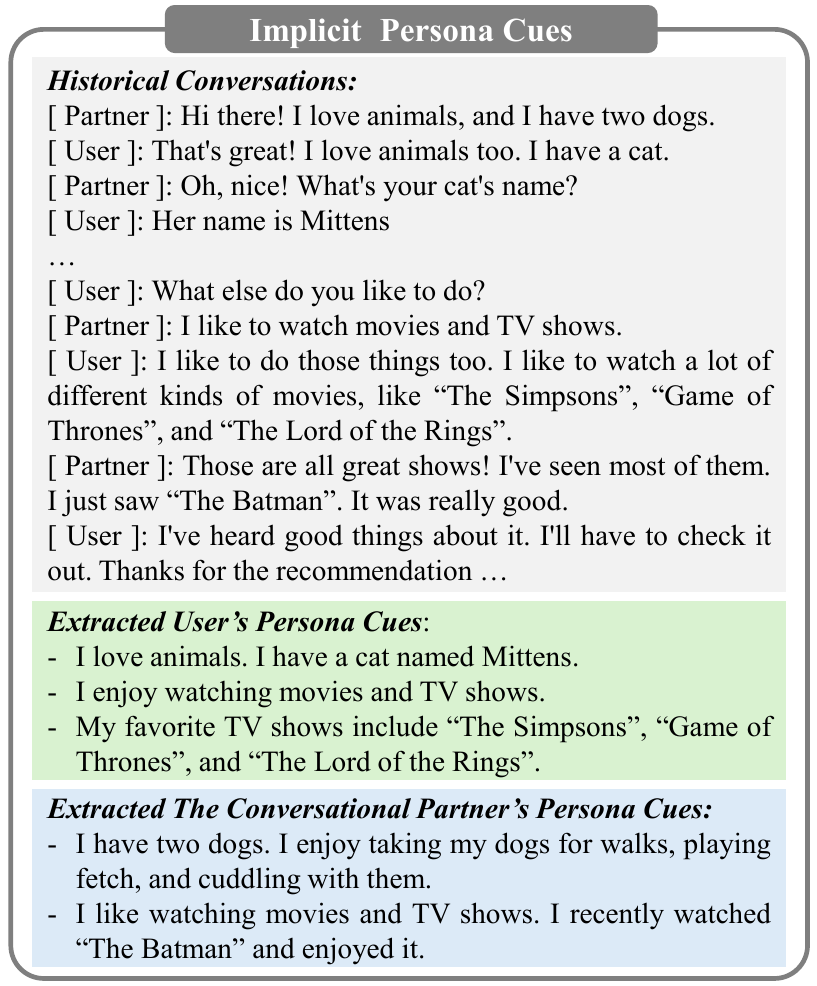}
\vspace{-1.5em}
  \caption{
  Examples of the extracted implicit persona cues from the
historical conversations.}
  % \vspace{-.5em}
\label{fig:implicit_persona_cues}
\end{minipage}
\hfill
  \begin{minipage}[t]{0.48\columnwidth}
     \centering
\includegraphics[width=1.\textwidth]{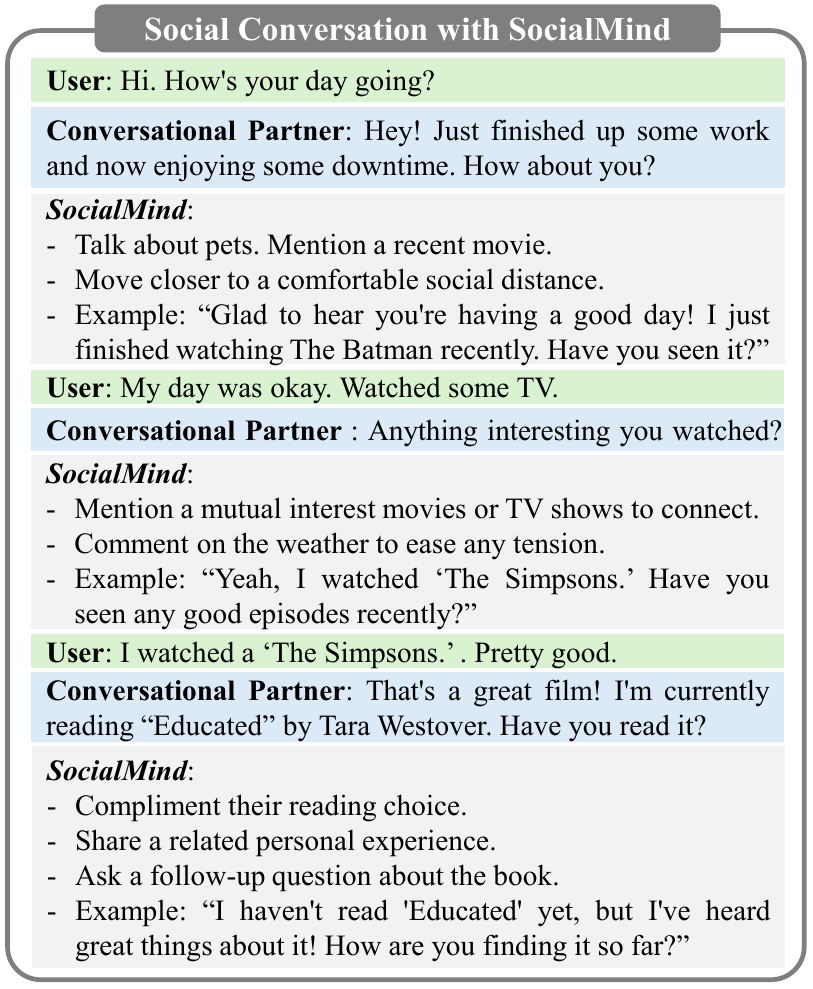}
\vspace{-1.5em}
  \caption{Examples of conversations and social suggestions provided by \workname, with implicit persona cues.
  }
  % \vspace{-.5em}
\label{fig:dialogue_implicit_persona_cues}
\end{minipage}
\vspace{-1.2em}
\end{figure}

\noindent\textbf{\textit{Observation 3: \workname~can generate customized social suggestions by considering the implicit persona cues of both parties.}}
Figure~\ref{fig:implicit_persona_cues} shows examples of the extracted implicit persona cues from the historical conversations of both parties.
Additionally, Figure~\ref{fig:dialogue_implicit_persona_cues} shows examples of the social suggestions provided by \workname, which recommend that users mention mutually interesting topics such as pets and movies.
These suggestions align with the personas of both parties, thereby boosting engagement.
Furthermore, even when the partner's persona cues are unavailable, such as during a first meeting without historical conversation information, \workname~can still steer the conversation to align with the user's implicit personas, thus enhancing engagement.

% \begin{figure}
%     \centering
%     \begin{subfigure}{0.48\columnwidth}
%         \centering
% \includegraphics[width=1\textwidth]{Evaluation/figs/implicit_persona_cues.png}
%         \vspace{-1.2em}
%         \caption{Examples of the extracted implicit persona cues from the historical conversations.}  \label{fig:implicit_persona_cues}
%     \end{subfigure}
%     \hfill
%     \begin{subfigure}{0.48\columnwidth}  
%         \centering 
%     \includegraphics[width=1\textwidth]{Evaluation/figs/dialogue_implicit_persona_cues.png}
%     \vspace{-1.2em}
%     \caption{Examples of conversations and social suggestions provided by SocialCoach, incorporating implicit persona cues.}    \label{fig:dialogue_implicit_persona_cues}
%     \end{subfigure}
%      \vspace{-1.em}
%     \caption{
%      Examples of social conversations and social suggestions SocialCoach and baselines.
%     }
% \label{fig:implicit_persona_cues}
%       % \vspace{-1.5em}
% \end{figure}

% \noindent\textbf{\textit{Observation 4: Social factor priors can help caching systems xxx.}}

% \noindent\textbf{\textit{Observation 4: \workname~ can leverage the external tools to xxx.
% }}

\subsection{Effectiveness of System Modules}
This section shows the effectiveness of each system module in \workname~and analyzes the impact of hyper-parameter in \workname.

\subsubsection{Effectiveness of Social Factor Prior.}
\begin{figure}
\begin{minipage}[t]{0.48\columnwidth}
     \centering
\includegraphics[width=0.80\textwidth]{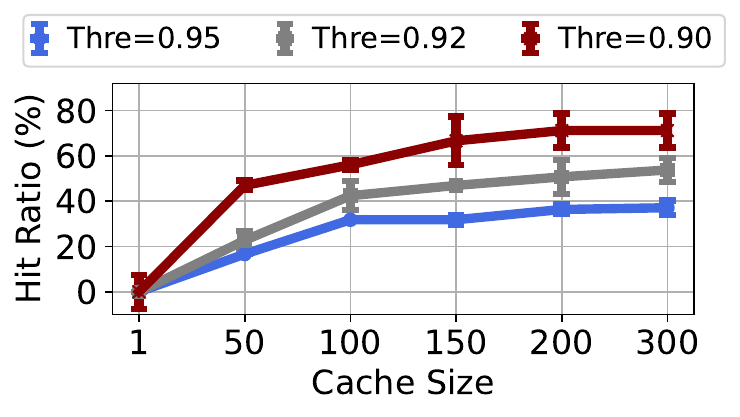}
\vspace{-1.5em}
  \caption{
  Impact of the cache size and threshold on hit ratio.}
  % \vspace{-.5em}
\label{fig:cache_size}
\end{minipage}
\hfill
  \begin{minipage}[t]{0.48\columnwidth}
     \centering
\includegraphics[width=0.80\textwidth]{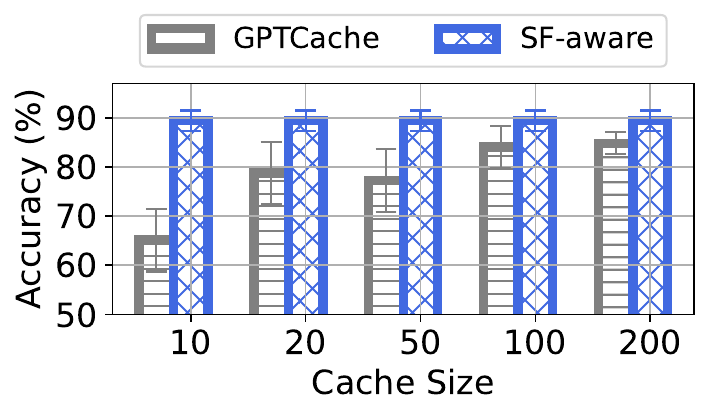}
\vspace{-1.5em}
  \caption{Effectiveness of social factor-aware cache.
  }
  % \vspace{-.5em}
  \label{fig:social_cache}
\end{minipage}
\vspace{-1em}
\end{figure}

We first conduct experiments to validate the effectiveness of social factor-aware cache (SF-aware), as shown in Figure~\ref{fig:social_cache}.
We collect the simulated conversations from our social assistant platform to construct a dataset, and split the dataset into 80\% and 20\% for caching and testing, respectively. 
Each sample contains the conversational partner's utterance, social suggestions, and the corresponding labels of social factors.
We use GPTcache \cite{bang2023gptcache} as the baseline, which directly uses the semantic similarity of the partner's utterance to select the social suggestions.
Accuracy reflects whether the retrieved results align with the query's social factor labels.
Figure~\ref{fig:cache_size} shows that the social factor-aware cache achieves 4.6\% higher accuracy than GPTCache.
This is because \workname~employs a social factor-aware cache that utilizes social factor priors to avoid semantically similar yet socially misaligned matches.
% matching queries that are semantically similar but violate social scenarios.
% Figure~\ref{fig:cache_size} shows that the social factor-aware cache achieves xxx\% higher accuracy, validating its effectiveness.
% With the cache size increases, xxx.

% \begin{figure}
% \begin{minipage}[t]{0.48\columnwidth}
%      \centering
% \includegraphics[width=0.80\textwidth]{Evaluation/figs/token_save.pdf}
% \vspace{-1.5em}
%   \caption{
%   Impact of the cache size and saved token ratio.}
%   % \vspace{-.5em}
% \label{fig:token_save}
% \end{minipage}
% \hfill
%   \begin{minipage}[t]{0.48\columnwidth}
%      \centering
% \includegraphics[width=0.80\textwidth]{Evaluation/figs/xx.pdf}
% \vspace{-1.5em}
%   \caption{xx.
%   }
%   % \vspace{-.5em}
%   \label{fig:xx}
% \end{minipage}
% \vspace{-1em}
% \end{figure}

% \textbf{\textit{w/o P}}.
% This method excludes the implicit persona adaptation module for generating social suggestions, while the remaining components are identical to those in \workname.

% \textbf{\textit{w/o N}}.
% This method omits the nonverbal cues integration module for generating social suggestions. All other components remain the same as in \workname.

\subsubsection{Effectiveness of Personas and Nonverbal Cues Integration.}
% Next, we validate the effectiveness of the Implicit Personas Adaptation module and the Nonverbal Cues Integration module in \workname.
To validate the effectiveness of the sub-modules in \workname, we omit the Implicit Personas Adaptation module and the Nonverbal Cues Integration module from \workname, denoting them as \textbf{\textit{w/o P}} and \textbf{\textit{w/o N}}, respectively.
% We compare the performance of social suggestions from \workname~ and two baselines: \textbf{\textit{w/o P}} and \textbf{\textit{w/o N}}.
% The \textbf{\textit{w/o P}} excludes the implicit persona adaptation module for generating social suggestions, while the \textbf{\textit{w/o N}} omits the nonverbal cues integration module for generating social suggestions.
All other components remain the same as in \workname.
Figure~\ref{fig:ablation_study} shows that \workname~achieves an average of 52.2\% higher \textit{Personalization} and 41.2\% higher \textit{Engagement} compared to \textbf{\textit{w/o P}}.
Additionally, results show that \workname~achieves average 35.7\% higher \textit{Nonverbal Cues Utilization} scores than \textbf{\textit{w/o N}}.
These findings validate the effectiveness of the implicit persona adaptation and nonverbal cues integration module in \workname.

\subsubsection{Impact of Hyper-parameter Settings.}
\label{Hyper-parameter Settings}
In this subsection, we conduct experiments to analyze the impact of hyper-parameters on \workname's performance.

\begin{figure}
\begin{minipage}[t]{0.48\columnwidth}
     \centering
\includegraphics[width=0.85\textwidth]{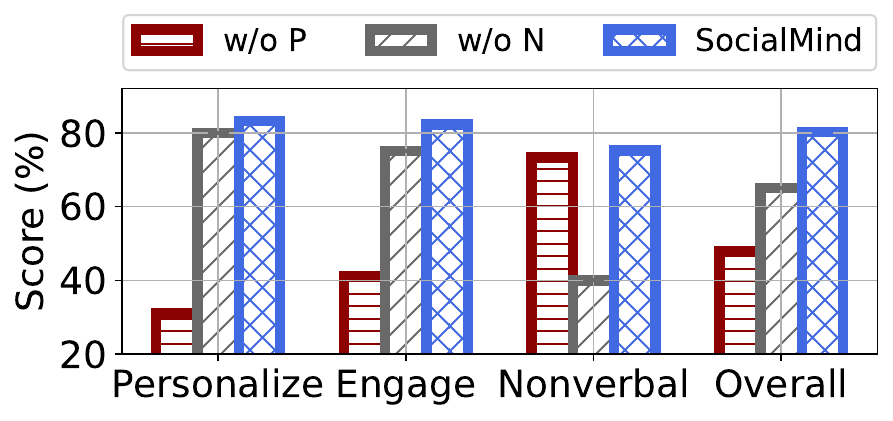}
\vspace{-1.em}
  \caption{
  % Effectiveness of the implicit personas adaptation module and the nonverbal cues integration module in \workname.
    Effectiveness of the personas and the nonverbal cues integration module in \workname.
    \textbf{\textit{w/o P}} and \textbf{\textit{w/o N}} means omitting the two modules from \workname, respectively.}
  % \vspace{-.5em}
\label{fig:ablation_study}
\end{minipage}
\hfill
  \begin{minipage}[t]{0.48\columnwidth}
     \centering
\includegraphics[width=0.9\textwidth]{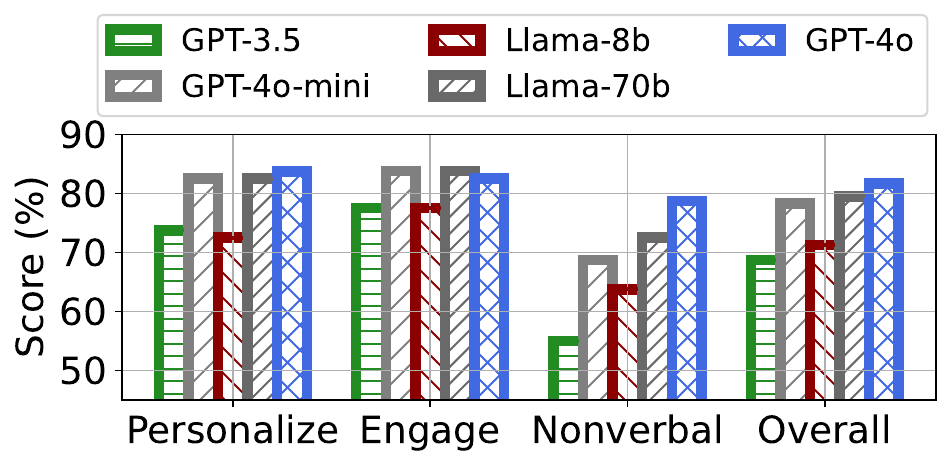}
\vspace{-1.em}
  \caption{Overall performance of the social suggestions generated by \workname~when using different LLMs as the base model.
  }
  % \vspace{-.5em}
  \label{fig:baseLLMs}
\end{minipage}
\vspace{-1em}
\end{figure}

\noindent\textbf{Impact of Base LLMs}.
First, we employ various LLMs as the base model in \workname~and compare their performance in generating social suggestions. 
The experimental LLMs include GPT-3.5-turbo, GPT-4o, GPT-4o-mini, Llama-3.1-8B-Instruct, and Llama-3.1-70B-Instruct.
Figure~\ref{fig:baseLLMs} shows that using GPT-4o as the base LLM achieves the highest overall performance among all the base LLMs. 
It achieves 6.2\% higher scores in \textit{nonverbal cues utilization} compared to the next top-performing base LLM but gains in \textit{personalization} and \textit{engagement} scores are not significant.
Notably, Llama-3.1-70B-Instruct performs only slightly lower than GPT-4o and achieves comparable overall performance. However, its open-source nature makes it a promising solution to reduce costs.
% However, its open-source make it a promising way to dep

% Whey using xxx, can achieves highest scores, xxx.
% Llama lower, but open-source and low cost.

\begin{figure}
    \centering
    \begin{subfigure}{0.48\columnwidth}
        \centering
\includegraphics[width=0.75\textwidth]{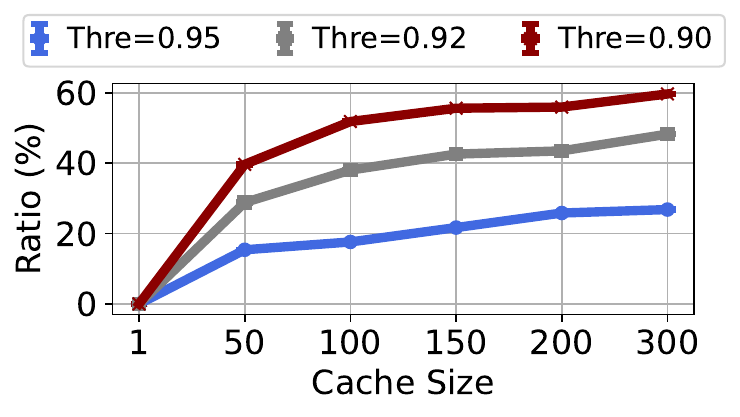}
        \vspace{-1.em}
        \caption{Token saving ratio of LLM inputs.}  \label{fig:token_saveinput}
    \end{subfigure}
    \hfill
    \begin{subfigure}{0.485\columnwidth}  
        \centering 
    \includegraphics[width=0.75\textwidth]{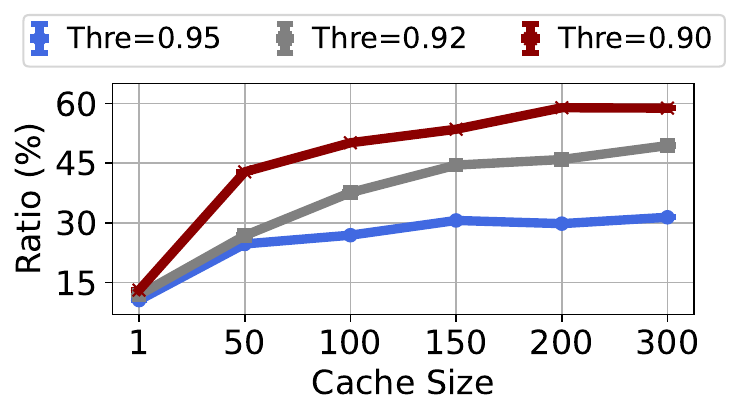}
    \vspace{-1.em}
    \caption{Token saving ratio of LLM outputs.}    \label{fig:token_saveoutput}
    \end{subfigure}
     \vspace{-1.em}
    \caption{
   Impact of the threshold and catch size on the LLM token saving ratio.
    }
\label{fig:token_save}
      \vspace{-1.5em}
\end{figure}

\noindent\textbf{Impact of Cache Size and Threshold}.
Next, we evaluate the impact of the cache size and threshold on the performance of the social factor-aware cache.
Figure~\ref{fig:cache_size} shows that the cache hit rate increases with cache size. 
In the initial phase of deployment, \workname~is still unfamiliar with the user's environment and background, making it difficult to achieve cache hits to speed up suggestion generation. 
However, \workname~ will continuously monitor the user's social interactions and update the cache.
When the cache size reaches 200, \workname~can achieve 36.3\% cache hit rate under a threshold of 0.95. To ensure high-quality social suggestions, \workname~employs a relatively high threshold in the cache to avoid delivering irrelevant responses.
Figure~\ref{fig:token_save} shows that with a 0.95 threshold and cache size of 300, the input token saving ratio is 26.8\% and the output token saving ratio is 31.4\%.
% that with a threshold of 0.95, a cache size of 300 can achieve input token savings of 26.8\% and output token savings of 31.4\%, respectively.

\begin{figure}
    \centering
    \begin{subfigure}{0.32\columnwidth}
        \centering
        \includegraphics[width=0.85\textwidth]{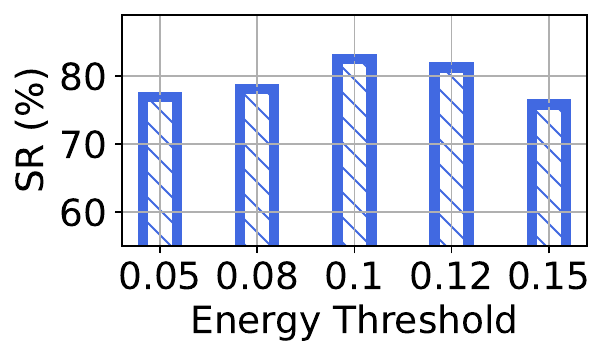}
        \vspace{-1.0em}
        \caption{Parameter impact on audio.}  \label{fig:para_speech}
    \end{subfigure}
    \hfill
    \begin{subfigure}{0.32\columnwidth}  
        \centering 
        \includegraphics[width=0.85\textwidth]{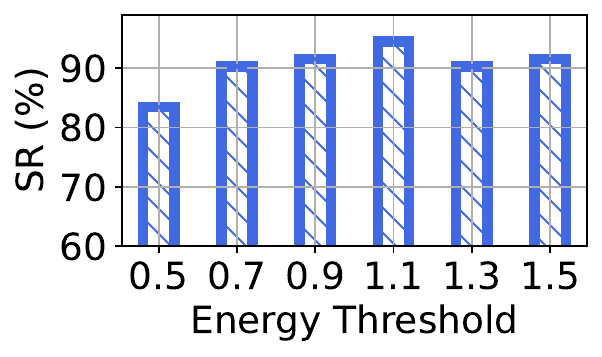}
        \vspace{-1.0em}
        \caption{Parameter impact on Vibration.}    
\label{fig:para_IMU}
    \end{subfigure}
    \hfill
    \begin{subfigure}{0.32\columnwidth}  
        \centering 
\includegraphics[width=0.75\textwidth]{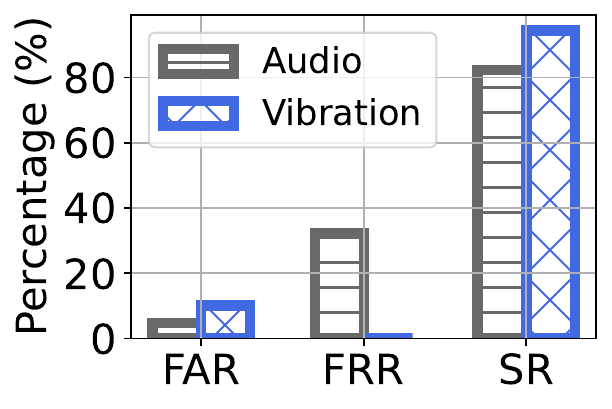}
        \vspace{-1.0em}
        \caption{Detection performance.}    
\label{fig:overall}
    \end{subfigure}
     \vspace{-1.0em}
    \caption{The primary user detection performance of \workname~and the impact of parameters on different solutions
.}
\label{fig:primary_user_detection_performance}
\vspace{-.5em}
\end{figure}

\noindent\textbf{Impact of Threshold in Primary User Detection}.
\label{performance_primary_user_detection}
We also evaluate the performance of primary user detection using audio-based and vibration-based solutions.
Since the fingerprint-based solution has privacy concerns \cite{microsoft,li2020vocalprint}, we employ the voice volume-based approach, following the settings in EarVoice \cite{chen2024enabling}.
% For the voice volume-based approach, we follow the settings in EarVoice \cite{chen2024enabling}, using the audio signal's energy in the 0$\sim$1000 Hz range for detection. 
For the vibration-based approach, we calculate the vibration signal's energy within the 3$\sim$10 Hz range.
We use a 1-second time window for identification and use false accept rate (FAR), false reject rate (FRR), and success rate (SR) metrics for evaluation \cite{chen2024enabling}.
Figure~\ref{fig:primary_user_detection_performance} shows that the energy threshold significantly affects SR for both solutions.
When the energy threshold is set to 0.1 and 1.1, the audio-based and vibration-based solutions achieve optimal SR, respectively. 
Additionally, Figure~\ref{fig:primary_user_detection_performance} shows that the vibration-based approach achieves a 32.3\% lower FRR and a 12.1\% higher SR than the voice volume-based approach. 
This is because the voice volume-based method struggles to accurately identify the primary user when the user's volume does not exceed that of the conversational partner, as illustrated in Figure~\ref{fig:audio_waveform_timefreq}.

% Additionally, vibration-based solutions achieve the highest primary detection performance, with a 12.1\% higher SR, validating the effectiveness of our design.

% Figure~\ref{fig:primary_user_detection} shows that the energy threshold significantly affects the performance of SR for both audio-based and vibration-based solutions.
%  When the energy threshold is set to 0.1 and 1.1, audio-based and vibration-based solution achieves its highest SR, respectively. Additionally, Figure\ref{fig:primary_user_detection} shows that vibration-based solutions achieve the highest primary detection performance, with a 12.1\% higher SR.

\vspace{-.5em}
\subsection{Real-world Evaluation}

\subsubsection{System Performance}
% We evaluate the system performance of \workname~in the real world, including energy consumption and system latency.
% The face and pose tracking in \workname~is locked to 3 frames per second at 640×480 resolution, with power consumption under 2 watts, comparable to a standard camera app. The RayNEO~X2 glasses can support continuous operation of \workname~for about 2 hours. In practical use, \workname~is activated during specific social interactions, allowing the system to operate for extended hours.
% Additionally, we measure the system latency of \workname. 
% The glasses communicate with the server via HTTPS, with data transfer rates below 100 KB/s.
% For nonverbal cues perception, the latency of pose recognition on glasses is within 70 ms. 
% Regarding social suggestion generation, \workname~employs a multi-tier collaborative reasoning strategy, where the average latencies for cache and LLM reasoning are 50 ms and 2.8 s, respectively.
% However, since \workname~uses an intention inference-based LLM reasoning strategy, the LLM takes partial utterances of the conversational partner for reasoning without waiting for the end of their speech. This allows \workname~to update social suggestions on AR glasses even while the partner is speaking, ensuring that the latency for user experience does not exceed that time.
We evaluated \workname's real-world performance, focusing on energy use and system latency. To conserve power, face and pose tracking is capped at three frames per second at 640×480 resolution, keeping power consumption under 2 watts—similar to a standard camera app—and supporting up to 70 minutes of use. Users can also activate the system manually to extend battery life.

System latency measurements show data transfer rates below 100 KB/s over HTTPS. Pose/face tracking latency for nonverbal cues is within 70 ms, while \workname’s multi-tier reasoning strategy achieves average latencies of 50 ms for cache and 2.8 s for LLM processing. Using intention-based inference, \workname~analyzes partial conversational utterances, allowing real-time updates on the AR glasses without waiting for speech completion, thus ensuring low latency and a smooth user experience.

% First, the pose recognition in \workname is limited to 3 frames per second at a 640$\times$480 resolution, keeping power consumption under 2 watts if—comparable to a standard camera app.
% RayNEO X2 glasses supports such \workname~continuously work about 2 hours.  
% Additonally, in practical use, \workname~is activated when certain social interactions occurs, so it our system can support more hours working.

% energy, estimated to 2 hrs if xx. But \workname works can ..
% Second, we measure the system latency.
% pose recognition time
% offloading time, bandwidth
% llm time
% cache time
% no more than such latency because xxx.

% We limit pose recognition to 3 frames per second at a 640$\times$480 resolution, keeping power consumption under 2 watts—comparable to a standard camera app. 
% The glasses communicate with the server via HTTPS, with data transfer rates below 100 KB/s.

\subsubsection{User Study}
\label{sec_user_study}
% This section presents the user study of \workname.
% We recruited 20 participants for real-world testing, which involved two types of social interaction scenarios. In each scenario, participants wore glasses and engaged in conversations with non-player characters with the assistance of \workname.
We recruited 20 participants for real-world testing. Participants wore glasses and engaged in conversations with a partner with the assistance of \workname.
% and engaged in conversations with non-player characters.

% \subsubsection{Settings of Social Interactions}
\noindent\textbf{Settings of Social Interactions}.
Our user survey shows that a significant number of participants experience social awkwardness in sudden interactions, such as engaging with company superiors or meeting unfamiliar colleagues.
Therefore, we evaluate \workname~in these scenarios to validate its effectiveness in helping users manage sudden social interactions and avoid embarrassment.
Specifically, each participant wears AR glasses equipped with \workname~(system implementation see \S~\ref{implementation}).
Participants are instructed to engage in a social conversation with a partner. They can either freely talk with their partner or refer to the social suggestions displayed on the glass screen.
The conversation topics can include daily experiences such as work or entertainment.
% Specifically, we assigned a fixed non-player character (NPC) to proactively initiate conversations with the participant, where their social relationship is peer-to-peer. The participant interacts with the NPC with the assistance of \workname.
Afterward, participants complete a six-question questionnaire to provide feedback and ratings on their experience with \workname.
% Additionally, each participant is required to complete a questionnaire with six questions after using \workname.
% This questionnaire is designed to gather participants' feedback and ratings regarding their experience with \workname. The details of the questionnaire are as follows:
The questionnaire details are as follows:
\begin{itemize}[leftmargin=*]
\item 
\textbf{\textit{Q1}}: Have you ever used an eyewear social assistant during live social interactions before?

\item
\textbf{\textit{Q2}}: Are you satisfied with the suggestions from the eyewear social assistant? If so, to what extent?

\item
\textbf{\textit{Q3}}: 
Is the latency of social suggestion generation acceptable during live social interactions? 
% If so, to what extent?
% Is the latency of social suggestion generation in the assistive system acceptable during live social interactions? If so, to what extent?

\item
\textbf{\textit{Q4}}: 
Are you willing to use this eyewear social assistant during live social interactions? If yes, to what extent?

\item
\textbf{\textit{Q5}}: 
Are you willing to communicate with
another person who is using such an
eyewear social assistant?

\item
\textbf{\textit{Q6}}: 
Do you find our social assistant's design to be innovative and functional?  If so, to what extent?

\end{itemize}

% determine whether the system can help users manage sudden social interactions and avoid embarrassment.
% Specifically, we designated a fixed non-player character to initiate conversations with the participant proactively, and the social relation between the participant and the non-player character is peer-to-peer.
% The participant interacts with the non-player character with the assistance of \workname.

\begin{figure}
    \centering
    \begin{subfigure}{0.32\columnwidth}
        \centering
        \includegraphics[width=0.85\textwidth]{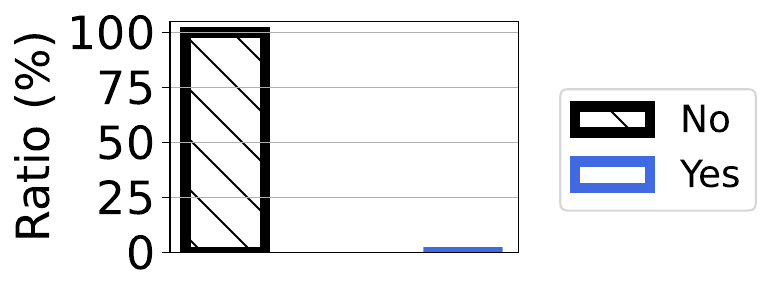}
        \vspace{-0.9em}
        \caption{Have you ever used an eyewear social assistant during live social interactions before?}  \label{fig:user_study_Q1}
    \end{subfigure}
    \hfill
    \begin{subfigure}{0.32\columnwidth}  
        \centering 
        \includegraphics[width=1\textwidth]{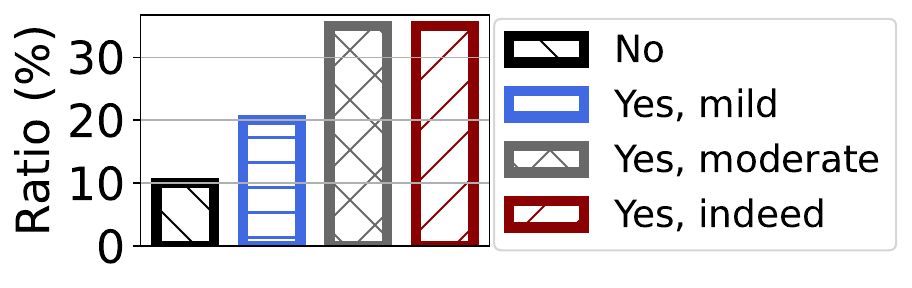}
        \vspace{-2.0em}
        \caption{Are you satisfied with the suggestions from the eyewear social assistant? If so, to what extent?}    
        \label{fig:user_study_Q2}
    \end{subfigure}
    \hfill
    \begin{subfigure}{0.32\columnwidth}  
        \centering 
        \includegraphics[width=1\textwidth]{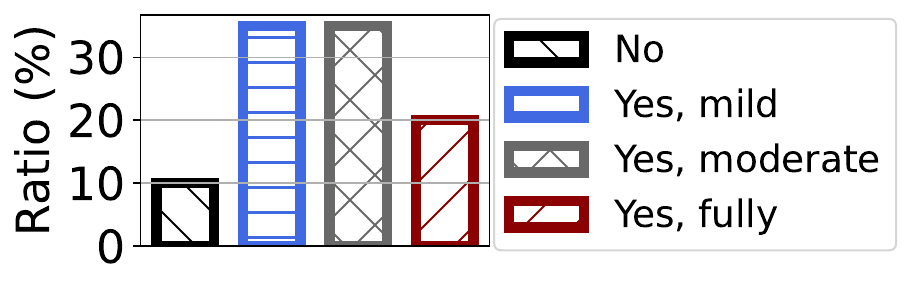}
        \vspace{-2.0em}
        \caption{Is the latency of social suggestion generation acceptable during live social interactions? If so, to what extent?}    
        \label{fig:user_study_Q3}
    \end{subfigure}
     % \vspace{-0.5em}
    \vskip\baselineskip
    \begin{subfigure}{0.32\columnwidth}   
        \centering 
        \includegraphics[width=1\textwidth]{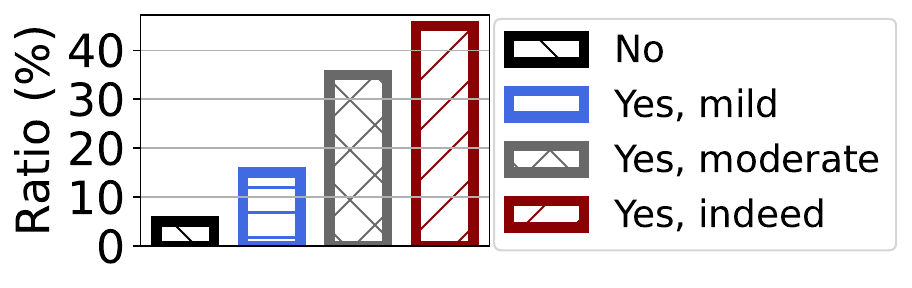}
        \vspace{-2.0em}
        \caption{Are you willing to use this eyewear social assistant during live social interactions? If yes, to what extent?}    \label{fig:user_study_Q4}
    \end{subfigure}
    \hfill
    \begin{subfigure}{0.32\columnwidth}   
        \centering 
        \includegraphics[width=1\textwidth]{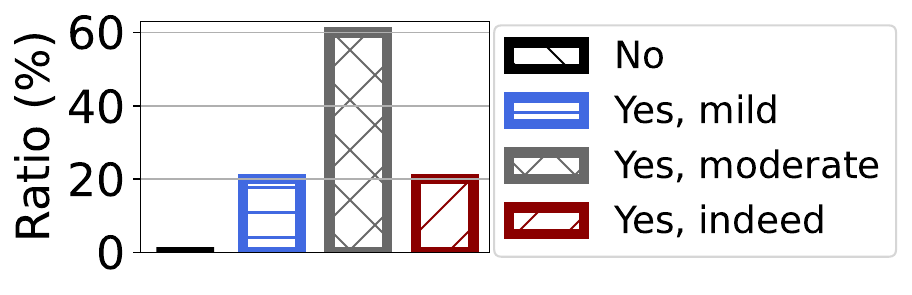}
        \vspace{-2.0em}
        \caption{Are you willing to communicate with
another person who is using such an
eyewear social assistant?}    \label{fig:user_study_Q5}
    \end{subfigure}
    \hfill
    \begin{subfigure}{0.32\columnwidth}  
        \centering 
    \includegraphics[width=1\textwidth]{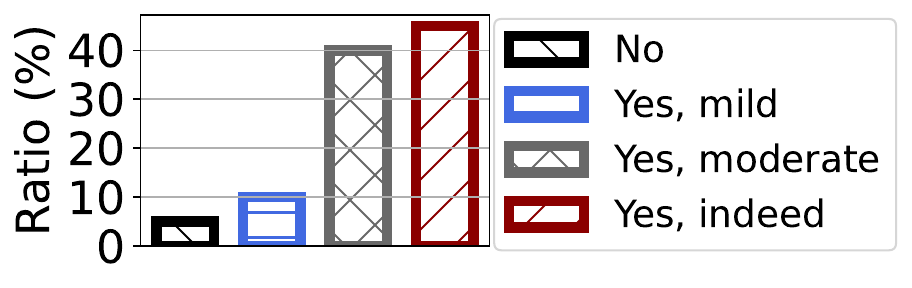}
        \vspace{-2.0em}
        \caption{Do you find our social assistant's design to be innovative and functional?  If so, to what extent?}    
        \label{fig:user_study_Q6}
    \end{subfigure}
     \vspace{-.5em}
    \caption{\workname's user study results.
    }
    \label{fig:user_study}
      \vspace{-1.5em}
\end{figure}

% \vspace{-1.5em}
\noindent\textbf{Results and Insights}.
% \subsubsection{Results and Insights} 
Figure~\ref{fig:user_study} shows participants' feedback after using \workname~in live social interactions.
Results show that none of the participants have prior experience with eyewear social assistants.
Consequently, 85\% believe our system to be both novel and practical.
Additionally, nearly 70\% are satisfied with the suggestions from the social assistant, considering them helpful during face-to-face live social interactions.
Moreover, over 90\% find the latency in suggestion generation acceptable, as it does not disrupt the natural flow of the dialogue.
% Moreover, over 90\% find the latency in suggestion generation acceptable during face-to-face live social conversations, as it does not disrupt the natural flow of the dialogue.

% that none of the participants ever use such eyewear social assistant during live social interactions, and thus xx\% believe our systems to be novel and practical.

Many participants expressed a willingness to use our system, believing it can genuinely assist in handling unexpected social situations. 
Interestingly, while over 95\% are eager to use such an assistive system during live social interactions, fewer participants, around 80\%, are inclined to interact with someone else who is also using the system.
This suggests that people prefer to have an assistant for their own benefit, rather than for others.
% Interestingly, while many volunteers are eager to use such an assistive system during live social interactions (more than xx\%), they are less inclined to interact with someone else who is also using the system (more than xx\% exhibit unwillingness). Additionally, several participants noted that it could be beneficial even if the other parties use this system because it may help them understand and interpret their intentions more clearly, thereby enhancing social interactions and relationships. However, some participants raised privacy concerns, feeling monitored by our system. They prefer the system to be used in public social scenarios rather than personal spaces. It’s important to note that our system only uses the camera to calculate pose and offload the data to the server without recording or offloading raw images.
Some participants find that \workname~is particularly useful for them, especially when they are unsure of what to say, as it provides helpful clues and prevents embarrassment.
Additionally, some participants think that while their spoken English is not very strong, their reading skills are sufficient.
\workname, as a social assistance system, not only enhances their social skills but also supports them in practicing and improving spoken English.
Moreover, some participants also highlight the need for a training process to become familiar with using the system, such as balancing using their own cognitive abilities versus referring to the text displayed on the glasses, ensuring a natural conversation flow.
% Moreover, some participants also highlight the need for a training process to become familiar with using the system, such as balancing using their own cognitive abilities versus referring to the text displayed on the glasses without interrupting the natural flow of live social conversations.
Additionally, some participants find the system helpful when they lose focus or ``zoned out'' during conversations, as it allows them to review the conversation and reduce cognitive workload during interactions.
The feedback from participants reveals a promising market potential for \workname.

%% file: Discussion/Discussion.tex
\section{Discussion and Limitations}
% \vspace{-1.em}
% Although we have validated the effectiveness of \workname, there are still limitations to address.
% In this section, we discuss the limitations, potential solutions, and future work.

% % \noindent\textbf{Adaptively Display of Social Suggestions}.
% \noindent\textbf{Pretrained Social LLMs as Base Models}.
% In this work, we experiment with various general LLMs as base models in \workname, including GPT-3.5-turbo, GPT-4o, GPT-4o-mini, Llama-8b, and Llama-70b. 
% Our system is also scalable to other pretrained LLMs in the social domain, such as the Tianji series \cite{tianji2024}. 
% Leveraging these pretrained social LLMs as the base LLMs can further enhance social suggestion quality, as these models incorporate extensive knowledge of social etiquette and politeness.

% \noindent\textbf{Proactive Assistance Service Trigger}.
% Feedback from some participants in the user study indicates that the social suggestions are displayed too frequently. 
% For certain users, these suggestions may not be necessary during social interactions. 
% In fact, our user survey (\S~\ref{user survey}) shows that social suggestions are most needed in specific scenarios, such as interacting with superiors, feeling awkward or unsure of what to say, and not understanding the other person. 
% Enabling \workname~to proactively detect these situations and trigger the social assistance service is a crucial direction for future research.

\noindent\textbf{System Scalability}.
\workname~can be extended to multi-modal LLMs for nonverbal cue extraction in an open-ended manner~\cite{liu2024visual}, further improving the system's generalization. Given the limited resources, an edge-cloud collaborative framework can also be considered~\cite{yang2023edgefm}.
Additionally, \workname~is scalable to multi-person scenarios. We plan to utilize the camera view to identify individual speakers, enhancing the system’s capability to manage multiple conversational partners simultaneously.

\noindent\textbf{User’s Nonverbal Cues}. 
Integrating the wearer’s facial expressions could enhance assistance quality.
However, forward-facing cameras that capture these expressions—such as those used in devices like the VisionPro~\cite{apple2024visionpro}—add considerable weight, impacting comfort significantly. For daily use, our current solution avoids heavy AR goggles.
With hardware advances, like Meta Orion~\cite{meta2024orion}, we plan to incorporate nonverbal cues from both the user and conversational partners, enhancing the experience through more comprehensive social suggestions. 

\noindent\textbf{Next Steps}. 
% This study establishes a foundational design and validation framework for using AR glasses to assist in general social interactions.
% Moving forward, we aim to extend this solution to address  specific needs in professional cases, such as individuals with Social Anxiety Disorder (SAD), Autism Spectrum Disorder (ASD), Selective Mutism, and Agoraphobia.
% For these applications, we plan to collaborate closely with experts and therapists, integrating their specialized insights and domain knowledge to tailor AR interventions effectively.
% This approach will ensure that the design is therapeutic and practical for these user groups in real-world use.
% This study establishes a foundational framework for designing and validating AR glasses as aids in general social interactions. Building on this, our next steps involve adapting this solution for more complex, real-world applications, including multi-person conversations and specific use cases, such as supporting individuals with Social Anxiety Disorder (SAD), Autism Spectrum Disorder (ASD), Selective Mutism, and Agoraphobia. 
This study establishes a foundational framework for designing and validating AR glasses as aids in general social interactions. Building on this, our next steps involve adapting this solution for more complex, real-world applications, including multi-person conversations and specific use cases, such as supporting individuals with Social Anxiety Disorder (SAD) and Autism Spectrum Disorder (ASD). 
% For multi-person scenarios, we plan to leverage the camera view to identify individual speakers, enhancing the system’s capability to manage multiple conversational partners simultaneously.
In professional applications, we will collaborate with therapists and domain experts to incorporate tailored therapeutic insights, ensuring the AR interventions are both effective and responsive to these user groups' unique needs. 
% This approach aims to make our design both practical and therapeutically valuable in real-world scenarios.

% that the design is both therapeutic and practical, meeting the unique requirements of these user groups in real-world contexts.

% the feedback from some participants show that the social suggestions are too frequently display. 
% For some users, some of the social conversation, these users may not require these social suggestions.
% In fact, according to our user survey, most scenarios that user need social suggestions are When interacting with specific individuals, such as superiors, and When you feel awkward or don't know what to say, When you don’t understand what the other person is saying.
% Enable \workname~to proactively detect these situations and then trigger the social assistance service is the future work.

% \noindent\textbf{Adapt to Multi-modal Data}.
% We conduct experiments to validate the performance of \workname using different base LLMs, including the GPT series and Llama series. 